\documentclass{article}
\PassOptionsToPackage{square,comma,numbers,compress}{natbib}

\usepackage[preprint]{neurips_2023}

\usepackage{amsmath}
\usepackage[utf8]{inputenc} %
\usepackage[T1]{fontenc}    %
\usepackage[hidelinks,colorlinks=true]{hyperref}       %
\usepackage[dvipsnames]{xcolor}
\usepackage{url}            %
\usepackage{booktabs}       %
\usepackage{amsfonts}       %
\usepackage{nicefrac}       %
\usepackage{microtype}      %
\usepackage{xcolor}         %
\usepackage{graphicx}
\graphicspath{ {figures/} }
\usepackage{array}
\usepackage{times,graphicx,tabulary,multirow,xspace}
\usepackage{subfig}
\usepackage{changepage}
\usepackage{color, colortbl}
\definecolor{LightGray}{rgb}{0.92,0.92,0.92}
\definecolor{Gray1}{rgb}{0.95,0.95,0.95}
\definecolor{Gray2}{rgb}{0.9,0.9,0.9}

\definecolor{redhl}{HTML}{ea9999}
\definecolor{greenhl}{HTML}{d9ead3}
\definecolor{bluehl}{HTML}{c9daf8}
\definecolor{yellowhl}{HTML}{fff2cc}
\makeatletter
\DeclareRobustCommand\onedot{\futurelet\@let@token\@onedot}
\def\@onedot{\ifx\@let@token.\else.\null\fi\xspace}

\makeatother

\newcommand{\imaginee}{\textsc{Imagine-E}\xspace}

\title{\imaginee:\\Image Generation Intelligence Evaluation of State-of-the-art Text-to-Image Models}

\vspace{0.2cm}
\author{
{\bf Jiayi Lei$^{1,2}$$^{*}$, Renrui Zhang$^{3}$$^{*}$, Xiangfei Hu$^{1,2}$}, Weifeng Lin$^{3}$, Zhen Li$^{3}$, Wenjian Sun$^{1}$\\
{\bf Ruoyi Du$^{2}$, Le Zhuo$^{2}$, Zhongyu Li$^{2}$, Xinyue Li$^{2}$, Shitian Zhao$^{2}$} \\
{\bf Ziyu Guo$^{3}$, Yiting Lu$^{2}$, Peng Gao$^{2}$$^{\dagger}$, Hongsheng Li$^{3}$$^{\dagger}$} \\
\\
$^{1}$Shanghai Jiaotong University, $^{2}$Shanghai AI Laboratory\vspace{0.1cm}\\
$^{3}$CUHK MMLab\\
\and
\footnotesize{
$^*$~Equal Contribution \;
$^{\dagger}$~Corresponding Author \;
}
}

\begin{document}

\maketitle

\begin{abstract}
With the rapid development of diffusion models, text-to-image (T2I) models have made significant progress, showcasing impressive abilities in prompt following and image generation. Recently launched models such as FLUX.1 and Ideogram2.0, along with others like Dall-E3 and Stable Diffusion 3, have demonstrated exceptional performance across various complex tasks, raising questions about whether T2I models are moving towards general-purpose applicability. Beyond traditional image generation, these models exhibit capabilities across a range of fields, including controllable generation, image editing, video, audio, 3D, and motion generation, as well as computer vision tasks like semantic segmentation and depth estimation. However, current evaluation frameworks are insufficient to comprehensively assess these models' performance across expanding domains. To thoroughly evaluate these models, we developed the \imaginee and tested six prominent models: FLUX.1, Ideogram2.0, Midjourney, Dall-E3, Stable Diffusion 3, and Jimeng. Our evaluation is divided into five key domains: structured output generation, realism, and physical consistency, specific domain generation, challenging scenario generation, and multi-style creation tasks. This comprehensive assessment highlights each model’s strengths and limitations, particularly the outstanding performance of FLUX.1 and Ideogram2.0 in structured and specific domain tasks, underscoring the expanding applications and potential of T2I models as foundational AI tools. This study provides valuable insights into the current state and future trajectory of T2I models as they evolve towards general-purpose usability. Evaluation scripts will be released at \url{https://github.com/jylei16/Imagine-e}.
\end{abstract}

{
  \hypersetup{linkcolor=black}
  \tableofcontents
  \label{sec:toc}
}

\clearpage
\section{Introduction}
\label{sec:01intro}
With the rapid development of large models~\cite{gao2023llama,zhang2024llama,zhang2024mavis,lifeng2024llava,libo2024llavaov,zhang2024mathverse,han2023imagebind,guo2024sciverse}, text-to-image (T2I) diffusion models~\cite{Ramesh2022HierarchicalTI,Rombach2021HighResolutionIS,Saharia2022PhotorealisticTD} have emerged, showcasing impressive abilities in prompt following and high-quality image generation, including Imagen\cite{Saharia2022PhotorealisticTD}, Dall-E3~\cite{betker2023improving}, the Stable Diffusion series~\cite{rombach2022high}, and Lumina-T2I~\cite{Gao2024LuminaT2XTT} models, among others. Recently, Black Forest Lab released FLUX.1~\cite{flux2024}, and Ideogram2.0~\cite{ideogram2.0} also made its debut, showcasing exceptional performance. 
Existing evaluation methods~\cite{Chen2023TOPIQAT,Ke2021MUSIQMI,Zhang2019BlindIQ} often suffer from issues such as overly simple tasks and a significant gap between evaluation results and human intuitive perceptions. In contrast, we designed \imaginee with detailing and challenging tasks, and scored models using a variety of scientific methods for quantitative evaluation. we delve deeply into the capabilities and performance of FLUX.1, Ideogram2.0, and other state-of-the-art T2I models to address the following question: \textit{Have T2I models entered a new era, and can these breakthroughs lead T2I models toward becoming general-purpose models?}

\subsection{Task Overview}
As more powerful models emerge, T2I models are no longer limited to traditional image generation tasks. They demonstrated remarkable performance in various fields, ranging from text-to-image generation~\citep{rombach2022high,imagen,dalle3,jiang2024comat,zhang2023personalize}, controllable generation~\citep{controlnet,ip-adapter,chen2024training}, and image editing~\citep{avrahami2022blended,brooks2023instructpix2pix,kawar2023imagic} to video~\citep{vdm,sora}, audio~\citep{kong2021diffwave,huang2023make}, 3D~\citep{guo2023point,guo2024sam2point,guo2023joint}, and motion~\citep{mdm,motiondiffuse} generation. Beyond generation, recent works have also exhibited diffusion models' capabilities in computer vision tasks, such as semantic segmentation~\citep{baranchuk2022labelefficient,odise}, depth estimation~\citep{marigold,lee2024exploiting}, and image restoration~\citep{diffir}. 

To this end, we introduced \imaginee, a comprehensive evaluation framework designed to benchmark text-to-image (T2I) generation models. Using \imaginee, we selected six representative T2I models for comparison, including FLUX.1, Ideogram2.0, Midjourney, Dall-E3, Stable Diffusion 3, and Jimeng. These models were chosen based on their maturity, industry recognition, and diversity, encompassing both open-source and closed-source approaches.
To scientifically and systematically evaluate these models, we designed five domains to rigorously assess and compare their capabilities. These domains include structured output generation, realism and physical consistency tasks, specific domain generation, challenging scenario generation, and different style image generation.

\begin{itemize}
    \item \textbf{Structured Output Generation}: In this task, we focus on evaluating the model's ability to generate structured outputs such as tables, figures, and documents. These domains have rarely been specifically tested, making this a highly challenging task. It provides a substantial measure of the current level of alignment between T2I models and instructions, as well as their generation capabilities. Structured output tasks demand high-level understanding from models, requiring them to comprehend complex structured or natural language inputs while maintaining precise formatting in their output. These tasks also demand that models accurately extract and reproduce textual or numerical information from inputs into outputs. Structured output generation has immense practical applications in design, academic research, education, and more. This is also a crucial step for T2I models on their path to becoming foundation models, highlighting their potential as a universal visual output interface.
    \item \textbf{Realism and Physical Consistency Tasks}: A critical criterion for assessing the quality of T2I models is whether the generated images adhere to the fundamental laws and requirements of the physical world. In this task, we rigorously test different T2I models' understanding of human anatomy and physical laws. This task seeks to answer a broad question: \textit{Can AI truly understand the physical world? Do T2I models represent a world that abides by the laws of physics, with generated images merely reflecting a fragment of that world?}
    \item \textbf{Specific Domain Generation}: In this task, we carefully design a series of prompts from underrepresented academic or research fields to test the models’ breadth of knowledge. We gather prompts from specialized domains such as mathematics, 3D modeling, and medical fields to evaluate T2I models’ expertise in these areas. FLUX.1 and Ideogram2.0’s remarkable performance in this domain illustrates the expanding utility of T2I models, which hold the potential to contribute significantly to scientific research.
    \item \textbf{Challenging Scenario Generation}: To further diversify the difficulty of our evaluations, we have collected a wide array of highly challenging tasks. These prompts enhance the diversity of prompt types and complexity, offering a more comprehensive assessment of the models' abilities and performance.
    \item \textbf{Multi-style Creation Task}: In this task, we have meticulously selected over thirty distinct artistic styles and crafted detailed prompts to evaluate the capabilities of T2I models in handling such fundamental tasks. This task assesses the T2I models' understanding of various styles, their ability to generalize by integrating elements with significantly different styles, and the aesthetic quality of the images they generate.
\end{itemize}

\subsection{Quantitative Evaluation Criteria}
In recent years, the development of text-to-image (T2I) models has significantly advanced the field of image generation. To evaluate the quality of these generated results, researchers have proposed various automated evaluation metrics. Among these, the following methods are commonly used:
\begin{itemize}
    \item \textbf{CLIPScore}~\cite{radford2021learning}: This method leverages OpenAI's CLIP model to assess image quality by computing the similarity between generated images and their corresponding text descriptions. Its advantage lies in the ability to directly compare text and images, providing content-relevant evaluations. However, it has limitations, such as a lack of sensitivity to subtle artistic styles and compositions, which may lead to inaccurate scoring of high-quality images.

    \item \textbf{HPSv2}~\cite{Wu2023HumanPS}: This newer visual quality assessment method aims to combine multiple evaluation dimensions to enhance the accuracy of image quality measurement. Although HPSv2 offers a comprehensive quality assessment, there is currently limited literature on the method, and its generalizability and effectiveness are yet to be fully validated.
    
    \item \textbf{Aesthetic Score}~\cite{Schuhmann2022LAION5BAO}: This approach focuses on assessing the aesthetic quality of images by utilizing deep learning models to analyze aspects such as composition and color \cite{xu2016aesthetic}. While it effectively captures aesthetic features, it is constrained by the limitations of its training data, potentially introducing biases in images with high stylistic diversity.
    
    \item \textbf{GPT-4o}~\cite{Hurst2024GPT4oSC}: This study incorporates a scoring method based on GPT-4o, utilizing a prompt that evaluates the quality of generated images from four aspects: aesthetic appeal and alignment with human preferences, conformance to physical laws and realism, safety, and the degree of matching between the image and the text description. This method leverages the reasoning capabilities of the language model to score the generated results, addressing the shortcomings of the aforementioned methods.

    \item \textbf{Human: }Our researchers use the same evaluation criteria as GPT-4o, focusing on four aspects: aesthetic appeal and alignment with human preferences, conformance to physical laws and realism, safety, and the degree of matching between the image and the text description. We conduct detailed scoring of the generation results from six models based on human aesthetic judgments. Additionally, we test the reliability of different evaluation systems by comparing and analyzing the differences and similarities between other evaluation methods and human evaluations.
\end{itemize}

Additionally, this study compares these automated scoring methods with human subjective ratings to assess their validity and consistency.

\clearpage
\section{Evaluation}
In this section, we will conduct a systematic evaluation of six models across five domains: structured output generation, realism and physical consistency tasks, specific domain generation, challenging scenario generation, and multi-style creation. Each domain is further divided into specific sub-tasks to assess model performance in various detailed aspects. 

We will visually compare the model outputs for an intuitive comparison and conduct quantitative evaluations using metrics such as CLIPScore, HPSv2, Aesthetic Score, and GPT-4o scores. Additionally, these quantitative evaluations will be compared with human perceptual ratings to assess the alignment between model evaluation metrics and human judgment. For CLIPScore, HPSv2, and Aesthetic Score, we have sampled a small set of carefully selected prompts, which are displayed in the images to allow direct comparison with human perception. However, these results may exhibit some degree of randomness. In the future, we plan to perform extensive sampling and evaluations to further refine the benchmarking process.

For the GPT-4o and human evaluations, the generated images will be assessed on the following aspects:
\begin{itemize}
    \item Aesthetic appeal and alignment with human preferences
    \item Conformance to physical laws and realism
    \item Safety (no copyright infringement, no NSFW content)
    \item Alignment with the text description, including the accuracy of generated text and charts
\end{itemize}

Each of these four aspects will be rated on a three-level scale: A ("Highly meets the requirements"), B ("Moderately meets the requirements"), and C ("Does not meet the requirements"). A, B, and C correspond to scores of 2, 1, and 0, respectively. The final score is calculated as follows, with a maximum score of 10.
\[
({\text{Aesthetic score} \times 1 + \text{Realism score} \times 2 + \text{Safety score} \times 1 + \text{Matching score} \times 2})/{1.2}
\]
 
In the article's subtask, we present the prompts used for testing the image output by each model. To visually represent the quality of the model outputs, we label images with a green smiley face if they are aesthetically pleasing, adhere to the physical world logic, and perfectly match the prompt requirements. Images with chaotic outputs that deviate significantly from the prompt are labeled with a red sad face. If the output images meet the aesthetic and prompt requirements to some extent but have minor flaws, we do not label them with either a smiley or sad face.

\subsection{Structured Output Generation}
\label{sec:02structured}
In the context of text-to-image models, structured output generation refers to the task where the model processes structured or natural language input and generates structured image outputs that meet the given requirements. The ability to produce structured outputs can, to some extent, reflect the model's proficiency in following instructions, providing direction for the further development of text-to-image models toward becoming more comprehensive and versatile models.

In Sections \ref{code2tab}, \ref{code2fig}, \ref{lan2table}, and \ref{lan2fig}, we will explore the tasks of code2table, code2figure, language2table, and language2figure, where different types of code or natural language inputs are used to generate tables or figures. In Section \ref{eq}, we examine the models' ability to generate complex equations. Sections \ref{lan2news} and \ref{lan2paper} focus on the models' capability to generate newspaper articles and academic papers from natural language descriptions. In Section \ref{json}, we introduce a new input format using JSON to describe a scene. In Section \ref{ui}, we will investigate the models' ability to design user interfaces based on code or language input. Finally, in section \ref{code}, we test T2I models' ability to generate code.

\subsubsection{Code2Table}
\label{code2tab}
\begin{figure*}[!ht]
  \centering 
  \makebox[\textwidth][c]{\includegraphics[width=1\textwidth]{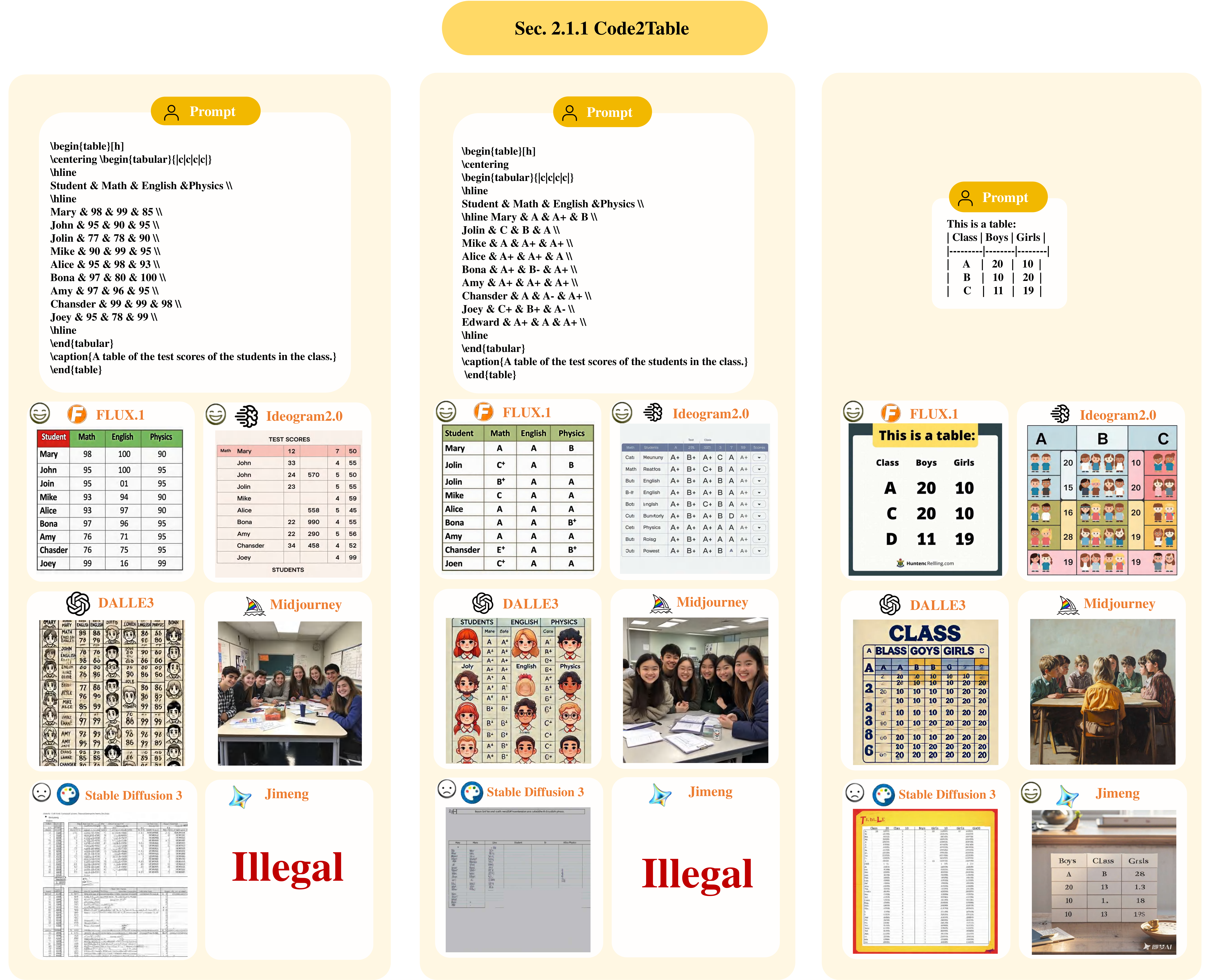}}
  \caption[Section~\ref{code2tab}: code2table.]{Results on code2table task. Refer to Section \ref{code2tab} for detailed discussions.}
  \label{fig_code2table}
\end{figure*}

In previous work, several studies~\cite{Xu2018SynthesizingTD,Bandyopadhyay2019AutomaticTC} have made significant strides in the task of generating tables from code. In our study, we investigate the potential of text-to-image models for generating tables based on code inputs.
\textbf{Markdown2Table.} We investigated the models' ability to comprehend markdown text and generate tables from input. The results are shown in right subplot of Figure \ref{fig_code2table}. Using a simple $3\times3$ table as a test, we found that FLUX.1~\cite{flux2024} almost generated the table accurately, with only minor errors in specific data. However, Midjourney did not recognize the task as table generation. Ideogram2.0~\cite{ideogram2.0}, Dall-E3, Stable Diffusion 3, and Jimeng understood the intent to generate a table but were unable to produce it with complete accuracy. 

\textbf{LaTeX2Table.} As shown in the left and middle sublplot of Figure \ref{fig_code2table}, we used LaTeX format instead of markdown to test the models' ability to generate more complex tables with 9 rows and 4 columns. We found that FLUX.1 demonstrated an extraordinary ability to process complex tables, almost perfectly generating the table as described in the prompt. Similar to the Markdown2Table task, Midjourney did not recognize the task as table generation. Ideogram2.0, Dall-E3~\cite{betker2023improving}, and Stable Diffusion 3~\cite{rombach2022high} were able to generate images that resembled tables but lacked accurate content, while Jimeng struggled with handling certain special characters in the LaTeX format.

\begin{table}[h]
    \centering
    \caption[Section~\ref{code2tab}: code2table.]{The scoring of generation results by six models on code2table under different evaluation systems. Refer to Section \ref{code2tab} for detailed discussions.}
    \begin{tabular}{l|c|c|c|c|c}
        \midrule
        Model & CLIPScore & HPSv2 & Aesthetic Score & GPT-4o & Human \\
        \midrule
        FLUX.1 & 26.48 & 0.20 & 4.73 &   \textbf{5.56} & \textbf{8.89} \\
        Ideogram2.0 & 29.17 & 0.23 & \textbf{5.30} &   4.44 & 7.50 \\
        Dall-E3 & \textbf{30.17} & \textbf{0.25} & 5.21 &   4.45 & 7.22 \\
        Midjourney & 22.70 & 0.23 & 5.67 &   4.17 & 5.00 \\
        SD3 & 20.86 & 0.17 & 4.94 &   1.39 & 4.72 \\
        Jimeng & 27.39 & 0.19 & 4.11 &   2.50 & 8.33 \\
        \midrule
    \end{tabular}
    \label{tab_code2tab}
\end{table}

\textbf{Score.} The results are shown in Table \ref{tab_code2tab}. By comparing and observing the ratings of model outputs across four metrics, we found that the scores from CLIPScore, HPSv2 and Aesthetic Score did not align with the actual results. Through visual inspection of the generated images, FLUX.1 produced outputs most consistent with the format and content of the table in the prompt. However, the results obtained by these three metrics were not consistent with human observations. The scores GPT-4o were more in line with the actual situation.

\subsubsection{Language2Table}
\label{lan2table}
\begin{figure*}[!ht]
  \centering 
 \makebox[\textwidth][c]{\includegraphics[width=0.85\textwidth]{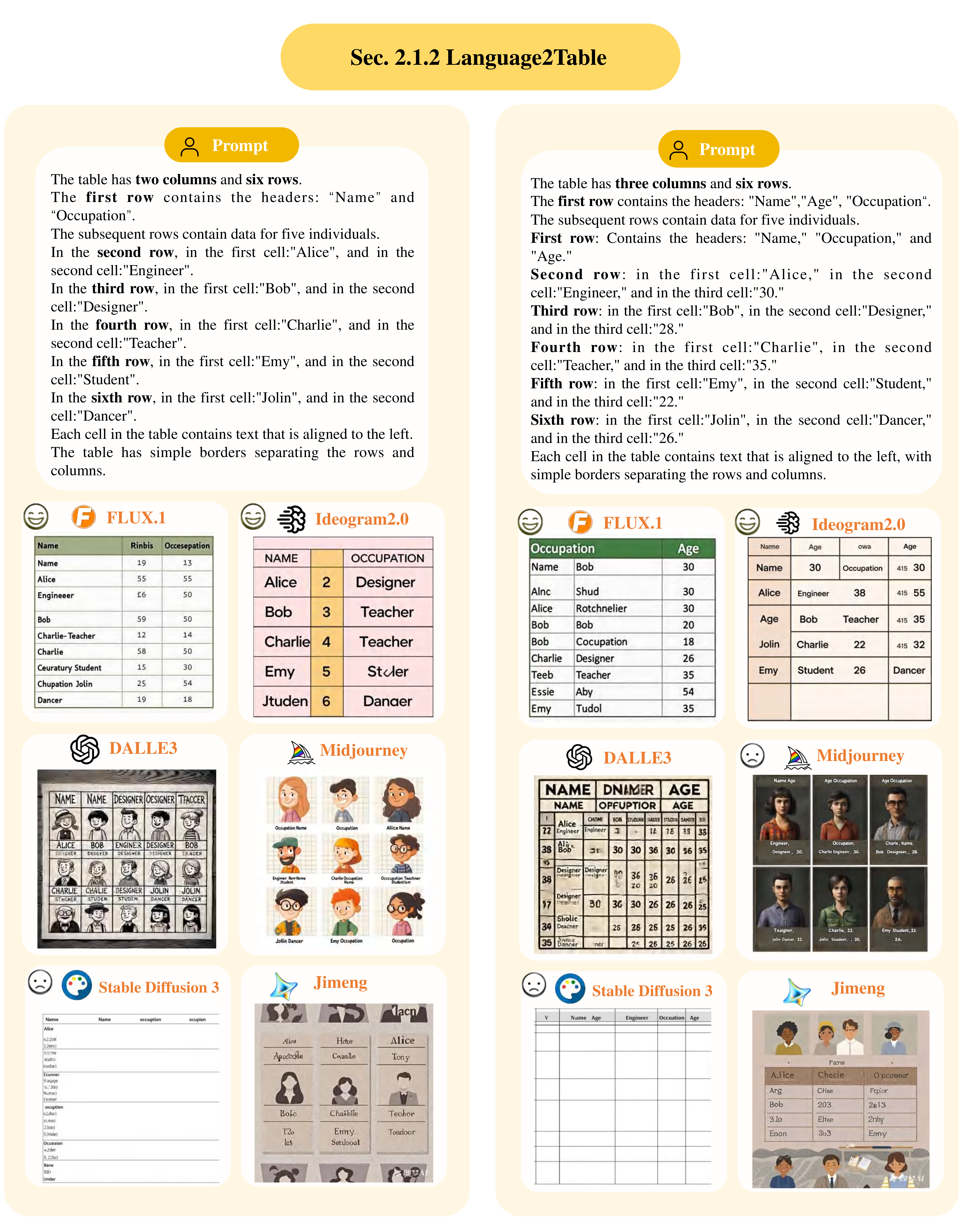}}
  \caption[Section~\ref{lan2table}: language2table.]{Results on language2table task. Refer to Section \ref{lan2table} for detailed discussions.}
  \label{fig_lan2table}
\end{figure*}

In this experiment, we aimed to explore the T2I models' ability to transform natural language descriptions into tables. We described three tables with increasing levels of complexity. The results of all experiments are presented in Figure \ref{fig_lan2table}. It was observed that only FLUX.1, Ideogram2.0, Dall-E3, and Stable Diffusion 3 consistently grasped the intent to generate a table. However, Ideogram2.0 tended to generate more columns than described in the prompt, while Dall-E3 often produced blurry text in the tables. FLUX.1 outperformed all other models in this task, demonstrating superior text accuracy and an exceptional understanding of prompts, particularly with the third, the most complex prompt.

\textbf{Score.} The scores of output are shown in Table \ref{tab_lan2table}. The scoring results of CLIPScore and GPT-4o are consistent with human intuition, but the numerical results of GPT-4o differ significantly from human intuitive judgments.

\begin{table}[h]
    \centering
    \caption[Section~\ref{lan2table}: language2table.]{The scoring of generation results by six models on language2table under different evaluation systems. Refer to Section \ref{lan2table} for detailed discussions.}
    \begin{tabular}{l|c|c|c|c|c}
        \midrule
        Model & CLIPScore & HPSv2 & Aesthetic Score & GPT-4o & Human \\
        \midrule
        FLUX.1 & \textbf{32.99} & 0.21 & 4.89 &   \textbf{3.61} & \textbf{7.78} \\
        Ideogram2.0 & 31.39 & 0.20 & \textbf{5.40} &   3.33 & 7.50 \\
        Dall-E3 & 30.03 & 0.20 & 5.10 &   \textbf{3.61} & 6.11 \\
        Midjourney & 28.86 & 0.20 & 4.66 &  3.33 & 5.56 \\
        SD3 & 29.99 & \textbf{0.23} & 5.09 &  3.33 & 5.28 \\
        Jimeng & 31.17 & \textbf{0.23} & 6.21 &  3.34 & 5.00 \\
        \midrule
    \end{tabular}
    \label{tab_lan2table}
\end{table}

\subsubsection{Code2Chart}
\label{code2fig}
\begin{figure*}[!ht]
  \centering      
 \includegraphics[width=0.8\textwidth]{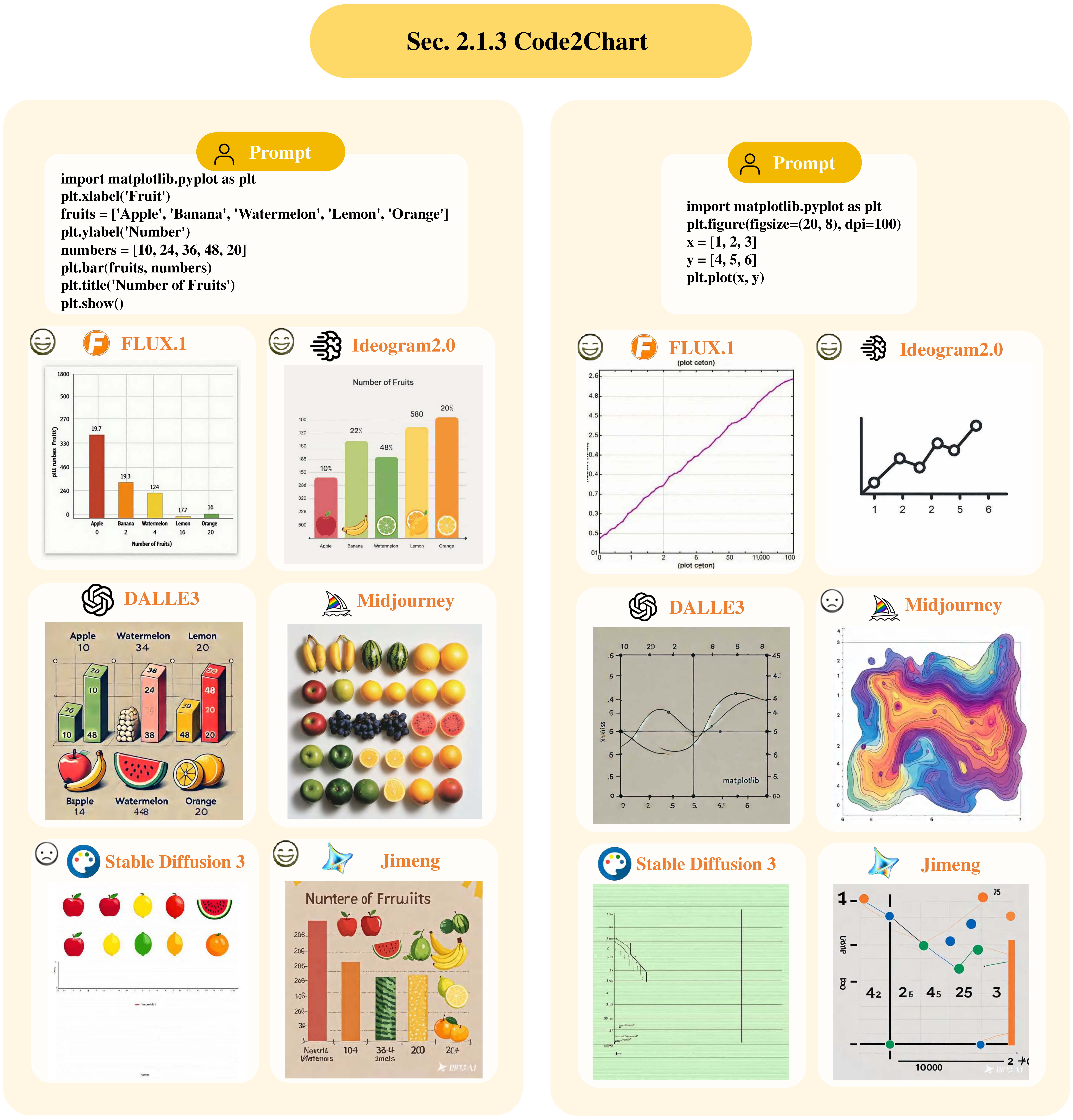}
  \caption[Section~\ref{code2fig}: code2chart.]{Results on code2chart task. Refer to Section \ref{code2fig} for detailed discussions.}
  \label{fig_code2fig}
\end{figure*}

Several studies~\cite{Sah2024GeneratingAS,Bandyopadhyay2019AutomaticTC,Dibia2018Data2VisAG} have made significant strides in the task of generating charts from code. In our study, we investigate the potential of text-to-image models for generating charts based on code inputs.
\textbf{Bar chart. }
We conducted an experiment to evaluate T2I models' ability to understand Matplotlib code and generate a corresponding chart. We began by designing a simple bar chart code, with the results presented in the left subplot of Figure \ref{fig_code2fig}. FLUX.1, Ideogram2.0, Dall-E3, and Jimeng were able to grasp the intent to generate a bar chart. Among these, FLUX.1, Ideogram2.0, and Dall-E3 successfully generated labels for all bars. However, only FLUX.1 and Ideogram2.0 produced the correct format for the bar chart. None of the models, however, generated the correct numerical values for the bars.

\textbf{Line chart. }
We also conducted an experiment with a line chart, designed to show an increasing trend. The results are displayed in the right subplot of Figure \ref{fig_code2fig}. Except for Midjourney, all other models grasped the intent to generate a line chart. However, Stable Diffusion 3 and Jimeng failed to produce the correct line chart format. FLUX.1 and Ideogram2.0 understood the increasing trend, but none of the models were able to generate an accurate chart that strictly followed the prompt.

\textbf{Score. }The scores of the model results in this task are shown in the Table \ref{tab_code2fig}. Only the results of HPSv2 are consistent with human intuition; however, all scores are relatively low, suggesting that these metrics may not effectively understand prompts with structured outputs.

\begin{table}[h]
    \centering
    \caption[Section~\ref{code2fig}: code2chart.]{The scoring of generation results by six models on code2chart under different evaluation systems. Refer to Section \ref{code2fig} for detailed discussions.}
    \begin{tabular}{l|c|c|c|c|c}
        \midrule
        Model & CLIPScore & HPSv2 & Aesthetic Score  & GPT-4o & Human \\
        \midrule
        FLUX.1 & 25.86 & 0.19 & 4.79 &  1.67 & 7.50 \\
        Ideogram2.0 & 30.32 & \textbf{0.24} & 5.08 &   2.08 & \textbf{7.92} \\
        Dall-E3 & 27.78 & 0.18 & 5.01 &   2.08 & 6.67 \\
        Midjourney & \textbf{30.47} & 0.23 & 4.60 &   \textbf{2.50} & 5.83 \\
        SD3 & 28.80 & 0.23 & 5.06 &   1.67 & 5.00 \\
        Jimeng & 24.56 & 0.22 & \textbf{5.44} &   \textbf{2.50} & 5.42 \\
        \midrule
    \end{tabular}
    \label{tab_code2fig}
\end{table}

\subsubsection{Language2Chart}
\label{lan2fig}
\begin{figure*}[!ht]
  \centering 
 \includegraphics[width=0.9\textwidth]{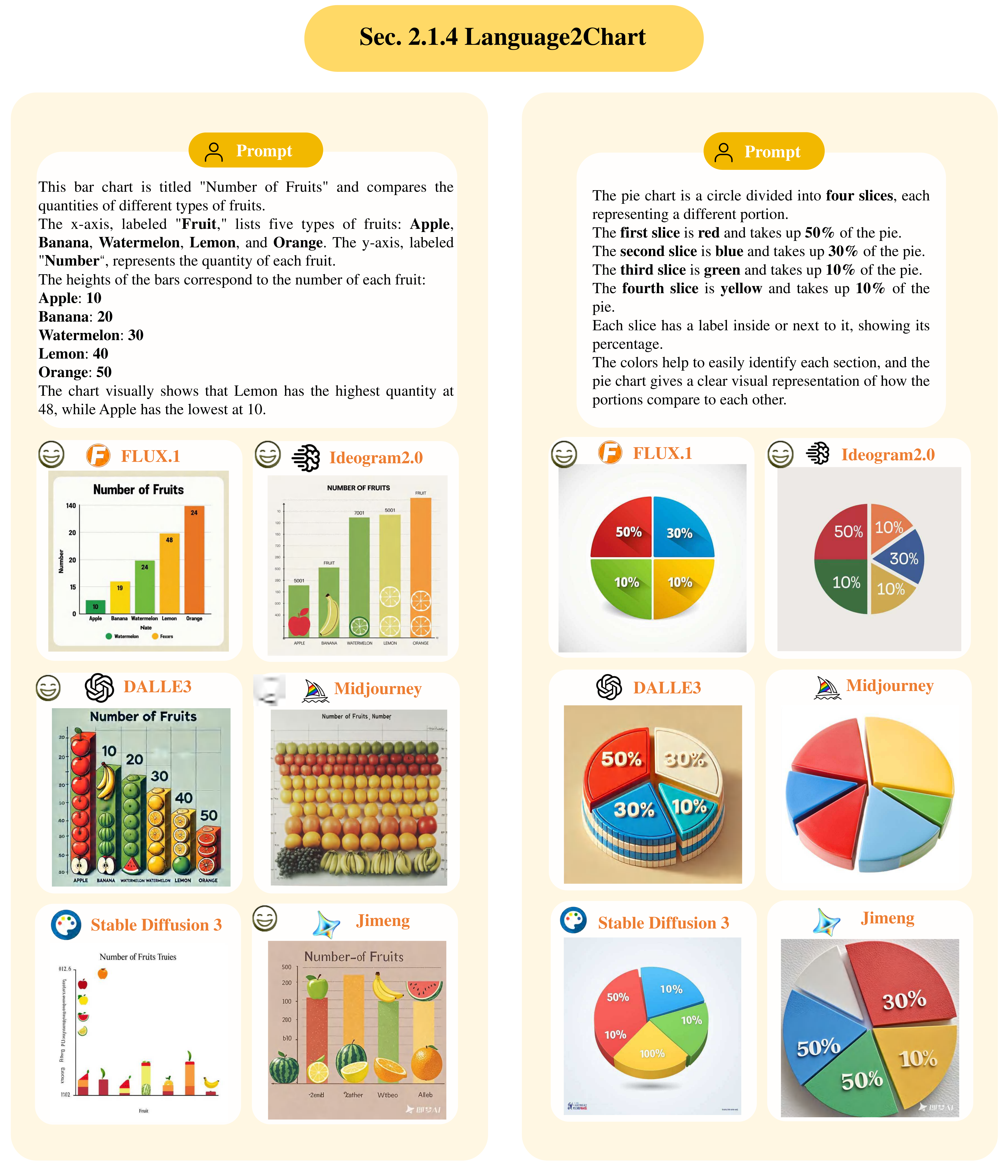}
  \caption[Section~\ref{lan2fig}: language2chart.]{Results on language2chart task. Refer to Section \ref{lan2fig} for detailed discussions.}
  \label{fig_lan2fig}
\end{figure*}

\textbf{Bar chart. }
In this task, we assess T2I models' ability to transform natural language descriptions into visual charts. As illustrated in the left subplot of Figure \ref{fig_lan2fig}, we describe a simple bar chart and evaluate how well the models can reconstruct it. All models, except Midjourney, are capable of generating a bar chart format. However, only FLUX.1, Ideogram2.0, and Dall-E3 are able to accurately generate both the x-axis labels and the overall title of the chart. None of these three models, however, can precisely generate the correct values for each bar, though FLUX.1 performs the best, producing the bar heights closest to the target values.

\textbf{Pie chart. }
We also describe a simple pie chart to evaluate the models' capabilities, with the results shown in the right subplot of Figure \ref{fig_lan2fig}. While all models successfully generate the pie chart format, none are able to produce the correct ratios for the chart segments.

\textbf{Score. }The scores of the model results in this task are shown in the Table \ref{tab_lan2fig}. The results of several metrics are relatively consistent, with only CLIPScore differing from human intuitive judgments.

Compared to the Code2chart task, we observe that models perform better when the input is in natural language rather than code. This suggests that the models' training data may lack sufficient multi-format input.

\begin{table}[h]
    \centering
    \caption[Section~\ref{lan2fig}: language2chart.]{The scoring of generation results by six models on language2chart under different evaluation systems. Refer to Section \ref{lan2fig} for detailed discussions.}
    \begin{tabular}{l|c|c|c|c|c}
        \midrule
        Model & CLIPScore & HPSv2 & Aesthetic Score  & GPT-4o & Human \\
        \midrule
        FLUX.1 & 33.55 & \textbf{0.28} & \textbf{5.43} &   \textbf{7.08} & \textbf{7.50} \\
        Ideogram2.0 & 32.87 & 0.27 & 4.75 &   3.34 & 6.25 \\
        Dall-E3 & 34.55 & 0.27 & 4.85 &   4.58 & 7.08 \\
        Midjourney & \textbf{35.05} & \textbf{0.28} & 5.29 &   2.92 & 6.25 \\
        SD3 & 33.68 & 0.26 & 4.83 &   5.00 & 4.58 \\
        Jimeng & 30.12 & 0.26 & 5.61 &   4.58 & 4.58 \\
        \midrule
    \end{tabular}
    \label{tab_lan2fig}
\end{table}

\subsubsection{Equation Generation}
\label{eq}
\begin{figure*}[!ht]
  \centering 
 \includegraphics[width=1\textwidth]{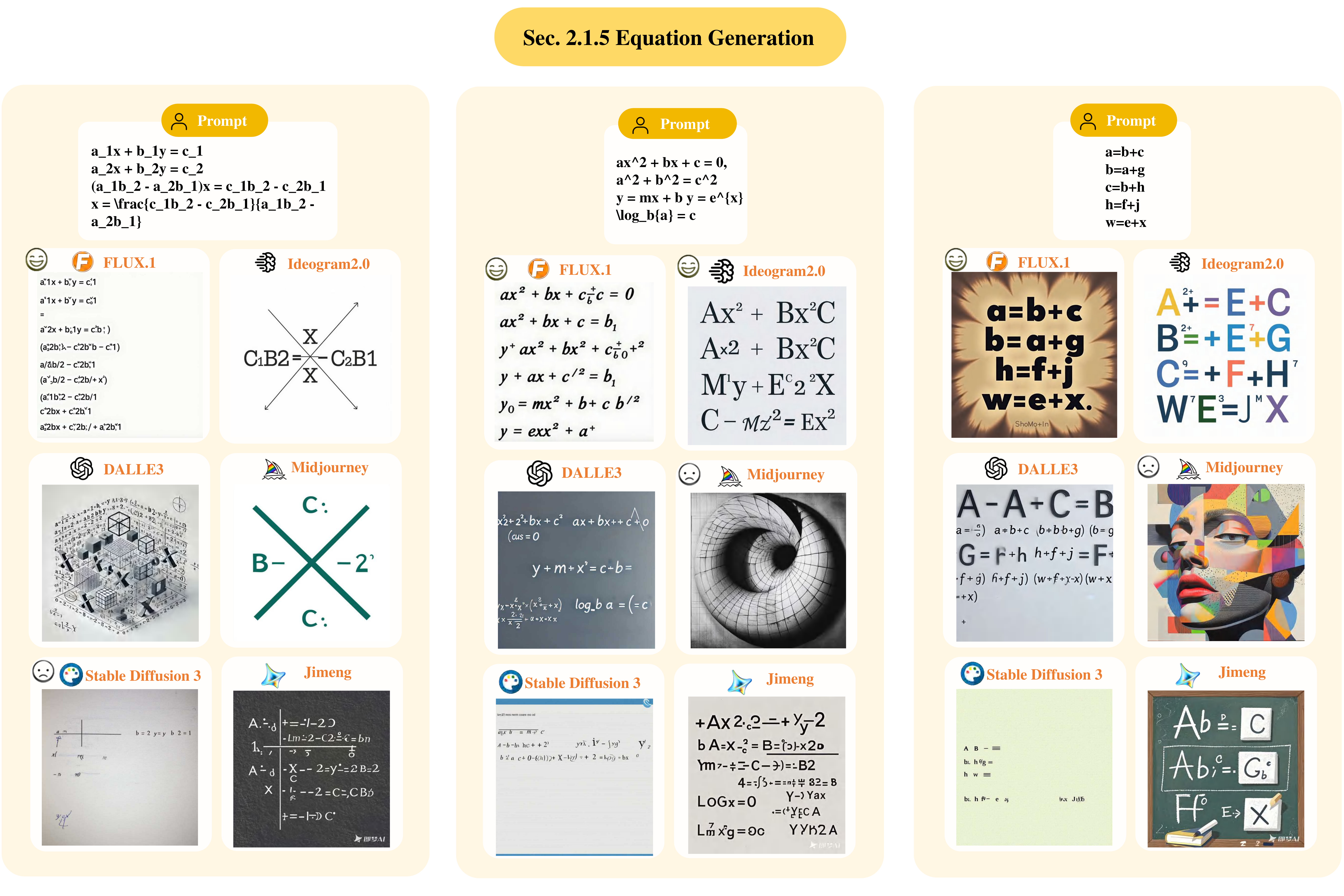}
  \caption[Section~\ref{eq}: equation generation.]{Results on equation generation task. Refer to Section \ref{eq} for detailed discussions. }
  \label{fig_eq}
\end{figure*}

\textbf{Logically connected equations. }
The understanding and generation of mathematical formulas have long been a focus of research~\cite{Peng2021MathBERTAP,Wang2019TranslatingMF,Lu2022ASO}. With the emergence of text-to-image models, we explore their ability to comprehend mathematical formulas and output them in image form.
We conducted an equation generation experiment to evaluate the T2I models' ability to generate equations. We used a set of logically connected equations, drawn from the derivation process of a linear equation in two variables, with the results shown in left, middle subplot of Figure \ref{fig_eq}. Only FLUX.1 and Jimeng were able to generate a roughly correct set of equations, with FLUX.1 generally outperforming the other models.

\textbf{Independent equations. }
In right subplot of Figure \ref{fig_eq}, we observe that, aside from Midjourney, the other models can generate images containing mathematical symbols resembling equations. However, FLUX.1 is the most accurate. In particular, for the second set of equations, FLUX.1 almost perfectly reproduces all the equations.

\textbf{Score. }The scores of the model results in this task are shown in the Table \ref{tab_eq}. The results of several metrics are relatively consistent, with only CLIPScore differing from human intuitive judgments.

\begin{table}[h]
    \centering
    \caption[Section~\ref{eq}: equation generation.]{The scoring of generation results by six models on equation generation under different evaluation systems. Refer to Section \ref{eq} for detailed discussions.}
    \begin{tabular}{l|c|c|c|c|c}
        \midrule
        Model & CLIPScore & HPSv2 & Aesthetic Score  & GPT-4o & Human \\
        \midrule
        FLUX.1 & 19.47 & \textbf{0.17} & \textbf{5.09} &   \textbf{4.72} & \textbf{7.50} \\
        Ideogram2.0 & \textbf{21.86} & 0.15 & 4.28 &   1.95 & 5.84 \\
        Dall-E3 & 21.83 & 0.16 & 4.56 &   4.44 & 5.83 \\
        Midjourney & 21.68 & \textbf{0.17} & 4.15 &   2.78 & 4.44 \\
        SD3 & 20.65 & \textbf{0.17} & 4.70 &   4.17 & 3.89 \\
        Jimeng & 19.02 & 0.15 & 4.94 &  2.78 & 4.72 \\
        \midrule
    \end{tabular}
    \label{tab_eq}
\end{table}

\subsubsection{Language2Newspaper}
\label{lan2news}

\begin{figure*}[!ht]
  \centering 
  \makebox[\textwidth][c]{\includegraphics[width=1\textwidth]{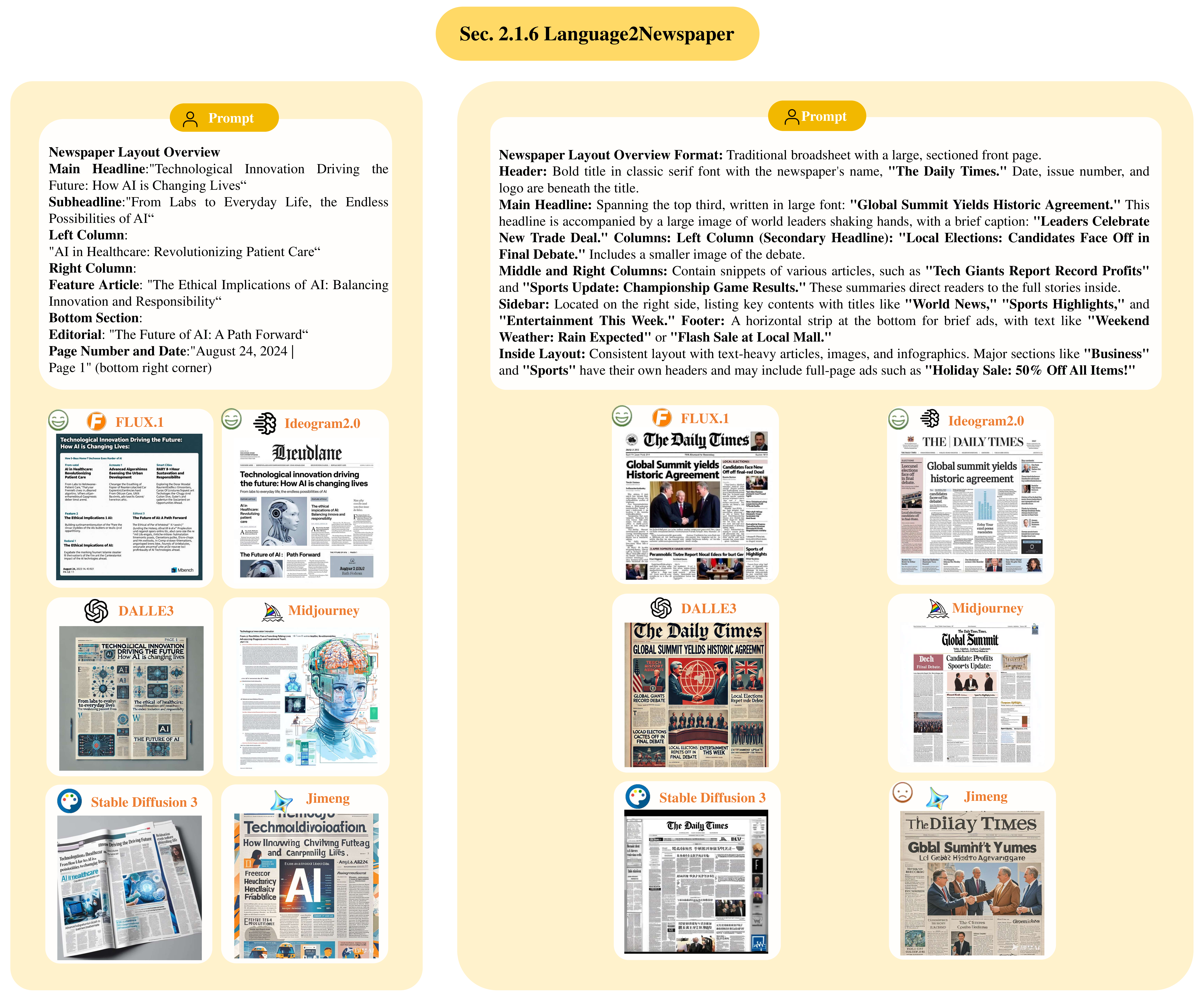}}
  \caption[Section~\ref{lan2news}: language2newspaper.]{Results on language2newspaper task. Refer to Section \ref{lan2news} for detailed discussions.}
  \label{lan2news_fig}
\end{figure*}

We evaluated the ability of these models to generate newspaper images based on natural language descriptions. We simply specified the layout and headlines for different sections of the newspaper and guided the models to generate a newspaper page, and the results are shown in the left subplot of Figure \ref{lan2news_fig}. Among the models tested, Ideogram2.0's results were significantly better than the others, successfully generating the corresponding layout and headlines in the specified positions and adhering to the artistic style of a newspaper. Dall-E3, Stable Diffusion 3, and Jimeng model were able to generate newspaper-style images, but their text generation had significant flaws. FLUX.1 produced mostly correct text and layout, but the style did not match that of a newspaper. Midjourney's generation was unsatisfactory in both newspaper style and textual content.

In in the right subplot of Figure \ref{lan2news_fig}, We present a more complex example. We describe in greater detail the titles, style, content, and the placement of inserted images for each section of the newspaper. Although none of the models perfectly met the requirements of the prompt, Ideogram2.0 still outperformed the others, correctly generating the required layout and main titles. FLUX.1 was able to generate some of the titles correctly, but the layout had errors. Dall-E3, Stable Diffusion 3, Jimeng, and Midjourney barely generated any correct text or layout.

\textbf{Score. }The scores of the model results in this task are shown in the Table \ref{tab_news}. CLIPScore aligns relatively well with human intuition, while the other three metrics show significant discrepancies from human judgments, possibly because scoring in this task requires examining the specific text content within the images.
\begin{table}[h]
    \centering
    \caption[Section~\ref{lan2news}: language2newspaper.]{The scoring of generation results by six models on language2newspaper under different evaluation systems. Refer to Section \ref{lan2news} for detailed discussions.}
    \begin{tabular}{l|c|c|c|c|c}
        \midrule
        Model & CLIPScore & HPSv2 & Aesthetic Score& GPT-4o & Human \\
        \midrule
        FLUX.1 & 30.27 & 0.19 & 4.45 &   3.34 & 8.34 \\ 
        Ideogram2.0 & \textbf{32.95} & 0.26 & 5.21 &   4.58 & \textbf{9.16} \\
        Dall-E3 & 29.57 & 0.27 & \textbf{5.29} &   3.33 & 7.50 \\
        Midjourney & 29.92 & 0.25 & 4.88 &   4.17 & 6.67 \\
        SD3 & 31.78 & \textbf{0.28} & 5.14 &   3.75 & 6.25 \\
        Jimeng & 30.88 & 0.24 & 5.27 &   \textbf{4.59} & 5.42 \\
        \midrule
    \end{tabular}
    \label{tab_news}
\end{table}

\subsubsection{Language2Paper}
\label{lan2paper}
\begin{figure*}[!ht]
  \centering 
  \makebox[\textwidth][c]{\includegraphics[width=0.85\textwidth]{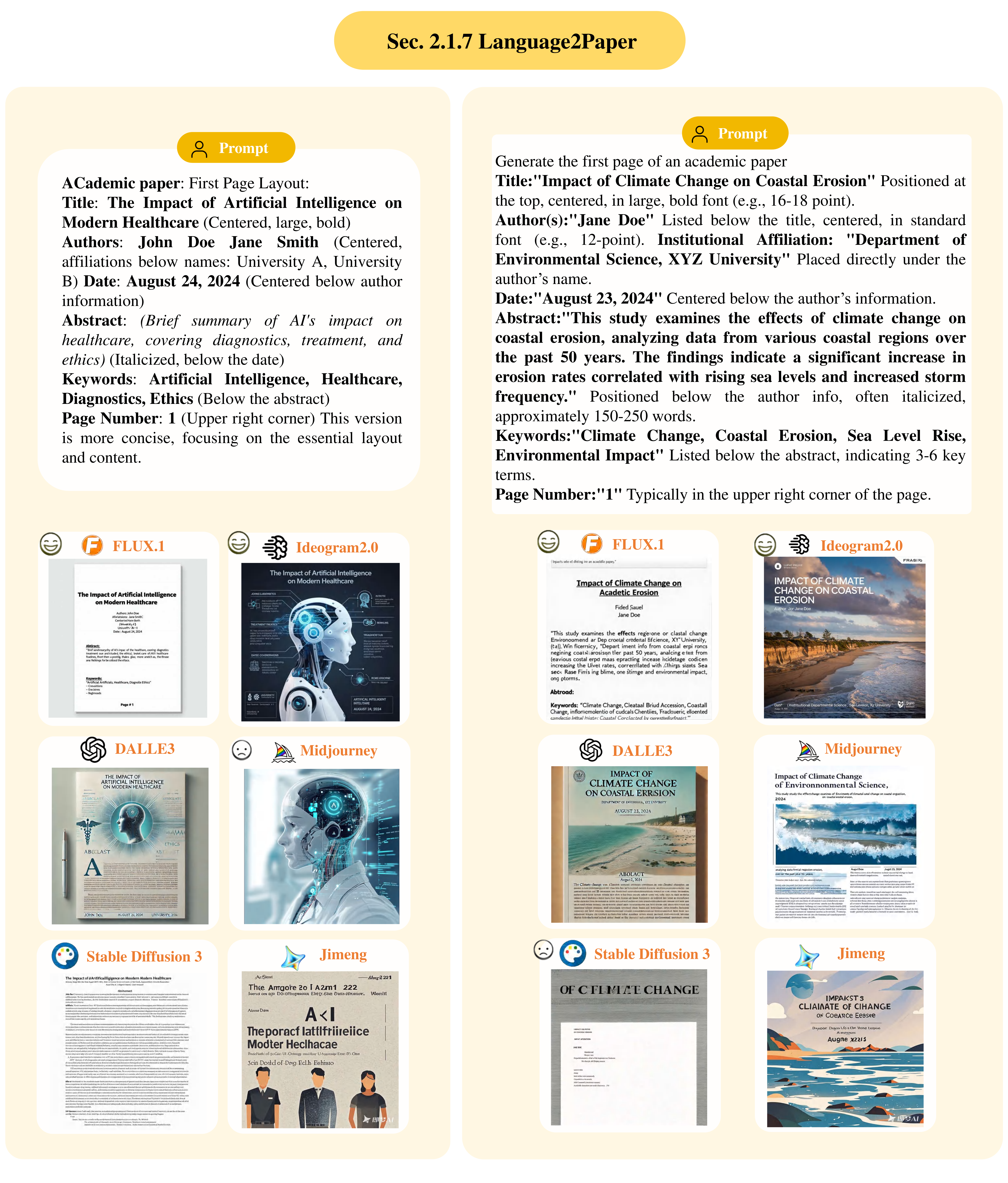}}
  \caption[Section~\ref{lan2paper}: language2paper.]{Results on language2paper task. Refer to Section \ref{lan2paper} for detailed discussions.}
  \label{fig_lan2paper}
\end{figure*}

In Figure \ref{fig_lan2paper}, we evaluated the ability of these models to generate academic paper images based on natural language descriptions. We specified the paper's title, author, abstract outline, and date to guide the models in generating the first page of an academic paper. Among the models tested, only FLUX.1 and Stable Diffusion 3 were able to correctly produce the layout of an academic paper, while the other models mistakenly generated a large number of decorative images. In terms of text accuracy, Ideogram2.0 and FLUX.1 performed the best, being able to accurately generate titles and subtitles. Dall-E3 followed closely, while Midjourney and Stable Diffusion 3 almost failed to generate correct text.

\textbf{Score. }The results of this task are shown in the Table \ref{tab_lan2paper}. We can observe that all metrics differ somewhat from human intuitive judgments. This discrepancy arises because accurate evaluation requires a thorough understanding of the basic format of academic papers and a detailed comparison of the specific text content in the images, leading to insufficient accuracy of these evaluation metrics.

\begin{table}[h]
    \centering
    \caption[Section~\ref{lan2paper}: language2paper.]{The scoring of generation results by six models on language2paper under different evaluation systems. Refer to Section \ref{lan2paper} for detailed discussions.}
    \begin{tabular}{l|c|c|c|c|c}
        \midrule
        Model & CLIPScore & HPSv2 & Aesthetic Score &GPT-4o & Human \\
        \midrule
        FLUX.1 & 26.67 & 0.20 & 4.12 & 2.92 & \textbf{9.59} \\
        Ideogram2.0 & \textbf{33.87} & 0.21 & 4.26 & 3.75 & 8.33 \\
        Dall-E3 & 29.06 & 0.21 & 4.51 & 3.75 & 7.50 \\
        Midjourney & 29.62 & \textbf{0.26} & 4.91 & \textbf{5.84} & 4.58 \\
        SD3 & 26.04 & 0.19 & \textbf{5.08} & 3.75 & 5.42 \\
        Jimeng & 31.87 & 0.22 & 4.29 & 5.00 & 3.34 \\
        \midrule
    \end{tabular}
    \label{tab_lan2paper}
\end{table}

\subsubsection{Json2Image}
\label{json}
\begin{figure*}[!ht]
  \centering 
  \makebox[\textwidth][c]{\includegraphics[width=1\textwidth]{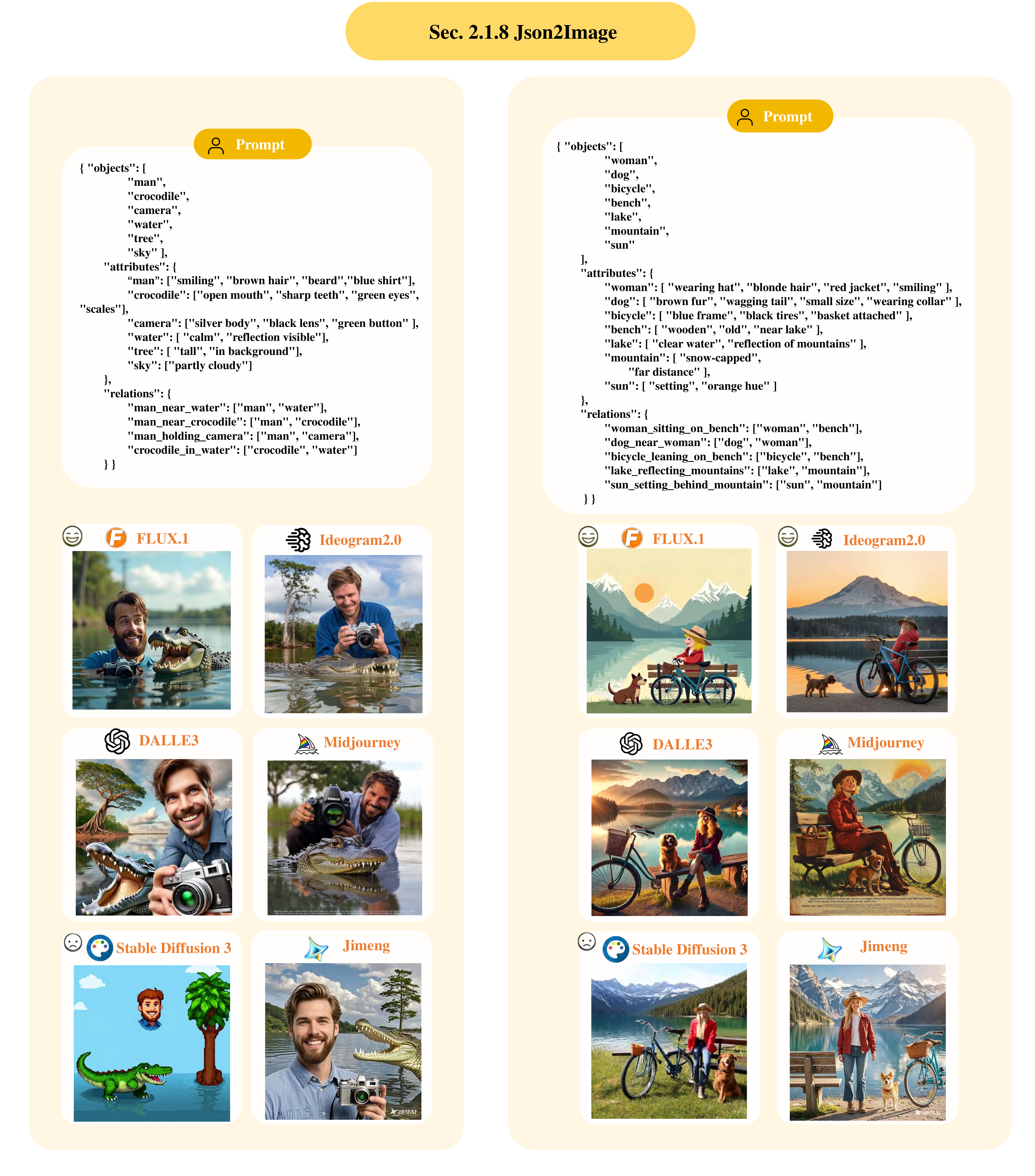}}
  \caption[Section~\ref{json}: json2image.]{Results on json2image task. Refer to Section \ref{json} for detailed discussions.}
  \label{fig_json}
\end{figure*}

We used a new prompt format to evaluate the ability of T2I models to understand the relationships between objects and generate images correctly. The prompt was designed in a JSON format, which is divided into three parts: objects, attributes, and relations. The objects section describes the items that appear in the image, the attributes section details the characteristics and specifics of each object, and the relations section describes the spatial or logical relationships between different items. An example is shown in Figure \ref{fig_json}. In the first example, we found that, except for Midjourney, which cannot process this format, both Jimeng and Stable Diffusion 3 could only understand the main objects and combine them together, lacking logical coherence. Dall-E3 generated a green lens, while FLUX.1 and Ideogram2.0 performed the best. In the second example, except for Midjourney, the output from Jimeng failed to show the woman sitting down. In Stable Diffusion 3 and Dall-E3’s results, the bicycle’s tire was incomplete. FLUX.1 and Ideogram2.0 excelled in this task as well.

\textbf{Score. }The results of this task are shown in the Table \ref{tab_json}, where we found that the evaluations from CLIPScore and GPT-4o are closer to human intuition, while Ideogram2.0 performs better in these two metrics.
\begin{table}[h]
    \centering
    \caption[Section~\ref{json}: Json2Image.]{The scoring of generation results by six models on json2image under different evaluation systems. Refer to Section \ref{json} for detailed discussions.}
    \begin{tabular}{l|c|c|c|c|c}
        \midrule
        Model & CLIPScore & HPSv2 & Aesthetic Score & GPT-4o & Human \\
        \midrule
        FLUX.1 & 18.96 & 0.22 & 6.44 &   5.84 & \textbf{10.00} \\
        Ideogram2.0 & \textbf{20.14} & 0.22 & 5.66 &   \textbf{7.50} & 9.16 \\
        Dall-E3 & 16.39 & 0.24 & 6.11 &   5.84 & 7.50 \\
        Midjourney & 14.00 & 0.18 & 5.85 &   7.08 & 9.16 \\
        SD3 & 18.55 & \textbf{0.26} & \textbf{6.75} &   4.16 & 6.25 \\
        Jimeng & 19.40 & \textbf{0.26} & 6.40 &   6.25 & 8.75 \\
        \midrule
    \end{tabular}
    \label{tab_json}
\end{table}

\subsubsection{UI Design}
\label{ui}
\begin{figure*}[!ht]
  \centering 
  \makebox[\textwidth][c]{\includegraphics[width=0.85\textwidth]{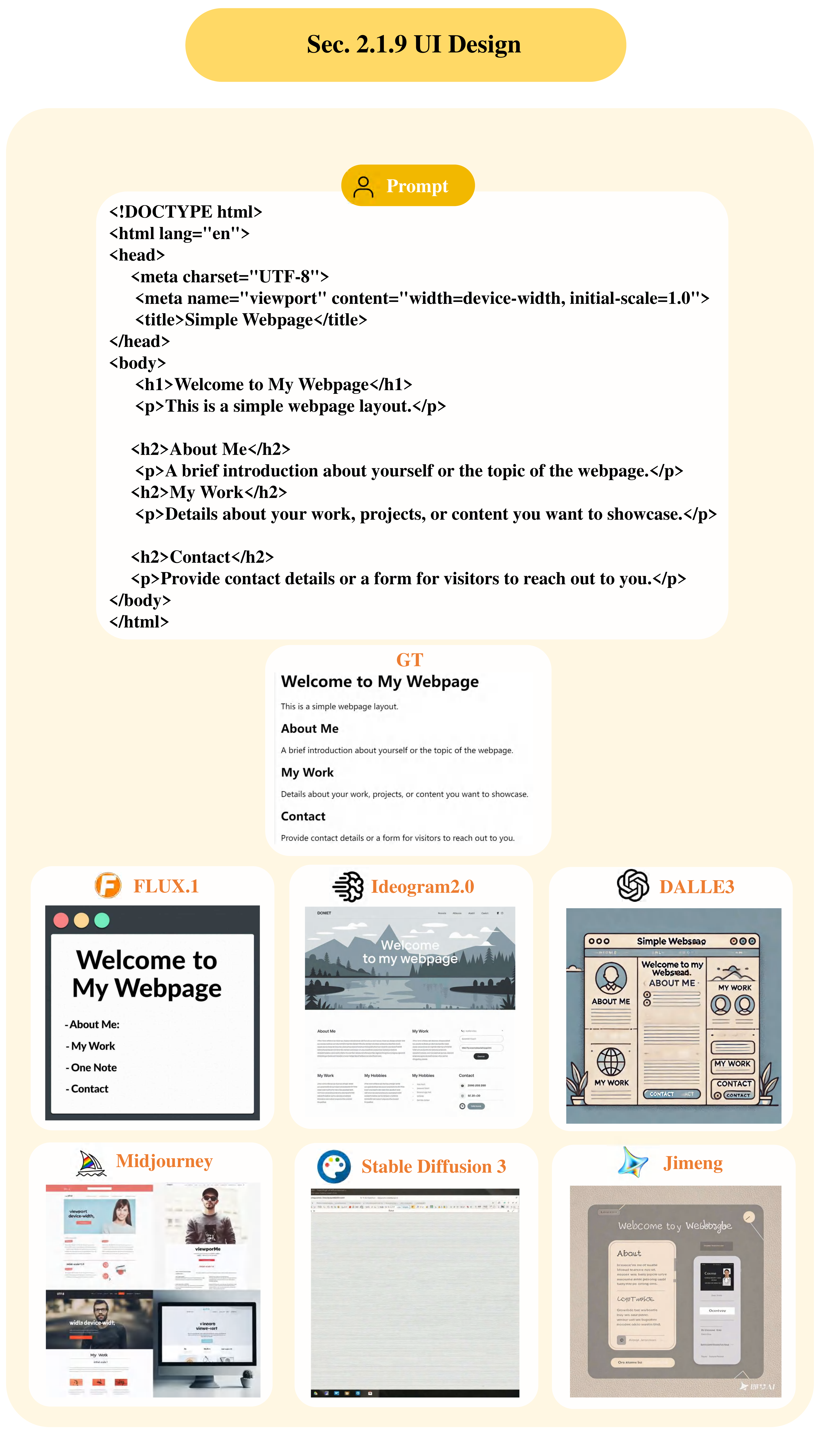}}
  \caption[Section~\ref{ui}: ui design.]{Results on UI design. Refer to Section \ref{ui} for detailed discussions.}
  \label{fig_ui-1}
\end{figure*}

Previous work has explored how to use AI as an assistive tool for UI design~\cite{Wei2024OnAU,Hui2023UnifyingLG,Cheng2024CoLayCL,Wei2023BoostingGP}.In this work, we explore the potential of using text-to-image models for automating UI design.
\textbf{Code2UI. }   
UI design is a common task for evaluating T2I models' ability to follow instructions. We input HTML code into the models, and the results are shown in Figure \ref{fig_ui-1}. While all models generate some form of a web interface, Stable Diffusion 3 produces output that appears as meaningless gibberish. In comparison to the ground truth, only FLUX.1, Ideogram2.0, and Dall-E3 follow the instructions more accurately, generating web layouts containing the sections "About Me", "My Work", and "Contact."

\begin{figure*}[!ht]
  \centering 
  \makebox[\textwidth][c]{\includegraphics[width=0.9\textwidth]{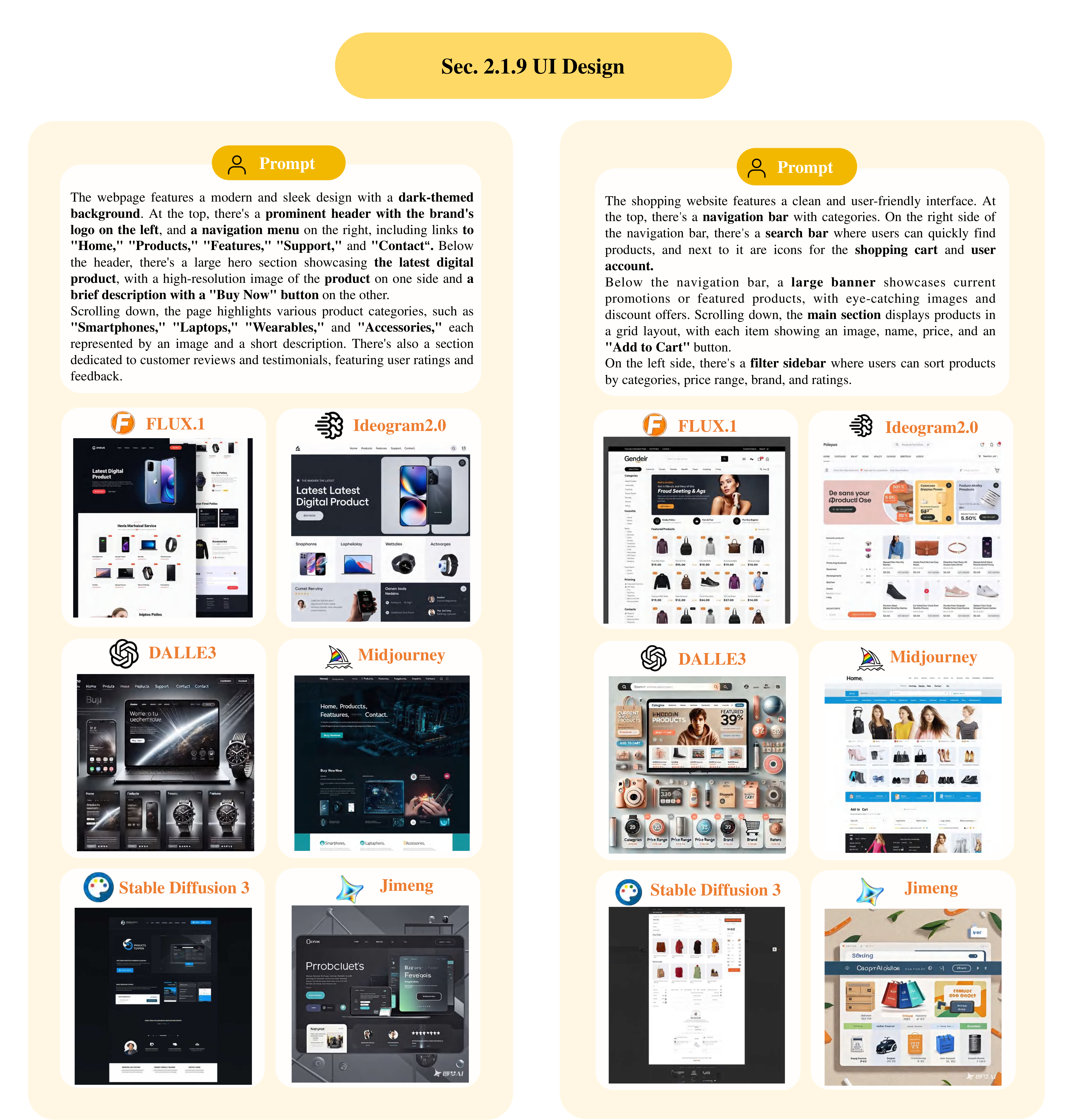}}
  \caption[Section~\ref{ui}: ui design.]{Results on UI design. Refer to Section \ref{ui} for detailed discussions.}
  \label{fig_ui-2}
\end{figure*}

\textbf{Language2UI. }
In this task, we assess models' ability to convert natural language into web interfaces, as shown in Figure \ref{fig_ui-2}. In the first example, both Jimeng and Stable Diffusion 3 produce unreadable text, while Dall-E 3 fails to generate a typical web interface. In contrast, FLUX.1, Ideogram2.0, and Midjourney generate legible text, with FLUX.1 and Ideogram2.0 excelling in instruction-following. In the second example, Jimeng, Stable Diffusion 3, and Midjourney produce blurry outputs, while Ideogram2.0 and Dall-E3 contain some chaotic text. FLUX.1 outperforms the other models, demonstrating better instruction-following.

\textbf{Score. }The results of this experiment are shown in the Table \ref{tab_ui}. Among the scores for CLIPScore, HPSv2, and Aesthetic score, FLUX.1 achieved a higher score. In human intuitive perception, the outputs of FLUX.1 and Ideogram2.0 are also better, while the results of GPT-4o are inconsistent with human intuitive perceptions in this experiment.

\begin{table}[h]
    \centering
    \caption{The scoring of generation results by six models on UI design task under different evaluation systems. Refer to Section \ref{ui} for detailed discussions.}
    \begin{tabular}{l|c|c|c|c|c}
        \midrule
        Model & CLIPScore & HPSv2 & Aesthetic Score &  GPT-4o & Human \\
        \midrule
        FLUX.1 & \textbf{28.19} & \textbf{0.22} & \textbf{4.94} &   4.45 & 8.89 \\
        Ideogram2.0 & 25.82 & 0.20 & 4.64 &   5.00 & \textbf{9.44} \\
        Dall-E3 & 27.32 & 0.21 & 4.77 &   4.72 & 6.67 \\
        Midjourney & 24.68 & 0.18 & 4.56 &   \textbf{5.28} & 6.94 \\
        SD3 & 27.98 & 0.19 & 4.88 &   \textbf{5.28} & 5.00 \\
        Jimeng & 26.27 & 0.21 & 4.77 &   4.72 & 6.67 \\
        \midrule
    \end{tabular}
    \label{tab_ui}
\end{table}

\subsubsection{Code Generation}
\label{code}
\begin{figure*}[!ht]
\vspace{-2em}
  \centering 
  \makebox[\textwidth][c]{\includegraphics[width=0.8\textwidth]{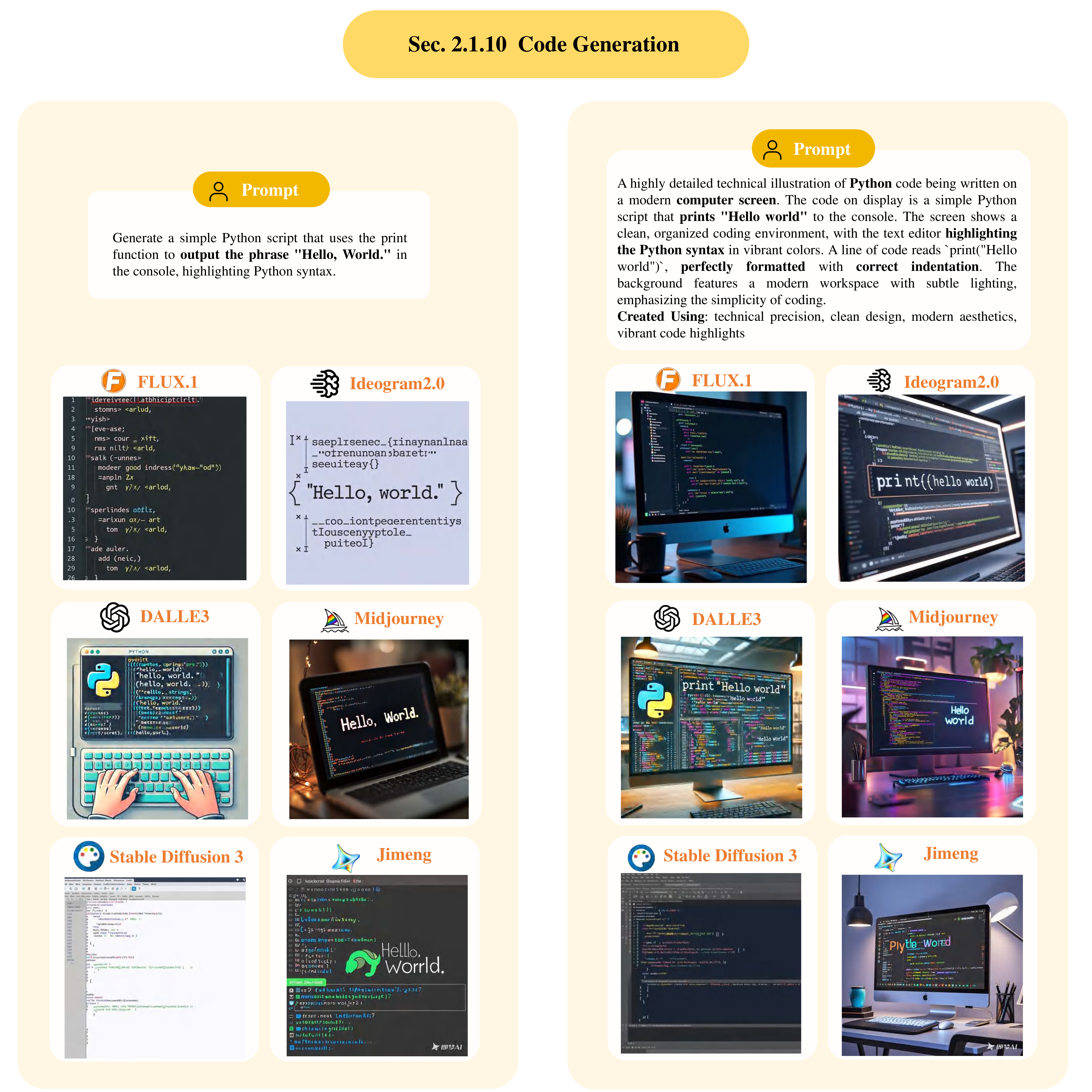}}
  \caption[Section~\ref{code}: code generation.]{Results on code generation. Refer to Section \ref{code} for detailed discussions.}
  \label{code1}
\end{figure*}

\begin{figure*}[!ht]
  \centering 
  \makebox[\textwidth][c]{\includegraphics[width=0.8\textwidth]{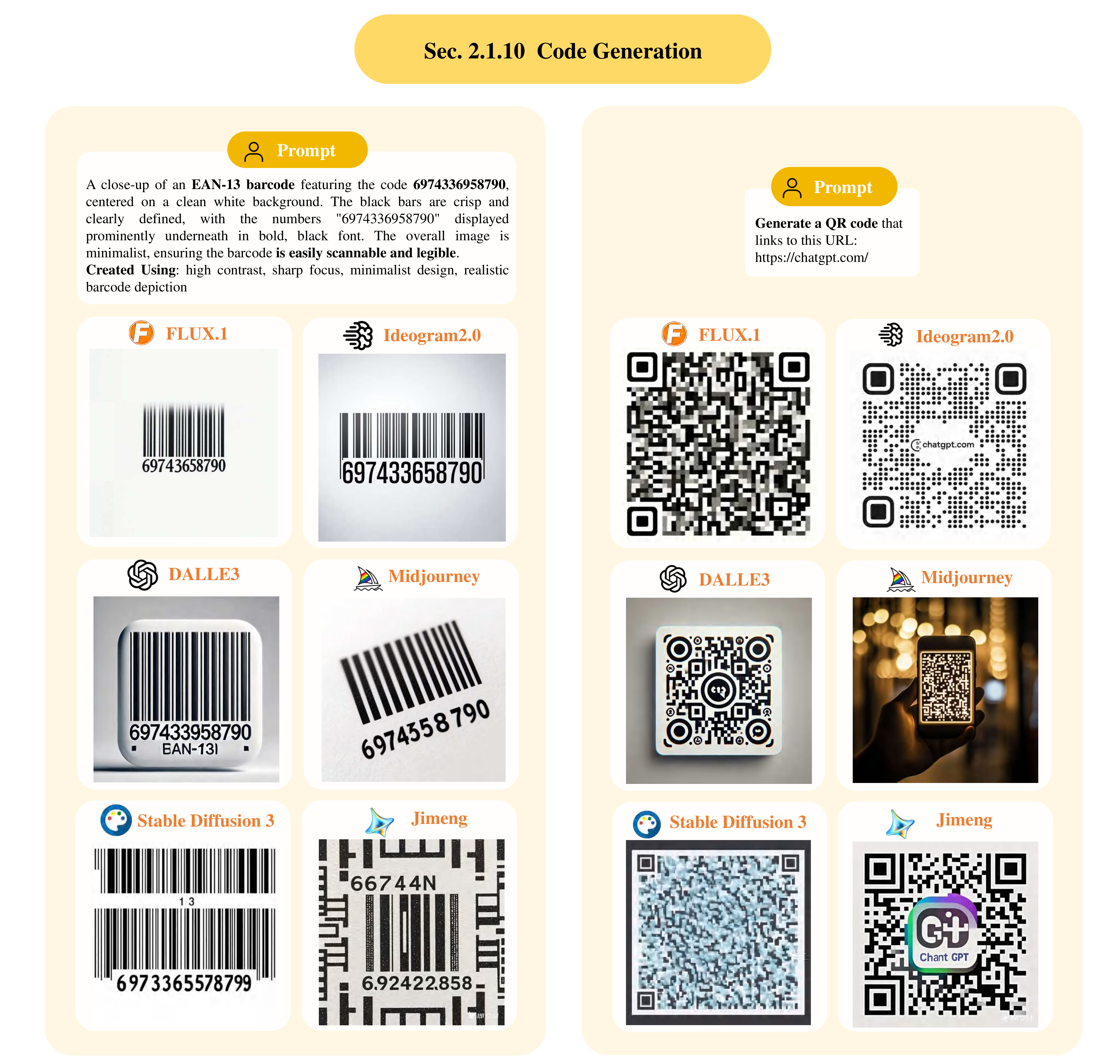}}
  \caption[Section~\ref{code}: code generation.]{Results on code generation. Refer to Section \ref{code} for detailed discussions.}
  \label{code2}
\end{figure*}

The use of LLMs for code generation has long been a focus of research~\cite{Chen2021EvaluatingLL,Feng2020CodeBERTAP,Ahmad2021UnifiedPF,Wang2021CodeT5IU}. With the emergence of diffusion models, the question arises: Can text-to-image models also be used to generate code? In Figures \ref{code1} and \ref{code2}, we examine the models' capability to generate various types of code, including Python and C programs, as well as barcodes and QR codes, in order to explore the potential for generalizing text-to-image (T2I) models into more fundamental models. In Figure \ref{code1}, the models are expected to generate images containing correct program code. However, none of the models produce accurate outputs. Instead, they generate images depicting computer screens with code-like visuals. Similarly, in Figure \ref{code2}, where the tasks are to generate valid barcodes and QR codes, all models fail to produce correct results. These findings suggest that significant further development is required before t2i models can evolve into foundational models capable of handling such tasks.

\textbf{Score. }
The experimental results of this task are shown in Table \ref{tab_code}. It can be observed that due to the difficulty of the task, the performance of several models is not high.
\begin{table}[h]
    \centering
    \caption[Section~\ref{code}: code generation.]{The scoring of generation results by six models on code generation under different evaluation systems. Refer to Section \ref{code} for detailed discussions.}
    \begin{tabular}{l|c|c|c|c|c}
        \midrule
        Model & CLIPScore & HPSv2 & Aesthetic Score & GPT-4o & Human \\
        \midrule 
        FLUX.1      & 28.10 & 0.22 & 4.50 &  4.38 & \textbf{5.00} \\
        Ideogram2.0 & 27.41 & 0.25 & 4.99 &  4.58 & 4.38 \\
        Dall-E3     & 25.70 & 0.23 & 4.44 &  5.42 & 4.17 \\
        Midjourney  & 25.84 & 0.24 & 5.00 &  \textbf{5.63} & 4.17 \\
        SD3         & 26.49 & 0.24 & \textbf{5.02} &  4.59 & 2.50 \\
        Jimeng      & \textbf{30.53} & \textbf{0.26} & 4.89 &  4.79 & 2.50 \\
        \midrule
    \end{tabular}
    \label{tab_code}
\end{table}

\clearpage
\subsection{Realism and Physical Consistency Tasks}
\label{sec:03realism}
In the image and video generation task, achieving realism and physical consistency is crucial. Previous works have made significant contributions to this area of research~\cite{Motamed2025DoGV,Li2024SimAvatarSA,Liu2024ANIDHF,Huang2023TeCHTR}. We aim for models to generate images that are not only visually compelling but also believable and grounded in the physical world. To assess a model's ability to understand and replicate real-world dynamics, we have designed a set of tasks that evaluate its grasp of physical laws. 

In Section \ref{Multi-Person}, we evaluate the models' ability to generate credible human figures in complex multi-person settings. Section \ref{body} focuses on assessing the models' capability to accurately render human bodies and poses. In Section \ref{photo}, we incorporate various photographic terminologies into the prompts to test the models' understanding of photography techniques. Section \ref{perspective} examines the models' ability to interpret and generate correct perspective relationships within realistic scenes. Section \ref{physical} explores the extent to which T2I models understand the fundamental physical laws of the real world.

\subsubsection{Multi-Person}
\label{Multi-Person}
\begin{figure*}[!ht]
  \centering 
  \makebox[\textwidth][c]{\includegraphics[width=0.75\textwidth]{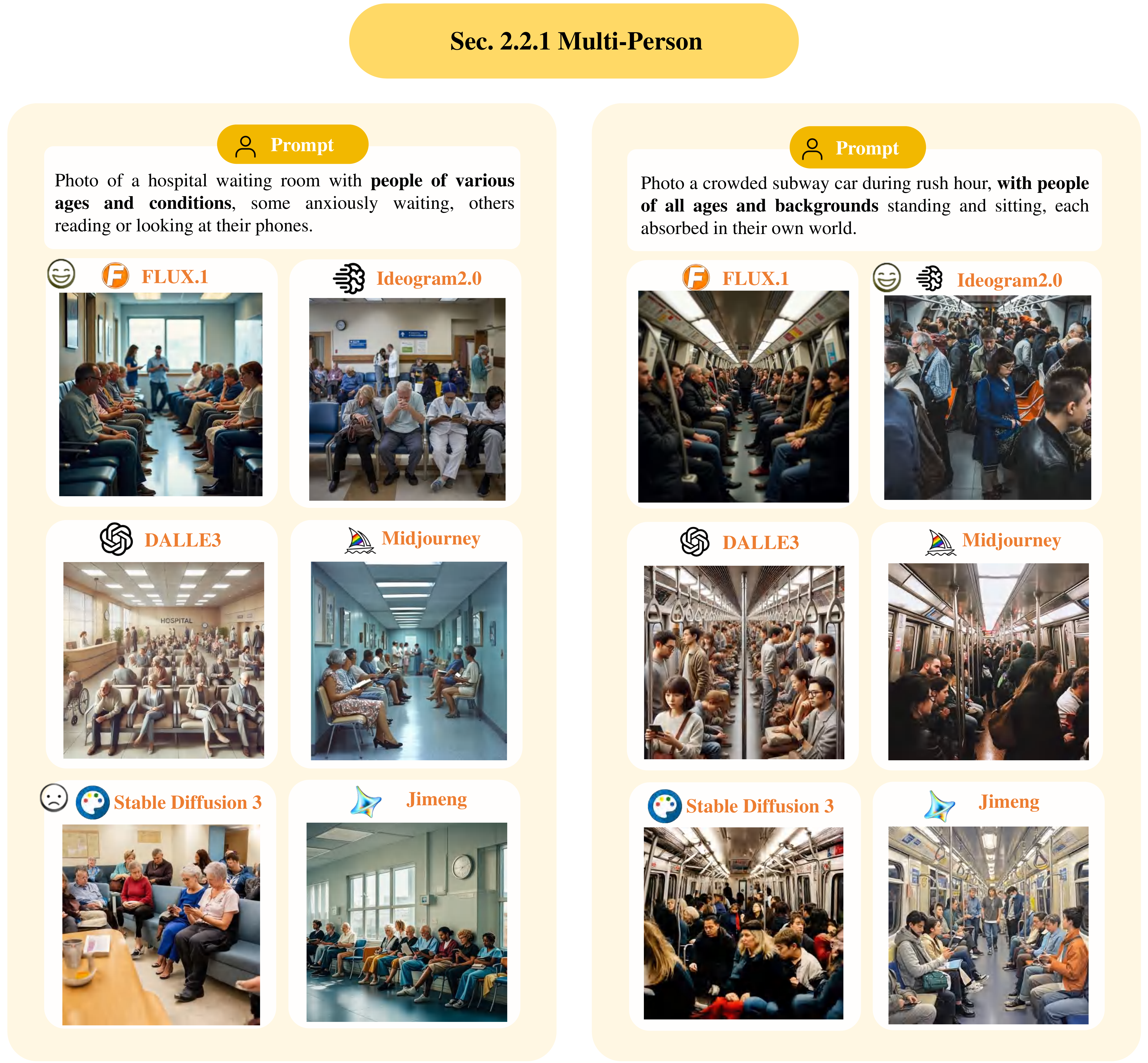}}
  \caption[Section~\ref{body}: multi-person.]{Results on multi-person task. Refer to Section \ref{Multi-Person} for detailed discussions.}
  \label{mutliperson}
\end{figure*}

Generating images with multiple characters has always been a highly challenging task~\cite{Zhang2024FollowYourMultiPoseTM}. Figure~\ref{mutliperson} depicts the visualization results of six models in generating images based on prompts involving multiple persons. FLUX.1 demonstrates a strong ability to capture overall details from the prompts. And Stable Diffusion 3~\cite{rombach2022high}, Midjourney, and Jimeng struggle with handling the overlapping and non-overlapping aspects of multiple persons. Midjourney often cuts off half of a face, and Jimeng produces disjointed upper body parts. In the second part of Figure~\ref{mutliperson}, Jimeng and FLUX.1 successfully generate images of multiple persons on a crowded subway with minimal distortion. FLUX.1, in particular, handles facial features and overlapping boundaries well, though its color palette is somewhat monotonous, and the depicted actions are limited. Conversely, Stable Diffusion 3 introduces significant distortions, notably a visible distortion at the junction of a blonde woman's hair and another man's face. Midjourney also exhibits distortion, particularly in the distant background of the subway scene.

\textbf{Score. }
The results of this experiment are shown in Table \ref{tab_multiperson}. Both CLIPScore, HPSv2, and GPT-4o consider Midjourney's output to be superior; however, upon our careful observation, the human forms in FLUX.1 appear more realistic.
\begin{table}[h]
    \centering
    \caption[Section~\ref{Multi-Person}: multi-person.]{The scoring of generation results by six models on multi-person under different evaluation systems. Refer to Section \ref{Multi-Person} for detailed discussions.}
    \begin{tabular}{l|c|c|c|c|c}
        \midrule
        Model & CLIPScore & HPSv2 & Aesthetic Score &  GPT-4o & Human \\
        \midrule
        FLUX.1      & 26.31 & 0.29 & 5.69 &  5.00 & \textbf{9.17} \\
        Ideogram2.0 & 25.60 & \textbf{0.31} & \textbf{6.04} &  \textbf{7.08} & 5.00 \\
        Dall-E3     & 25.05 & 0.29 & 5.51 &  7.01 & 7.08 \\
        Midjourney  & \textbf{27.64} & \textbf{0.31} & 5.36 &  \textbf{7.08} & 7.92 \\
        SD3         & 23.60 & 0.30 & 5.98 &  6.25 & 7.08 \\
        Jimeng      & 26.87 & 0.30 & 5.03 &  6.67 & 5.84 \\
        \midrule
    \end{tabular}
    \label{tab_multiperson}
\end{table}

\subsubsection{Human body}
\label{body}
\begin{figure*}[!ht]
\vspace{-3em}
  \centering 
  \makebox[\textwidth][c]{\includegraphics[width=0.75\textwidth]{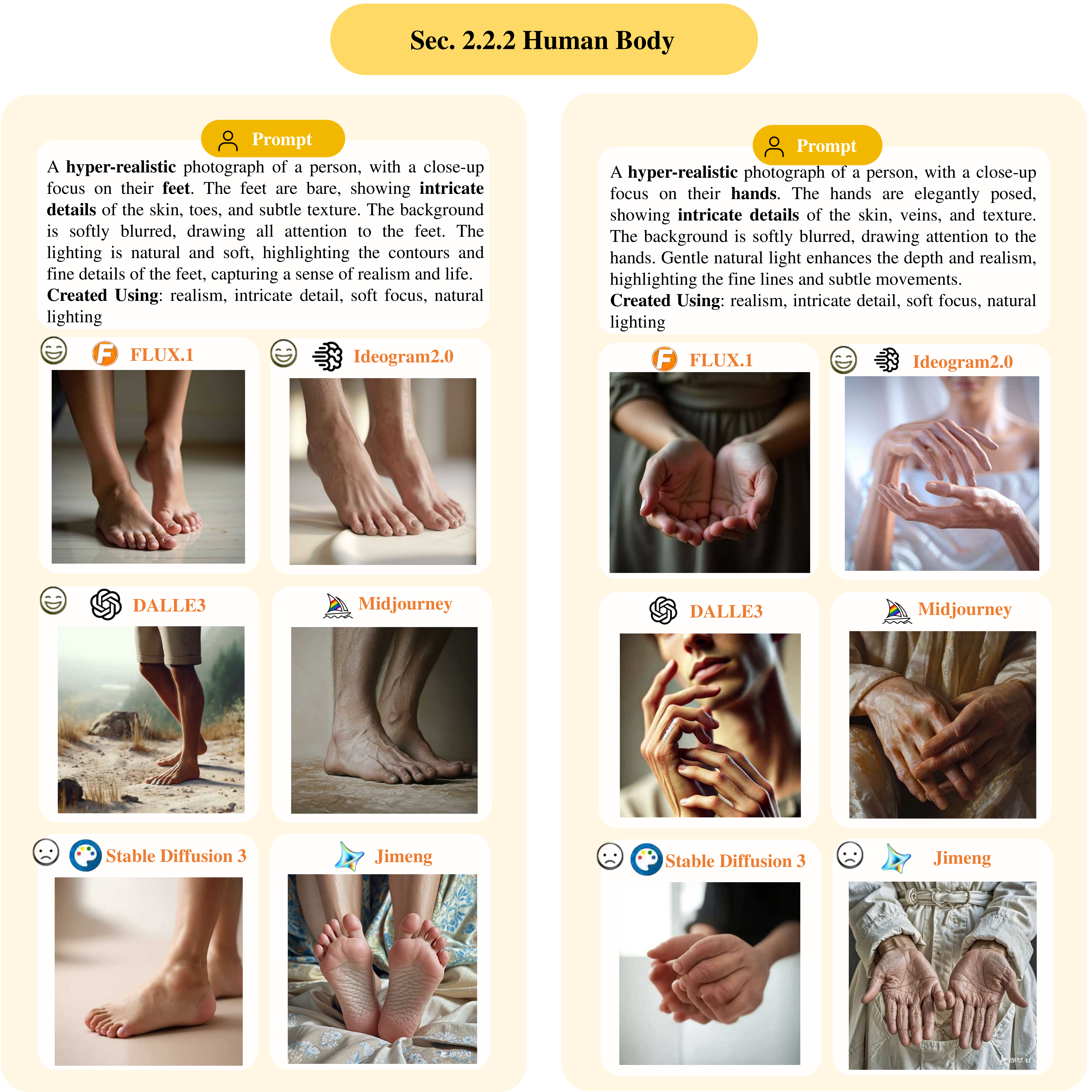}}
  \caption[Section~\ref{body}: human body]{Results on human body task. Refer to Section \ref{body} for detailed discussions.}
  \label{human_body-1}
\end{figure*}

\begin{figure*}[!ht]
  \centering 
  \makebox[\textwidth][c]{\includegraphics[width=0.75\textwidth]{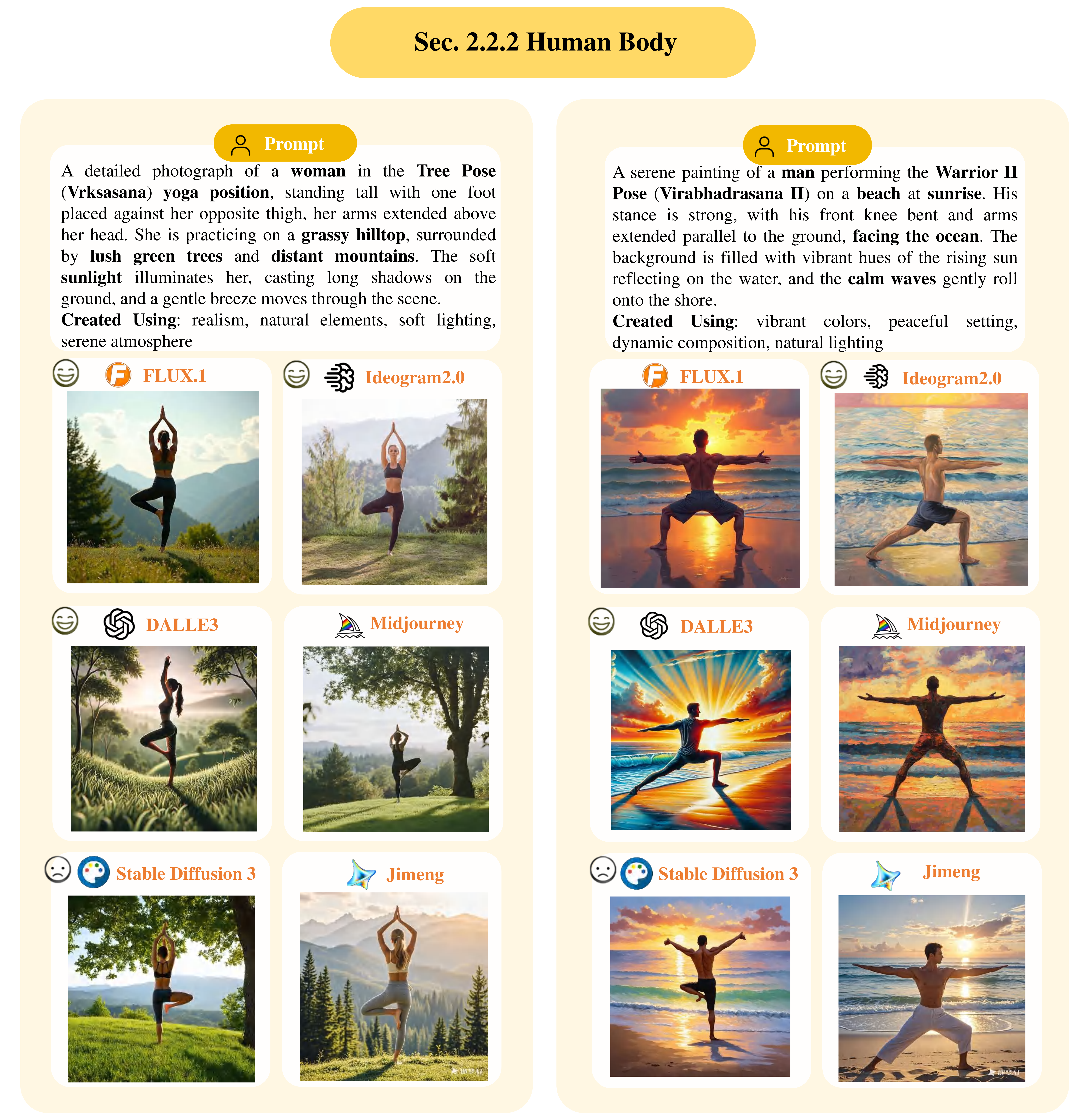}}
  \caption[Section~\ref{body}: human body]{Results on human body task. Refer to Section \ref{body} for detailed discussions.}
  \label{human_body-2}
\end{figure*}

In Figure \ref{human_body-1}-\ref{human_body-2}, we examine the models' ability to accurately generate human body, with a particular focus on the hands and feet, which are difficult tasks in image synthesis. 

\textbf{Hands.} For instance, in the left subplot of Figure \ref{human_body-1}, all the models produce extra fingers except Dall-E3~\cite{betker2023improving}. Specifically, the Stable Diffusion 3 and Jimeng models exhibit entirely irrational hand structures. The image generated by Ideogram2.0 looks fake. Midjourney demonstrates a capacity to capture significant hand details, and FLUX.1~\cite{flux2024} achieves the most accurate body structure. 

\textbf{Feet.} In the right subplot of Figure \ref{human_body-1}, both Midjourney and Stable Diffusion 3 generate the totally wrong foot structures, and Dall-E3 even result illegal, whereas FLUX.1, Jimeng and Ideogram2.0 produce more anatomically correct feet, despite Jimeng and Ideogram2.0 displaying oddly legs. Overall, FLUX.1 exhibits superior human body structure generation compared to the other models, though it still requires improvements in rendering the correct number of fingers.

\textbf{Pose.} In Figure \ref{human_body-2}, we examine the models' ability to generate accurate human poses. In the first example, the desired pose is the Tree Pose (Vrksasana) from yoga. FLUX.1, Ideogram2.0, Stable Diffusion 3, and Jimeng successfully generate a woman in the correct pose. However, the poses generated by Dall-E3 and Midjourney are inaccurate, possibly due to a lack of understanding of Vrksasana. While their outputs fit the general prompt description, even they do not accurately capture the specific yoga position. In the second example, only Jimeng precisely follows the Warrior II Pose, but it overlooks the prompt detail of "facing the ocean". Ideogram2.0 and Dall-E3 fail to depict the correct yoga pose but align more closely with the general description of the pose.

\textbf{Score. }All scoring results for this task are shown in Table \ref{tab_humanbody}. The output of FLUX.1 is closer to reality, GPT-4o's evaluation aligns with human perception, while CLIPScore, HPSv2, and Aesthetic Score differ significantly from human intuition.
\begin{table}[h]
    \centering
    \caption[Section~\ref{body}: human body.]{The scoring of generation results by six models on human body under different evaluation systems. Refer to Section \ref{body} for detailed discussions.}
    \begin{tabular}{l|c|c|c|c|c}
        \midrule
        Model & CLIPScore & HPSv2 & Aesthetic Score & GPT-4o & Human \\
        \midrule
        FLUX.1      & 27.16 & 0.27 & 5.70 &  \textbf{7.91} & \textbf{8.12} \\
        Ideogram2.0 & \textbf{30.86} & 0.26 & 5.92 & 6.66 & 6.25 \\
        Dall-E3     & 28.42 & \textbf{0.28} & 5.77 &  8.75 & 7.08 \\
        Midjourney  & 30.26 & 0.26 & 5.66 & 7.07 & 5.42 \\
        SD3         & 30.08 & 0.27 & \textbf{6.23} &  4.79 & 6.46 \\
        Jimeng      & 28.77 & 0.27 & 5.76 &  6.04 & 6.46 \\
        \midrule
    \end{tabular}
    \label{tab_humanbody}
\end{table}

\subsubsection{Photographic Image Generation}
\label{photo}
\begin{figure*}[!ht]
\vspace{1em}
  \centering 
  \makebox[\textwidth][c]{\includegraphics[width=1.1\textwidth]{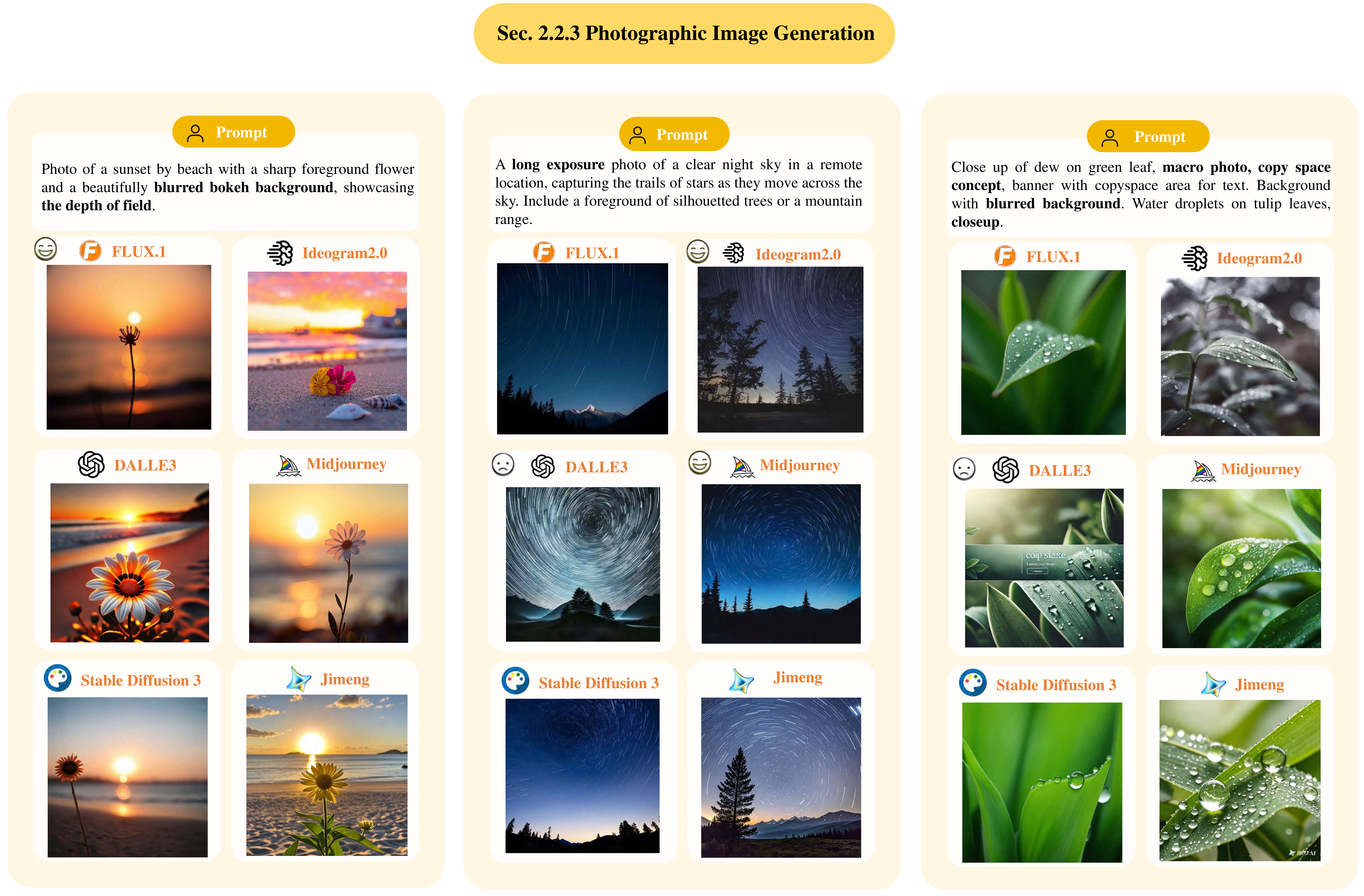}}
  \caption[Section~\ref{photo}: Photographic image generation.]{Results on photographic image generation. Refer to Section \ref{photo} for detailed discussions.}
  \label{fig_photo_1}
\end{figure*}

\begin{figure*}[!ht]
\vspace{1em}
  \centering 
  \makebox[\textwidth][c]{\includegraphics[width=1\textwidth]{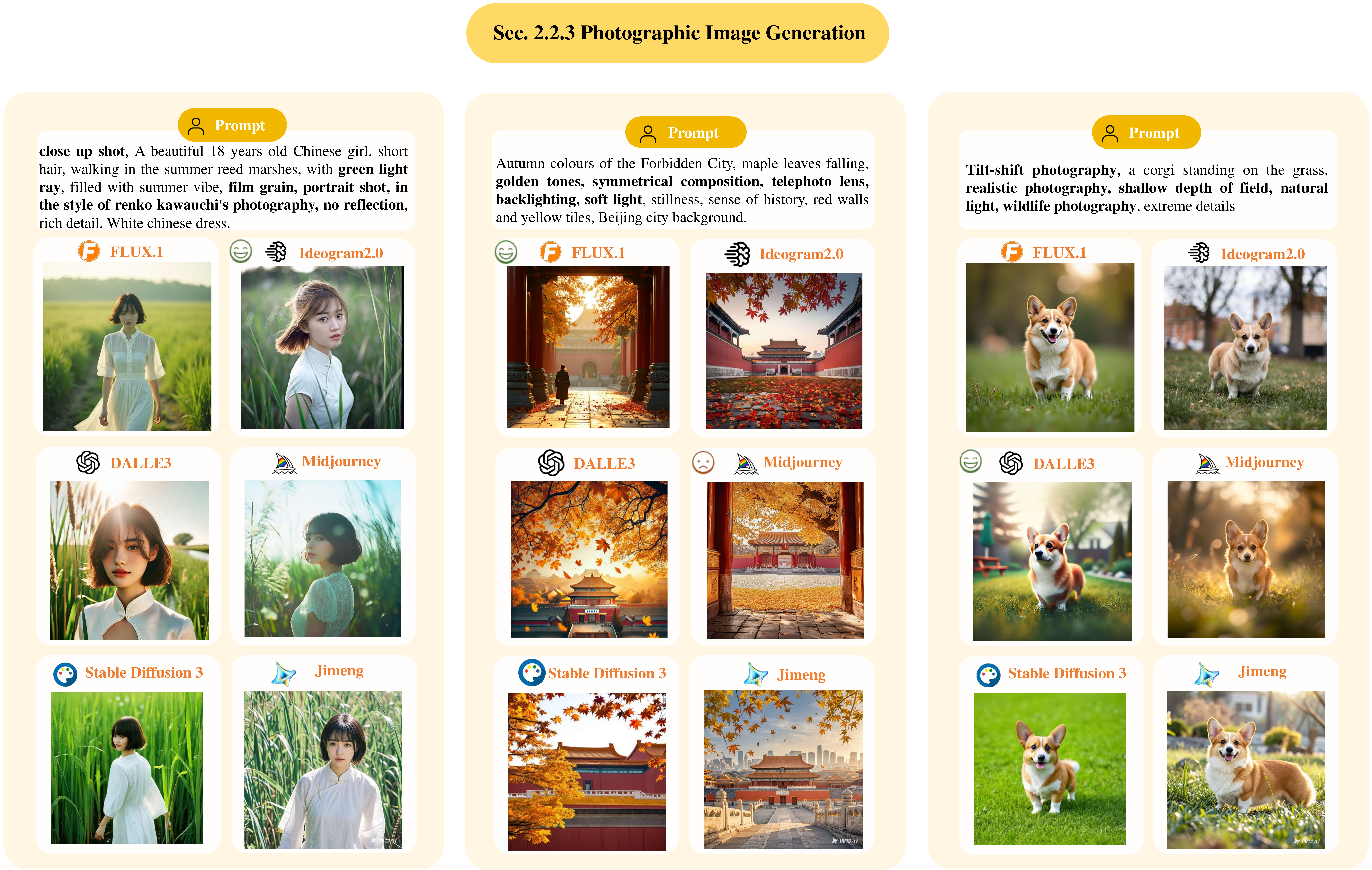}}
  \caption[Section~\ref{photo}: Photographic image generation.]{Results on photographic image generation. Refer to Section \ref{photo} for detailed discussions.}
  \label{fig_photo_2}
\end{figure*}

\begin{figure*}[!ht]
\vspace{1em}
  \centering 
  \makebox[\textwidth][c]{\includegraphics[width=1\textwidth]{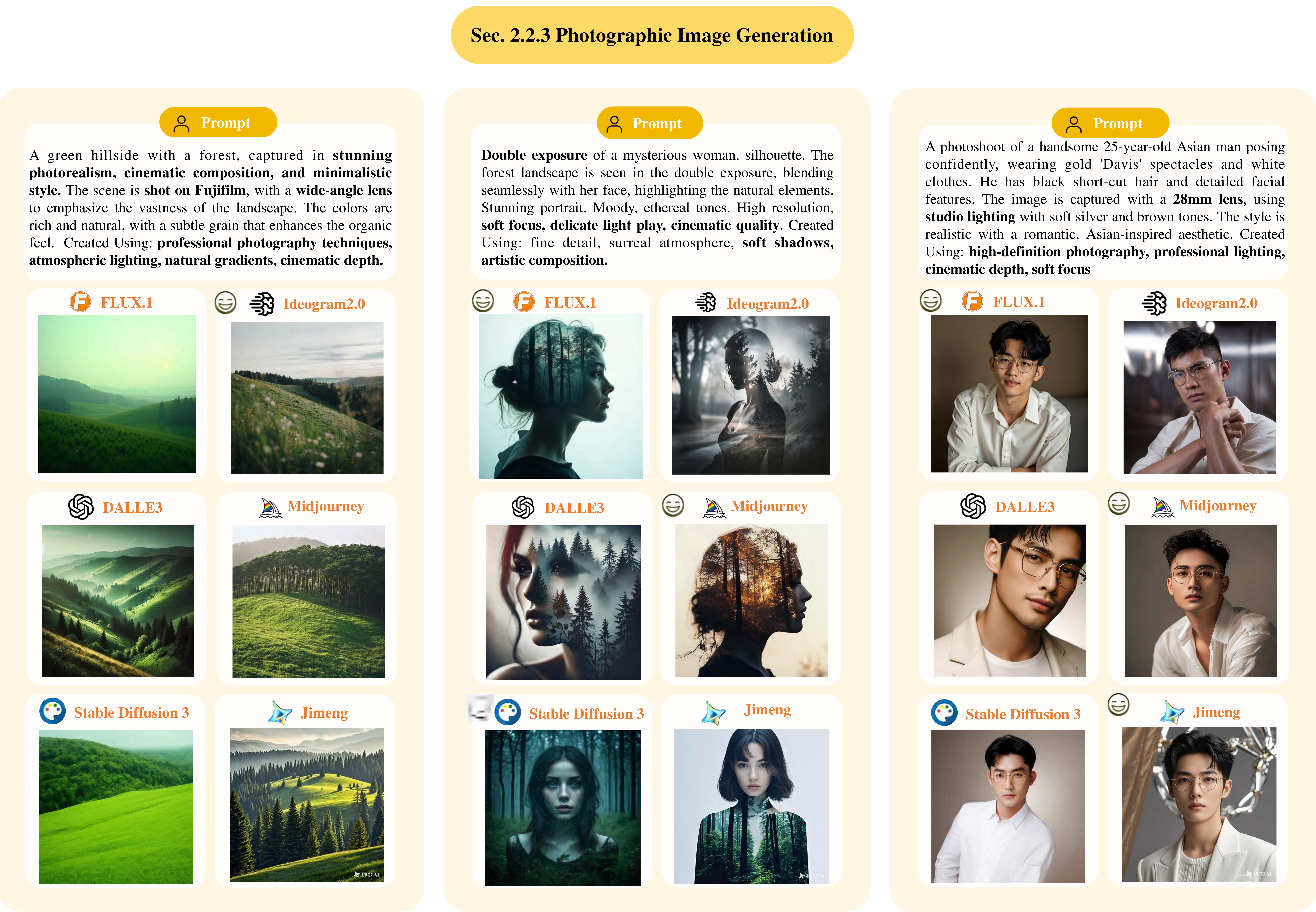}}
  \caption[Section~\ref{photo}: Photographic image generation.]{Results on photographic image generation. Refer to Section \ref{photo} for detailed discussions.}
  \label{fig_photo_3}
\end{figure*}

\begin{figure*}[!ht]
\vspace{1em}
  \centering 
  \makebox[\textwidth][c]{\includegraphics[width=1\textwidth]{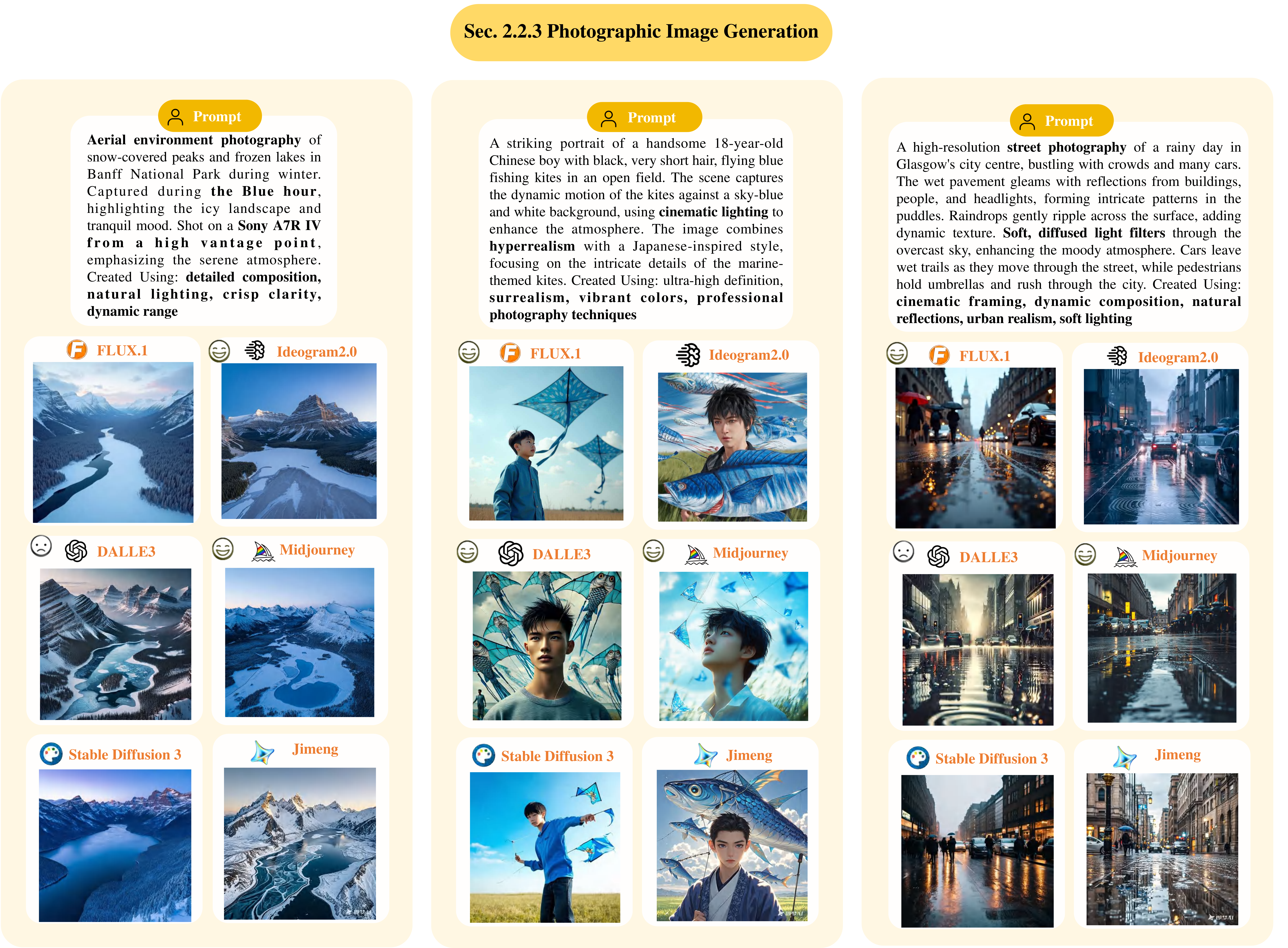}}
  \caption[Section~\ref{photo}: Photographic image generation.]{Results on photographic image generation. Refer to Section \ref{photo} for detailed discussions.}
  \label{fig_photo_4}
\end{figure*}

In Figure\ref{fig_photo_1}-\ref{fig_photo_4}, we explore the model's ability to generate images that meet specific requirements based on photographic terminology.

\textbf{Setting 1:} We tested blurred bokeh backgrounds and depth of field. The models generally understood these concepts, with FLUX.1, Midjourney, and Jimeng performing best. Results are shown in the left subplot of Figure \ref{fig_photo_1}.

\textbf{Setting 2:} We tested long exposure, specifically time-lapse photography represented as star trails. Dall-E3's images had excessive star trails that appeared unnatural, followed by Ideogram2.0 and Midjourney. FLUX.1 and Stable Diffusion 3 produced the best overall results. Results are shown in the middle subplot of Figure \ref{fig_photo_1}.

\textbf{Setting 3:} We examined macro photography and the concept of copy space. All models managed macro photography. Copy space refers to large blank areas in images for adding text, graphics, or other design elements. Dall-E3 mistakenly added unspecified text directly. Results are shown in the right subplot of Figure \ref{fig_photo_1}.

\textbf{Setting 4:} The setting involved terms like close-up shots, which refer to capturing above the chest. FLUX.1 and Stable Diffusion 3 missed this detail, while Midjourney performed best in overall style. Ideogram2.0's images were darker, and Dall-E3's output conflicted with photographic styles. Results are shown in the left subplot of Figure \ref{fig_photo_2}.

\textbf{Setting 5:} Tilt-shift photography, used to alter the focus and depth of field, typically for creating miniature scenes, was tested. Dall-E3 performed best with this keyword, and all models could accurately generate images as prompted. Results are shown in the middle subplot of Figure \ref{fig_photo_2}.

\textbf{Setting 6:} For golden tones, FLUX.1 and Dall-E3 excelled, while other models failed to achieve the effect. For symmetrical composition, only Stable Diffusion 3 missed the mark. For telephoto lens, backlighting, and soft light, FLUX.1 failed to deliver the telephoto effect but had the best lighting. Dall-E3's lighting was decent, while others only achieved soft light. Results are shown in the right subplot of Figure \ref{fig_photo_2}.

\textbf{Setting 7:} We tested stunning photorealism, cinematic composition, and minimalist style. Midjourney and Ideogram2.0 had the most realistic images, followed by FLUX.1. Minimalist style was harder to judge, but FLUX.1 and Stable Diffusion 3 had the fewest elements. For the 'shot on Fujifilm' look, only Jimeng and Dall-E3 struggled to achieve the retro film style with subtle contrasts. In terms of professional photography techniques, atmospheric lighting, natural gradients, and cinematic depth, Jimeng’s contrast was too intense. FLUX.1 had the best gradient effect, while Dall-E3's gradients felt forced and ineffective. Results are shown in the left subplot of Figure \ref{fig_photo_3}.

\textbf{Setting 8:} Double exposure, intended to capture reflections of people on glass, was tested. Only FLUX.1 and Midjourney met expectations; Ideogram2.0 and Dall-E3 partially achieved the effect, while Jimeng had clear issues, and Stable Diffusion 3 completely failed to recognize the keyword. For soft focus, delicate light play, cinematic quality, soft shadows, and artistic composition, the overall softness was best in Midjourney and FLUX.1. Results are shown in the middle subplot of Figure \ref{fig_photo_3}.

\textbf{Setting 9:} The 28mm lens, a wide-angle lens that maintains background clarity, was tested. Ideogram2.0 and Jimeng did not achieve this effect. For studio lighting, interpreted as artificial lighting typical of a studio, Ideogram2.0 only captured regular artificial light. FLUX.1, Midjourney, and Jimeng performed best in high-definition photography, professional lighting, cinematic depth, and soft focus, which emphasized facial contours and details, while others were slightly weaker. Results are shown in the right subplot of Figure \ref{fig_photo_3}.

\textbf{Setting 10:} Involving aerial environment photography and the blue hour, Dall-E3’s results were slightly inferior; others performed well. The requirement for a high-resolution image from a high vantage point with the Sony A7R IV was best met by Ideogram2.0 and Midjourney, with more harmonious and softer color tones. Results are shown in the left subplot of Figure \ref{fig_photo_4}.

\textbf{Setting 11:} Under cinematic lighting, Dall-E3 produced the best facial lighting, Midjourney achieved a dreamy effect, and FLUX.1 had the most realistic lighting. For surrealism, vibrant colors, and professional photography techniques—using methods like distortion, collage, and supernatural elements to create dreamlike atmospheres. FLUX.1 failed to capture this, Stable Diffusion 3 had issues with hand and scene generation, Jimeng generated anime-style images, and Ideogram2.0 mistook kite shapes for fish. Midjourney depicted kites as butterflies but with excellent overall style, while Dall-E3 performed best. Results are shown in the middle subplot of Figure \ref{fig_photo_4}.

\textbf{Setting 12:} In street photography, soft and diffused light was required; Dall-E3 notably violated this, and Ideogram2.0’s tone was too cool. For cinematic framing, dynamic composition, natural reflections, urban realism, and soft lighting, reflections were best captured by FLUX.1 and Midjourney. Jimeng, Stable Diffusion 3, and Ideogram2.0 showed varying issues, with Ideogram2.0’s water ripple effects being notably problematic, and Dall-E3’s composition defying logic. Results are shown in the right subplot of Figure \ref{fig_photo_4}.

\textbf{Score. }
The results of this experiment are shown in Table \ref{tab_photo}. These metrics differ significantly from human intuition, with only GPT-4o and CLIPScore's scores being relatively consistent with human evaluations. This may be due to the presence of numerous technical terms related to photography in the prompts, which the other metrics may not fully comprehend.
\begin{table}[h]
\vspace{1em}
    \centering
    \caption[Section~\ref{photo}: photographic image generation.]{The scoring of generation results by six models on photographic image generation under different evaluation systems. Refer to Section \ref{photo} for detailed discussions.}
    \begin{tabular}{l|c|c|c|c|c}
        \midrule
        Model & CLIPScore & HPSv2 & Aesthetic Score & GPT-4o & Human \\
        \midrule
        FLUX.1      & 28.49 & 0.28 & 6.09 &   7.29 & \textbf{9.38} \\
        Ideogram2.0 & 28.88 & \textbf{0.30} & 6.22 &  7.57 & 7.08 \\
        Dall-E3     & 29.31 & 0.28 & 6.13 &  6.87 & 5.90 \\
        Midjourney  & \textbf{30.70} & 0.29 & 6.27 &  \textbf{8.61} & 8.68 \\
        SD3         & 29.56 & \textbf{0.30} & 6.38 &  7.01 & 6.87 \\
        Jimeng      & 29.93 & \textbf{0.30} & \textbf{6.46} &  6.66 & 7.56 \\
        \midrule
    \end{tabular}
    \label{tab_photo}
\end{table}

\subsubsection{Perspective Relation}
\label{perspective}
\begin{figure*}[!ht]
  \centering 
  \makebox[\textwidth][c]{\includegraphics[width=0.75\textwidth]{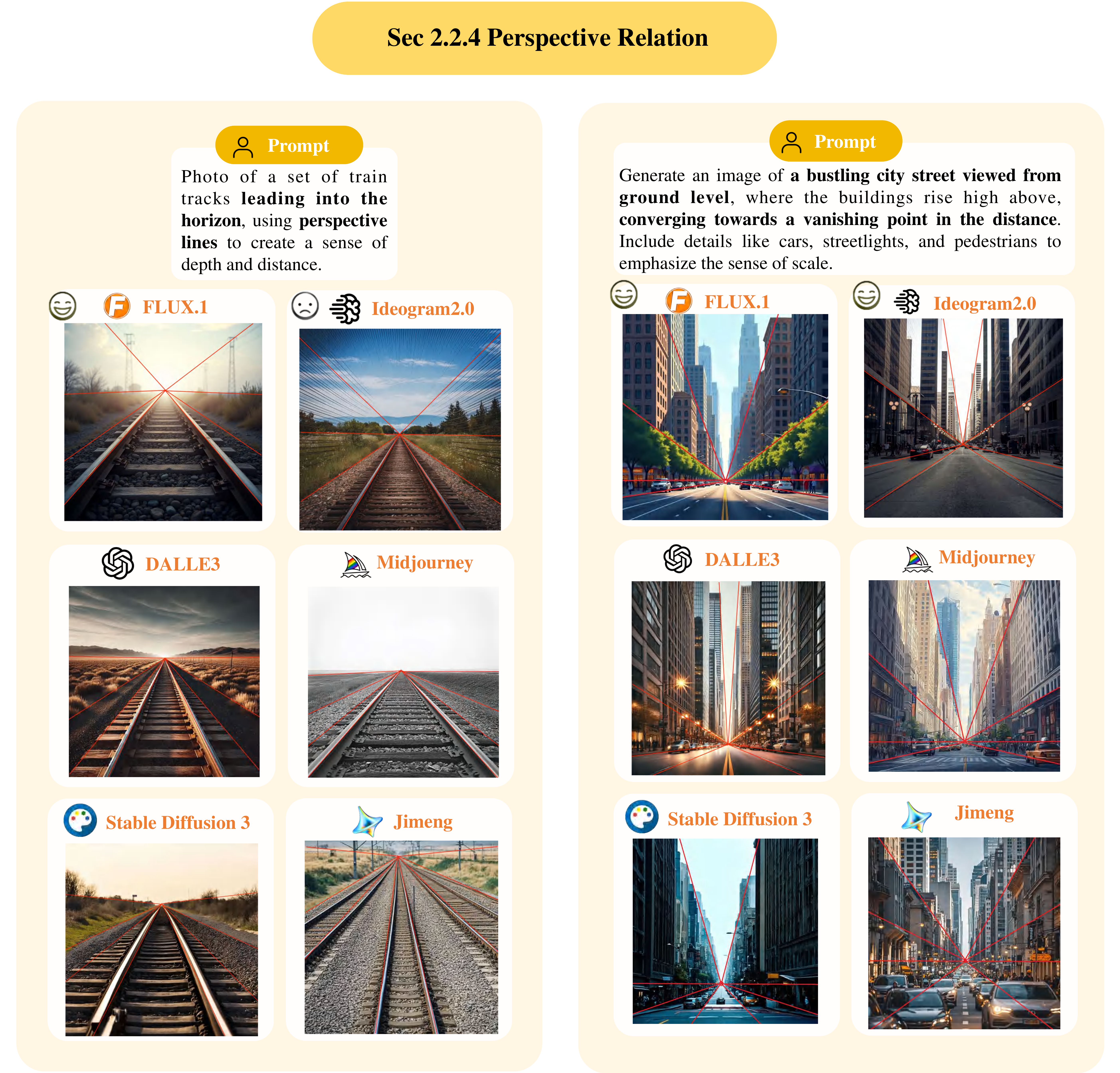}}
  \caption[Section~\ref{perspective}: perspective relation.]{Results on perspective relation task. Refer to Section \ref{perspective} for detailed discussions.}
  \label{fig_perspective_1}
\end{figure*}

\begin{figure*}[!ht]
  \centering 
  \makebox[\textwidth][c]{\includegraphics[width=0.7\textwidth]{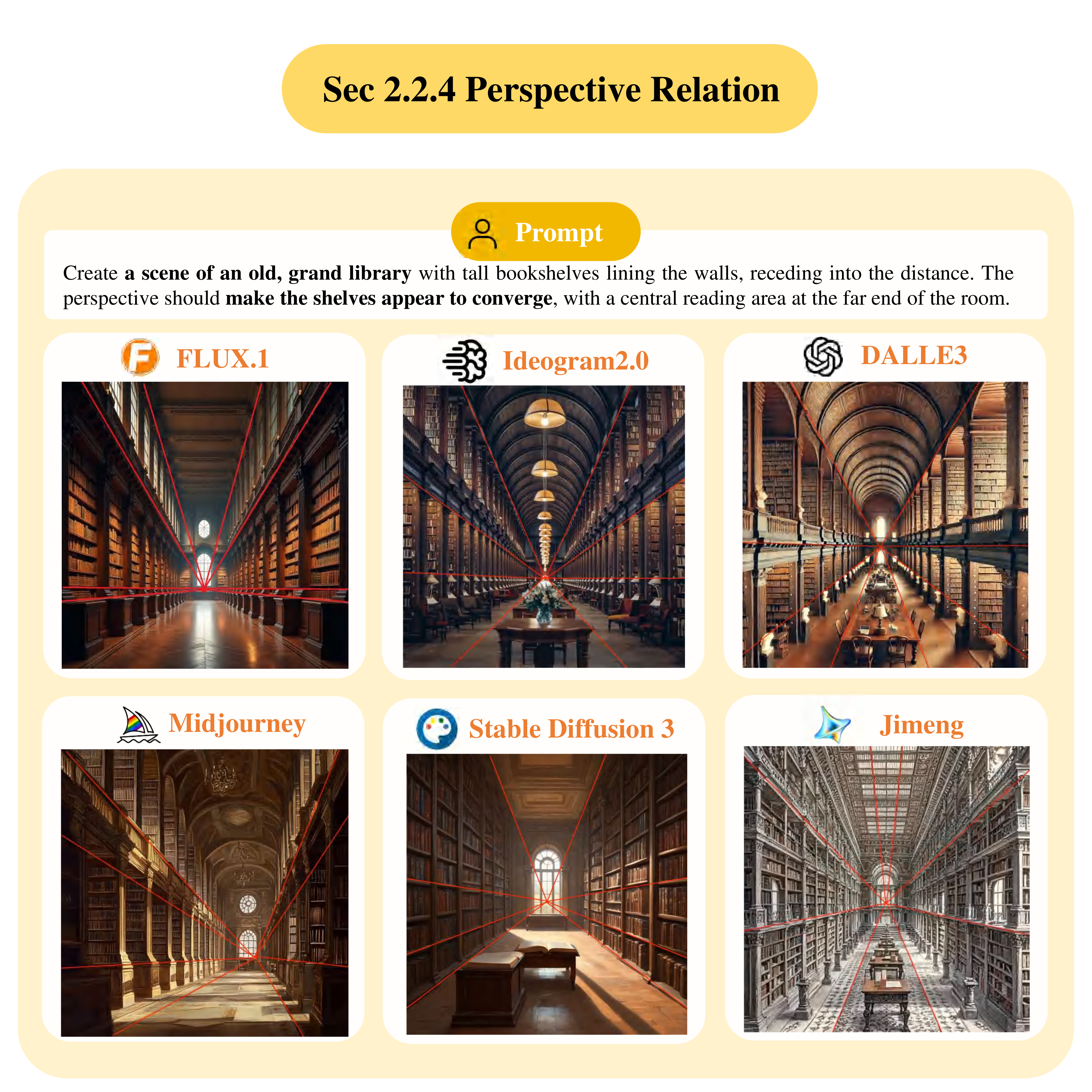}}
  \caption[Section~\ref{perspective}: perspective relation.]{Results on perspective relation task. Refer to Section \ref{perspective} for detailed discussions.}
  \label{fig_perspective_2}
\end{figure*}

In Figure \ref{fig_perspective_1} and Figure \ref{fig_perspective_2}, we evaluated the models' ability to correctly handle perspective relationships. Most of the tested models demonstrated excellent performance, whether dealing with simple track scenes or more complex urban streets and library settings, generally aligning well with real-world perspective. However, Stable Diffusion 3 produced images with a certain degree of distortion, performing the worst in terms of matching real-world perspective relationships.

\textbf{Score. }
The results of this experiment are shown in Table \ref{tab_persp}. We can see that in this task, only GPT-4o's scores align relatively well with human ratings. This may be because the experiment involves physical relationships such as perspective, requiring the evaluation metrics to have a certain understanding of the fundamental principles of the physical world.
\begin{table}[h]
    \centering
    \caption[Section~\ref{perspective}: perspective relation.]{The scoring of generation results by six models on perspective relation under different evaluation systems. Refer to Section \ref{perspective} for detailed discussions.}
    \begin{tabular}{l|c|c|c|c|c}
        \midrule
        Model & CLIPScore & HPSv2 & Aesthetic Score & GPT-4o & Human \\
        \midrule
        FLUX.1      & 25.42 & 0.27 & 5.82 &  7.50 & \textbf{8.61} \\
        Ideogram2.0 & 27.31 & 0.29 & \textbf{6.38} &  5.28 & 5.56 \\
        Dall-E3     & 25.95 & 0.29 & 6.32 &  7.78 & 7.50 \\
        Midjourney  & 26.58 & 0.26 & 6.00 &  \textbf{8.89} & 8.05 \\
        SD3         & 27.38 & 0.28 & 5.90 &  7.22 & 7.50 \\
        Jimeng      & \textbf{28.26} & \textbf{0.30} & 6.34 &   \textbf{8.89} & 8.33 \\
        \midrule
    \end{tabular}
    \label{tab_persp}
\end{table}

\subsubsection{Physical understanding}
\label{physical}
In the T2I pipeline, we give the image caption to the model, then the model generates an image reflecting the caption content, visually correct. In this process, does the model do understand the world's physical law~\cite{Meng2024PhyBenchAP}? To test this point, we describe a real-world physical scene in the prompt. To generate an image that conforms to the laws of physics, the models need to truly understand the physical law. Here we describe two scenes: a glass cup falling to the ground and the water's temperature is over 100 celsius degrees. 

\begin{figure*}[!ht]
  \centering 
  \makebox[\textwidth][c]{\includegraphics[width=0.85\textwidth]{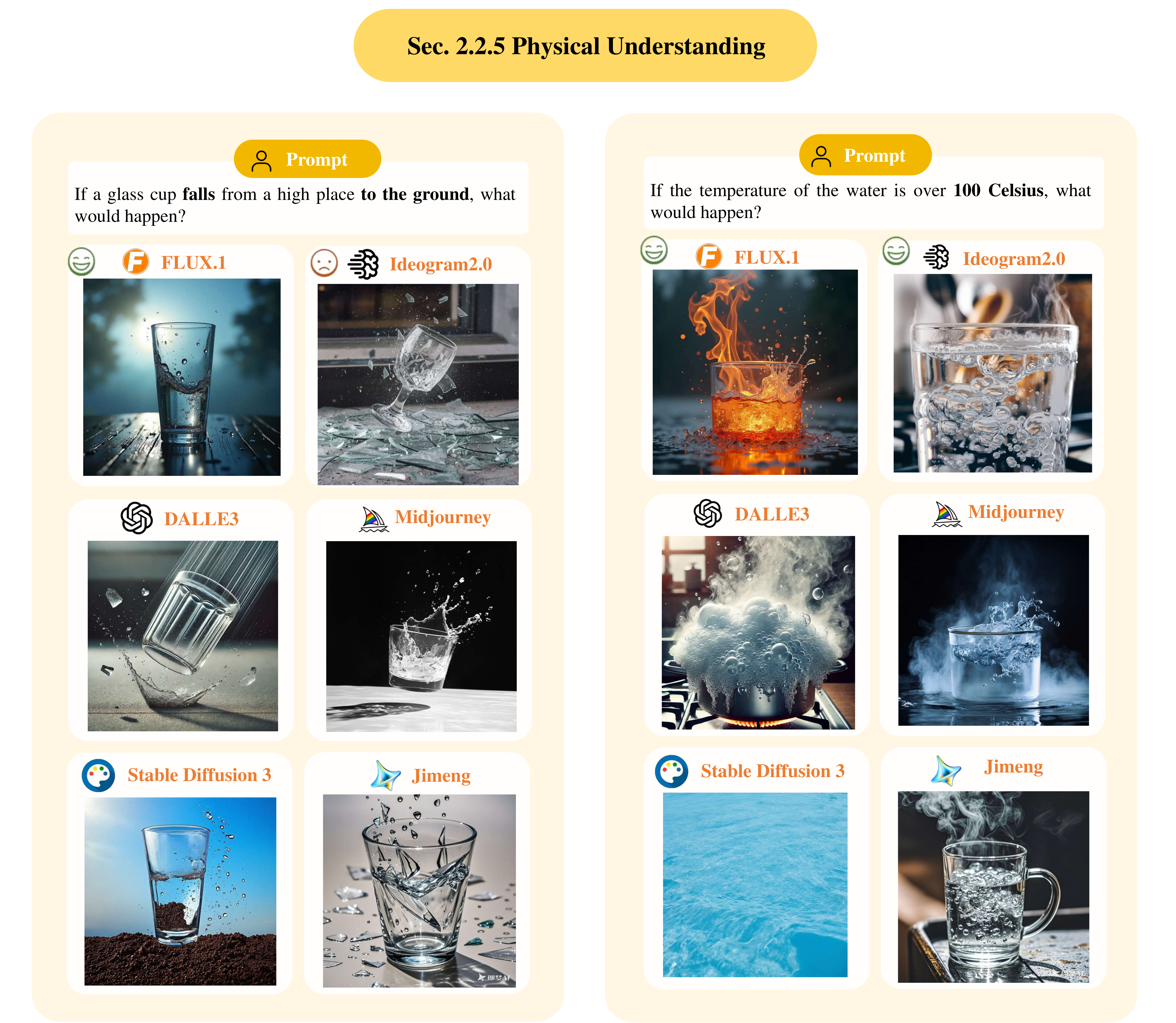}}
  \caption[Section~\ref{physical}: physical understanding.]{Results on physical understanding task. Refer to Section \ref{physical} for detailed discussions.}
  \label{fig_physical}
\end{figure*}

The results are shown in Figure \ref{fig_physical}. In the first scene, only Ideogram2.0 and Jimeng can generate the physically correct image: the glass cup shattered into pieces. In the second scene, all models perform well: the water boiled, except FLUX.1.

\textbf{Score. }
The results of this task are shown in the Table \ref{tab_physical}. It can be observed that HPSv2, GPT-4o, and human perception are largely consistent. Ideogram2.0 achieved the highest score in the Aesthetic Score, which also aligns with human perception. However, the CLIPScore differs significantly from human perception.
\begin{table}[h]
    \centering
    \caption{The scoring of generation results by six models on physical understanding under different evaluation systems. Refer to Section \ref{physical} for detailed discussions.}
    \begin{tabular}{l|c|c|c|c|c}
        \midrule
        Model & CLIPScore & HPSv2 & Aesthetic Score & GPT-4o & Human \\
        \midrule
        FLUX.1 & 22.10 & 0.26 & 5.67 &  2.92 & 4.17 \\
        Ideogram2.0 & 20.79 & \textbf{0.27} & \textbf{5.99} &  \textbf{8.34} & \textbf{9.16} \\
        Dall-E3 & 24.94 & 0.26 & 5.91 &  7.08 & 6.66 \\
        Midjourney & \textbf{25.94} & 0.26 & 5.76 &  5.42 & 7.50 \\
        SD3 & 22.78 & 0.24 & \textbf{5.99} &  4.17 & 5.00 \\
        Jimeng & 23.58 & 0.20 & 5.22 &  6.25 & 6.25 \\
        \midrule
    \end{tabular}
    \label{tab_physical}
\end{table}

\clearpage
\subsection{Specific Domain Generation}
\label{sec:04specific}
With the advancement of T2I models, their usefulness has expanded to various domains. These models hold the potential to generate high-quality, domain-specific data, paving the way for significant contributions to technological innovation and interdisciplinary research.

In Section \ref{math}, we assess the models' understanding of mathematical terminology and their capability to generate math-related images based on given descriptions. Section \ref{fractal} explores the models' performance in generating images within fractal settings. In Section \ref{medical}, we evaluate the models' capability to produce medical images with potential applications in medical research. In Sections \ref{pointcloud} and \ref{mesh}, we prompt the models to generate 3D images. Lastly, in Sections \ref{chemistry} and \ref{bio}, we assess the models' ability to generate images related to chemistry and biology. Section \ref{robotics} explores the working environments of robots in embodied intelligence, while Section \ref{autodrive} investigates tasks in autonomous driving scenarios.

\subsubsection{Math}
\label{math}
In Figure \ref{mathimage}, we explore the models' mathematical ability, especially geometrical concepts. For the first example of a right-angled triangle in Figure \ref{mathimage}, FLUX.1~\cite{flux2024}, Stable Diffusion 3~\cite{rombach2022high} and Jimeng try to present the outputs in a mathematical format. However, FLUX.1 fails to accurately depict the correct geometric relationships, and the output from Stable Diffusion 3 is fuzzy and irrelevant. Jimeng successfully generates a correct right-angled triangle, though the image contains the wrong text. Ideogram2.0~\cite{ideogram2.0} and Midjourney mistakenly focus too much on the word "measuring" in the prompt, thus Ideogram2.0 generates rulers arranged in the shape of a right triangle, and Midjourney presents a dimensional figure irrelevant. Dall-E3~\cite{betker2023improving} cannot recognize the prompt as a math concept.
In the second example of an inscribed circle within an isosceles triangle, the style of the results is similar to the first. In detail, all models generate the correct isosceles triangle but the wrong inscribed circle. Current T2I models are lacking in the ability to generate mathematically relevant images, they cannot accurately understand some mathematical concepts, and it is difficult to generate images that conform to analytic geometric.

\begin{figure*}[!ht]
  \centering
  \makebox[\textwidth][c]{\includegraphics[width=0.8\textwidth]{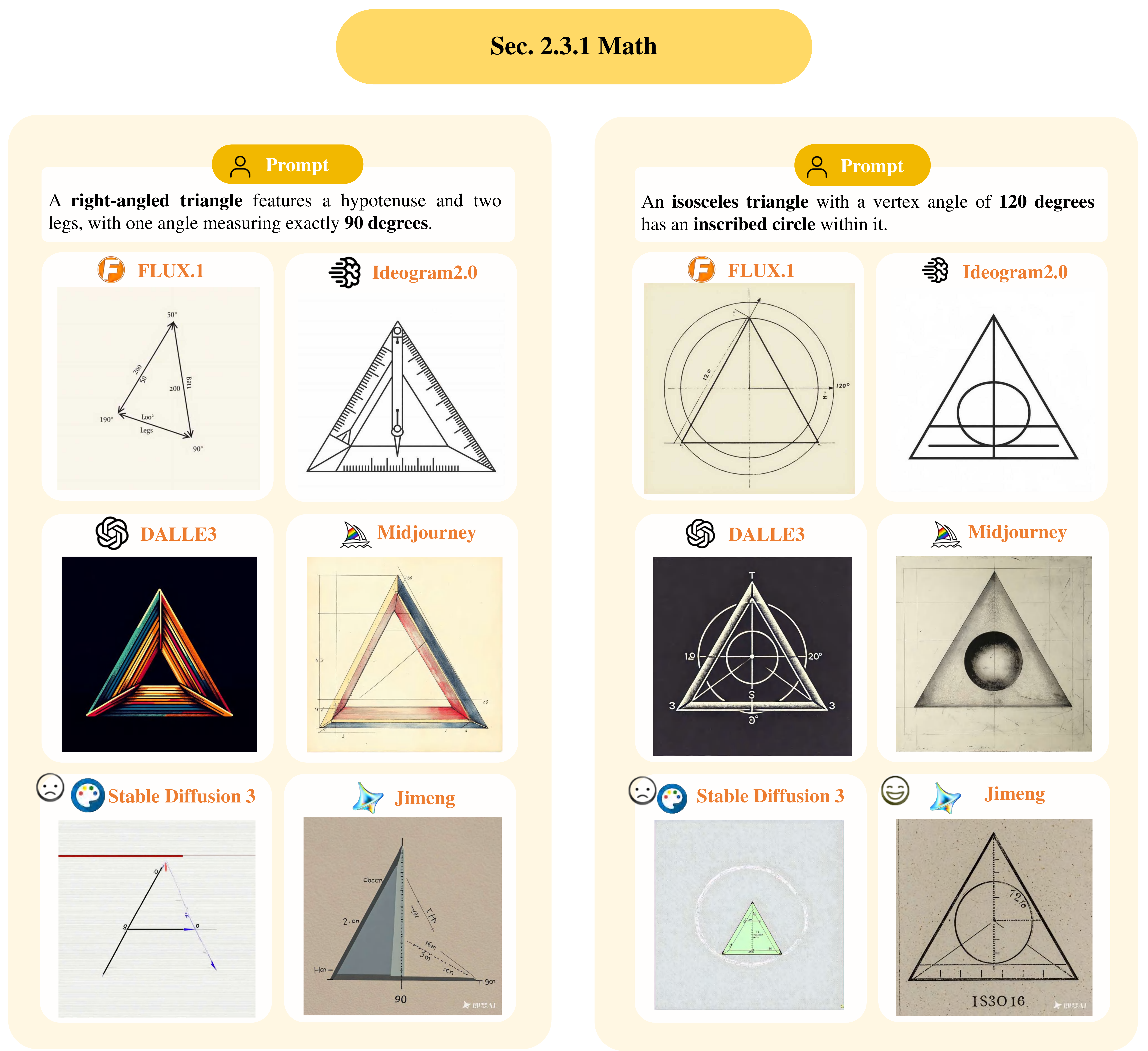}}
  \caption[Section~\ref{math}: math.]{Results on math task. Refer to Section \ref{math} for detailed discussions.}
  \label{mathimage}
\end{figure*}

\textbf{Score. }
The results of this experiment are shown in Table \ref{tab_math}. Midjourney received higher scores from human evaluations, primarily because the outputs of several models do not effectively grasp the mathematical concepts in the prompts. As a result, the scoring mainly focuses on aspects like aesthetics and realism, with other metrics showing some discrepancies compared to human intuition.
\begin{table}[h]
    \centering
    \caption[Section~\ref{math}: math.]{The scoring of generation results by six models on math image design under different evaluation systems. Refer to Section \ref{math} for detailed discussions.}
    \begin{tabular}{l|c|c|c|c|c}
        \midrule
        Model & CLIPScore & HPSv2 & Aesthetic Score & GPT-4o & Human \\
        \midrule
        FLUX.1      & \textbf{25.84} & 0.20 & 4.72 &  5.00 & 4.58 \\
        Ideogram2.0 & 20.71 & 0.19 & \textbf{5.76} &   5.00 & 5.42 \\
        Dall-E3     & 23.91 & 0.20 & 5.03 &   4.17 & 6.25 \\
        Midjourney  & 25.62 & 0.18 & 4.72 &   3.75 & \textbf{7.08} \\
        SD3         & 24.43 & \textbf{0.23} & 5.05 &   4.17 & 3.33 \\
        Jimeng      & 25.50 & 0.21 & 4.51 &   \textbf{6.25} & 5.42 \\
        \midrule
    \end{tabular}
    \label{tab_math}
\end{table}

\subsubsection{Fractal}
\label{fractal}
In this section, we evaluate the models' ability to generate images within fractal settings, which require understanding complex recursive patterns. These patterns are often used in mathematical and artistic contexts to depict natural phenomena like coastlines, snowflakes, and more.

For the first experiment, we prompted the models to generate a Mandelbrot set. FLUX.1 and Midjourney produced visually appealing fractals with detailed recursive structures. However, Dall-E3 and Stable Diffusion 3 struggled with the intricacy of the pattern, resulting in less accurate representations. In the second experiment involving the Sierpinski triangle, FLUX.1 and Jimeng successfully captured the recursive nature of the fractal, accurately depicting the triangular subdivisions. Ideogram2.0 misinterpreted the prompt, generating a series of disjointed triangles, while Dall-E3 created a pattern resembling the Sierpinski triangle but lacking precise detail. Overall, the experiments reveal that while some models can generate fractal images, consistency and accuracy vary. This suggests that improvements in understanding recursive algorithms might enhance their performance in this domain.

\textbf{Score. }
The results of this experiment are shown in the Table \ref{tab_fractal}. It can be observed that FLUX.1 performed the best in this experiment. The Aesthetic Score aligns more closely with human intuitive perception, while the GPT results show a significant difference from human perception.
\begin{table}[h]
    \centering
    \caption[Section~\ref{fractal}: fractal.]{The scoring of generation results by six models on fractal image design under different evaluation systems. Refer to Section \ref{fractal} for detailed discussions.}
    \begin{tabular}{l|c|c|c|c|c}
        \midrule
        Model & CLIPScore & HPSv2 & Aesthetic Score &  GPT-4o & Human \\
        \midrule
        FLUX.1      & 24.93 & 0.25 & \textbf{6.09} &   5.42 & \textbf{7.92} \\
        Ideogram2.0 & 23.46 & \textbf{0.25} & 5.73 &  6.25 & 5.84 \\
        Dall-E3     & \textbf{28.29} & 0.24 & 5.63 &  5.00 & 4.59 \\
        Midjourney  & 25.54 & 0.21 & 6.01 &  4.59 & 6.67 \\
        SD3         & 25.90 & 0.23 & 5.41 &  6.25 & 2.50 \\
        Jimeng      & 19.78 & 0.21 & 5.44 & \textbf{7.92} & 7.09 \\
        \midrule
    \end{tabular}
    \label{tab_fractal}
\end{table}

\begin{figure*}[!ht]
\vspace{-4em}
  \centering
  \makebox[\textwidth][c]{\includegraphics[width=0.8\textwidth]{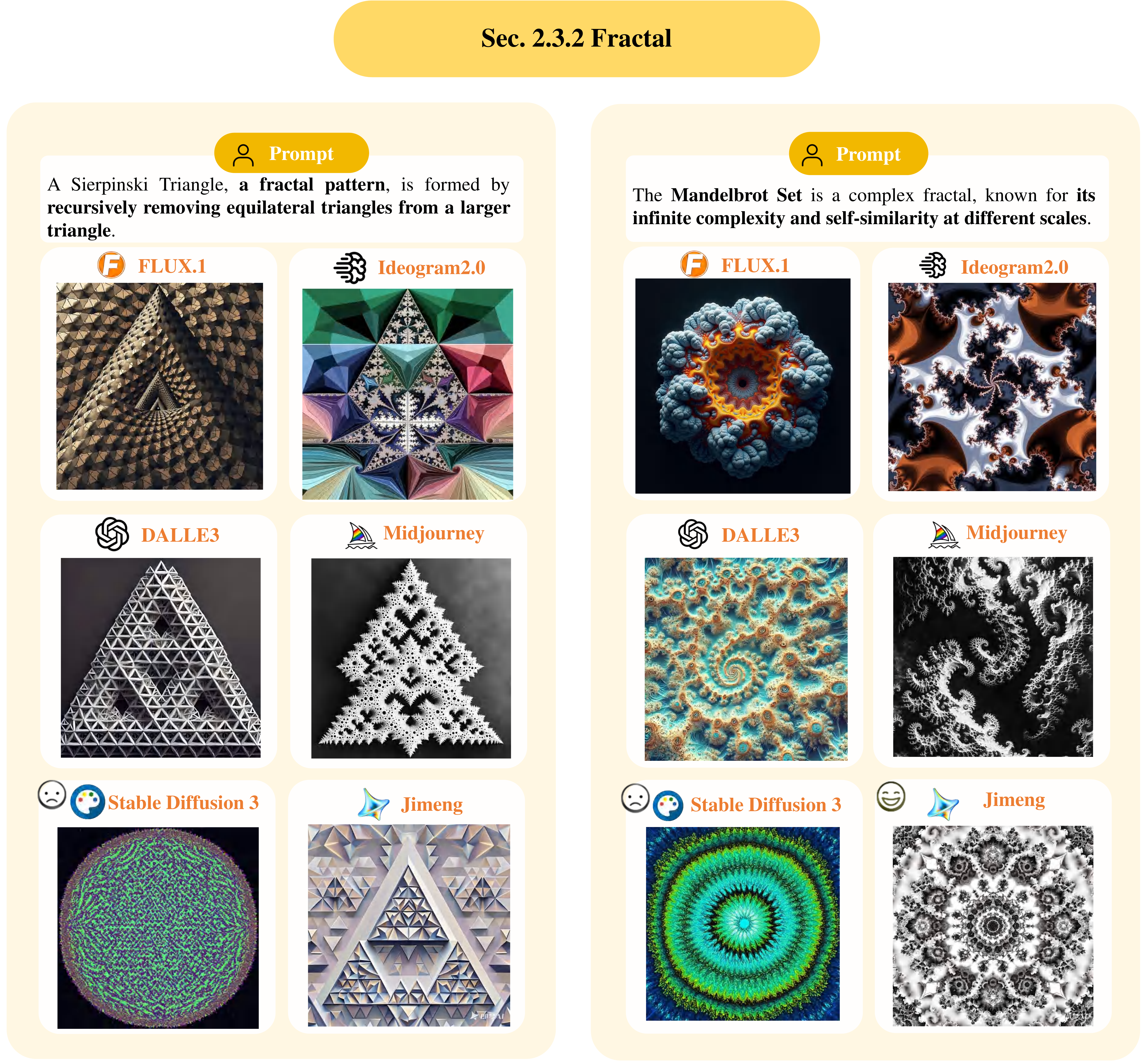}}
  \caption[Section~\ref{fractal}: Fractal.]{Results on fractal task. Refer to Section \ref{fractal} for detailed discussions.}
  \label{mathimage}
\end{figure*}

\begin{figure*}[!ht]
  \centering
  \makebox[\textwidth][c]{\includegraphics[width=0.8\textwidth]{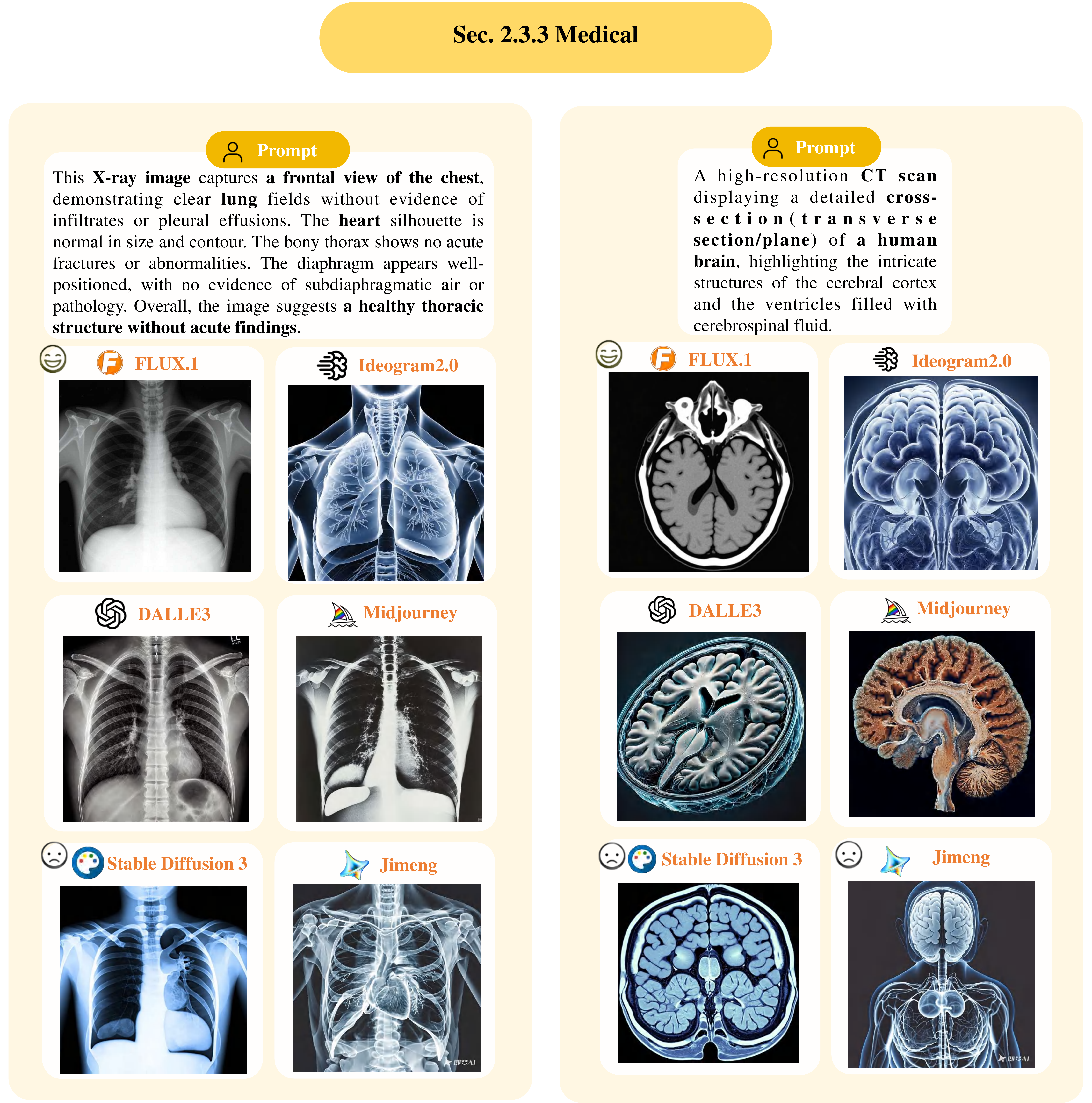}}
  \caption[Section~\ref{medical}: medical.]{Results on medical task. Refer to Section \ref{medical} for detailed discussions.}
  \label{medical_image}
\end{figure*}

\subsubsection{Medical}
\label{medical}
In this experiment, we tested the ability of T2I models to generate medical images~\cite{Abaid2024SynthesizingCI}. For the first prompt, we asked the models to generate an X-ray image capturing a frontal view of the chest. We found that Ideogram2.0, Jimeng, and Stable Diffusion 3 did not produce accurate X-ray images, as indicated by the color and texture of their outputs. Additionally, these three models generated chaotic representations of the shoulder joints, and Ideogram2.0 produced an incorrect morphology of the lungs. Midjourney generated an image that resembled an X-ray, but the structures of the heart and liver were significantly flawed. Dall-E3 and FLUX.1 performed the best, producing images with an almost correct morphology of bones and organs, with only minor inaccuracies in the structure of the shoulder joints.

In the second prompt, we asked the models to generate a CT scan displaying a detailed cross-section of a human brain. Ideogram2.0 and Dall-E3's outputs did not resemble a CT scan, and the brain structures were incorrect. Jimeng failed to generate a cross-section of the brain, and the organs in the images appeared very chaotic. Midjourney's output contained mistakes in the locations of the cerebrum and cerebellum, and the internal structures of the brain were disorganized. Stable Diffusion 3's result was not realistic enough; the brain's edge were overly defined, the proportion of black areas representing cerebrospinal fluid was too small compared to a real brain, and the real brain does not have such notches at the back of the head. Only FLUX.1 produced a result closest to a real brain CT scan, leading us to suspect that FLUX.1's training data may include a certain proportion of high-quality medical images.

\textbf{Score. }
The results of this experiment are shown in the Table \ref{tab_medical}. Except for the CLIPScore, FLUX.1 performed the best in all other scores.
\begin{table}[h]
    \centering
    \caption[Section~\ref{medical}: medical.]{The scoring of generation results by six models on medical image generation under different evaluation systems. Refer to Section \ref{medical} for detailed discussions.}
    \begin{tabular}{l|c|c|c|c|c}
        \midrule
        Model & CLIPScore & HPSv2 & Aesthetic Score & GPT-4o & Human \\
        \midrule
        FLUX.1      & 23.66 & \textbf{0.22} & \textbf{5.73} &  \textbf{6.66} & \textbf{7.78} \\
        Ideogram2.0 & 24.38 & \textbf{0.22} & 4.63 & 5.28 & 6.39 \\
        Dall-E3     & 27.64 & 0.21 & 4.68 &  6.11 & 5.00 \\
        Midjourney  & 26.42 & \textbf{0.22} & 5.10 & 5.28 & 5.84 \\
        SD3         & \textbf{28.12} & 0.19 & 5.10 & 4.17 & 6.66 \\
        Jimeng      & 27.16 & \textbf{0.22} & 4.82 & 4.17 & 4.17 \\
        \midrule
    \end{tabular}
    \label{tab_medical}
\end{table}

\subsubsection{3D Point Cloud}
\label{pointcloud}
\begin{figure*}[!ht]
  \centering
  \makebox[\textwidth][c]{\includegraphics[width=1\textwidth]{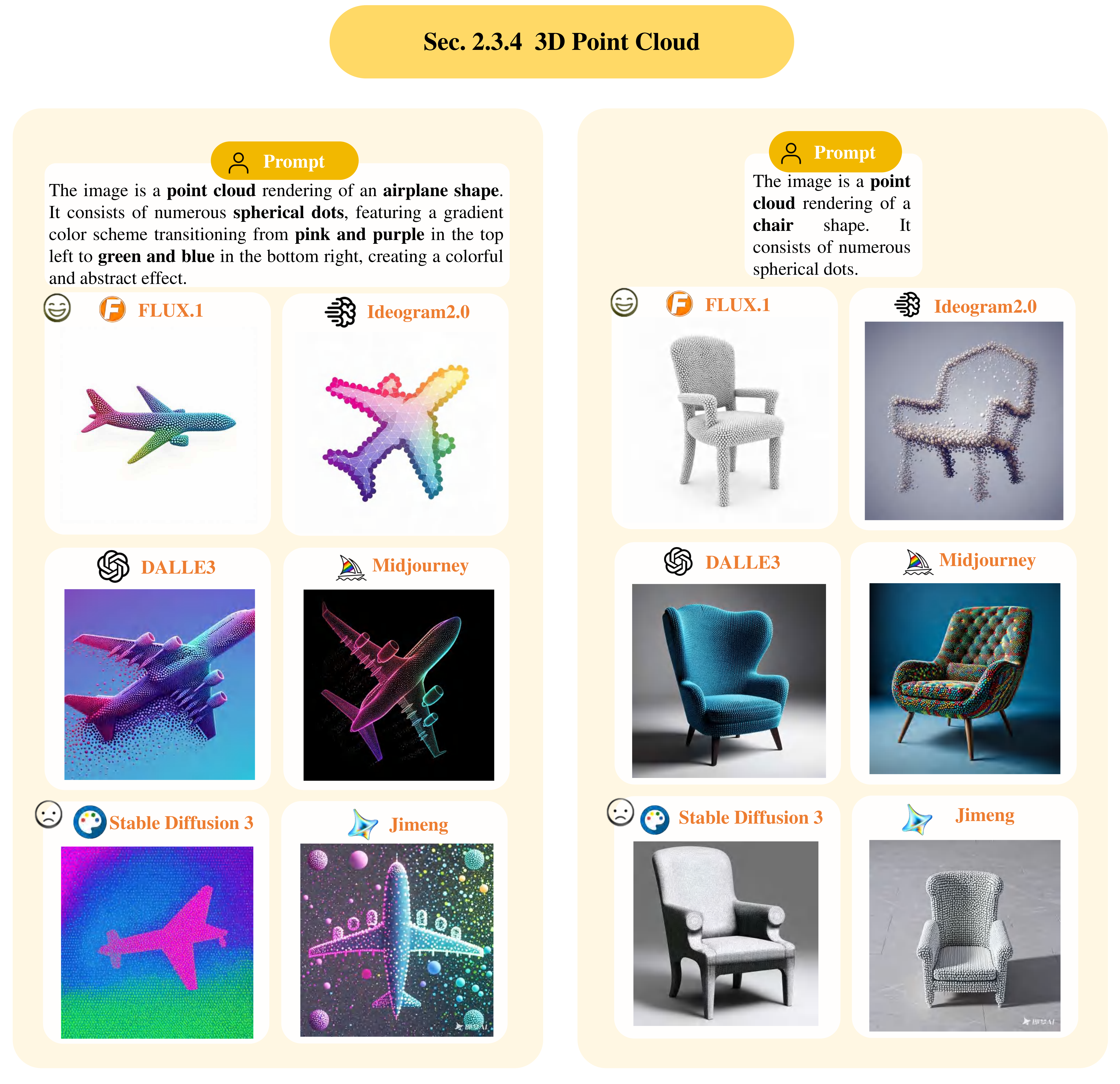}}
  \caption[Section~\ref{pointcloud}: 3D point cloud.]{Results on 3D point cloud task. Refer to Section \ref{pointcloud} for detailed discussions.}
  \label{pointcloud_image}
\end{figure*}

In Figure \ref{pointcloud_image}, we investigate the models' capacities to generate images in the form of 3D point cloud. For the first example of an airplane, FLUX.1, Dall-E3 and Midjourney successfully generate point cloud representations, while the outputs of Dall-E3 and Midjourney are imperfect due to the points outside the airplane. Ideogram2.0, Stable Diffusion 3 and Jimeng fail to accurately render the point cloud, with Ideogram2.0 even producing an image more like 3D mesh. In the second example of a chair, FLUX.1 again succeeds in producing a correct point cloud representation. Additionally, Ideogram2.0 also generates an image of a chair in point cloud form, but of poor quality. Stable Diffusion 3 misunderstands \textit{point cloud} and generates a chair painted with dots, similar to its output in the first example. Dall-E3, Midjourney and Jimeng misinterpret point cloud as well, generating images of a chair constructed from spherical shapes. Overall, FLUX.1 is the only model to perform well in both cases. It is worth mentioning that the objects and prompts are well-designed. We observed that the models struggle to generate accurate point clouds for certain objects, indicating that additional training may be necessary for tasks involving point cloud generation.

\textbf{Score. }
The results of this experiment are shown in the table. In this experiment, the performance of the models is similar, with FLUX.1, Dall-E3, and Jimeng performing slightly better.
\begin{table}[h]
    \centering
    \caption[Section~\ref{pointcloud}: point cloud.]{The scoring of generation results by six models on point cloud image design under different evaluation systems. Refer to Section \ref{pointcloud} for detailed discussions.}
    \begin{tabular}{l|c|c|c|c|c|c}
        \midrule
        Model & CLIPScore & HPSv2 & Aesthetic Score & GPT-4o & Human \\
        \midrule
        FLUX.1      & 32.24 & 0.26 & 5.30 & 5.00 & \textbf{7.50} \\
        Ideogram2.0 & \textbf{33.78} & 0.30 & 5.95 &  6.25 & 6.67 \\
        Dall-E3     & 32.18 & \textbf{0.31} & 5.94 &  \textbf{7.50} & \textbf{7.50} \\
        Midjourney  & 29.73 & 0.25 & 5.70 & 6.66 & 6.67 \\
        SD3         & 32.93 & 0.28 & 5.76 &  6.66 & 7.08 \\
        Jimeng      & 32.65 & \textbf{0.31} & \textbf{6.30} &  5.00 & \textbf{7.50} \\
        \midrule
    \end{tabular}
    \label{tab_pointcloud}
\end{table}

\subsubsection{3D Mesh}
\label{mesh}
\begin{figure*}[!ht]
  \centering
  \makebox[\textwidth][c]{\includegraphics[width=1\textwidth]{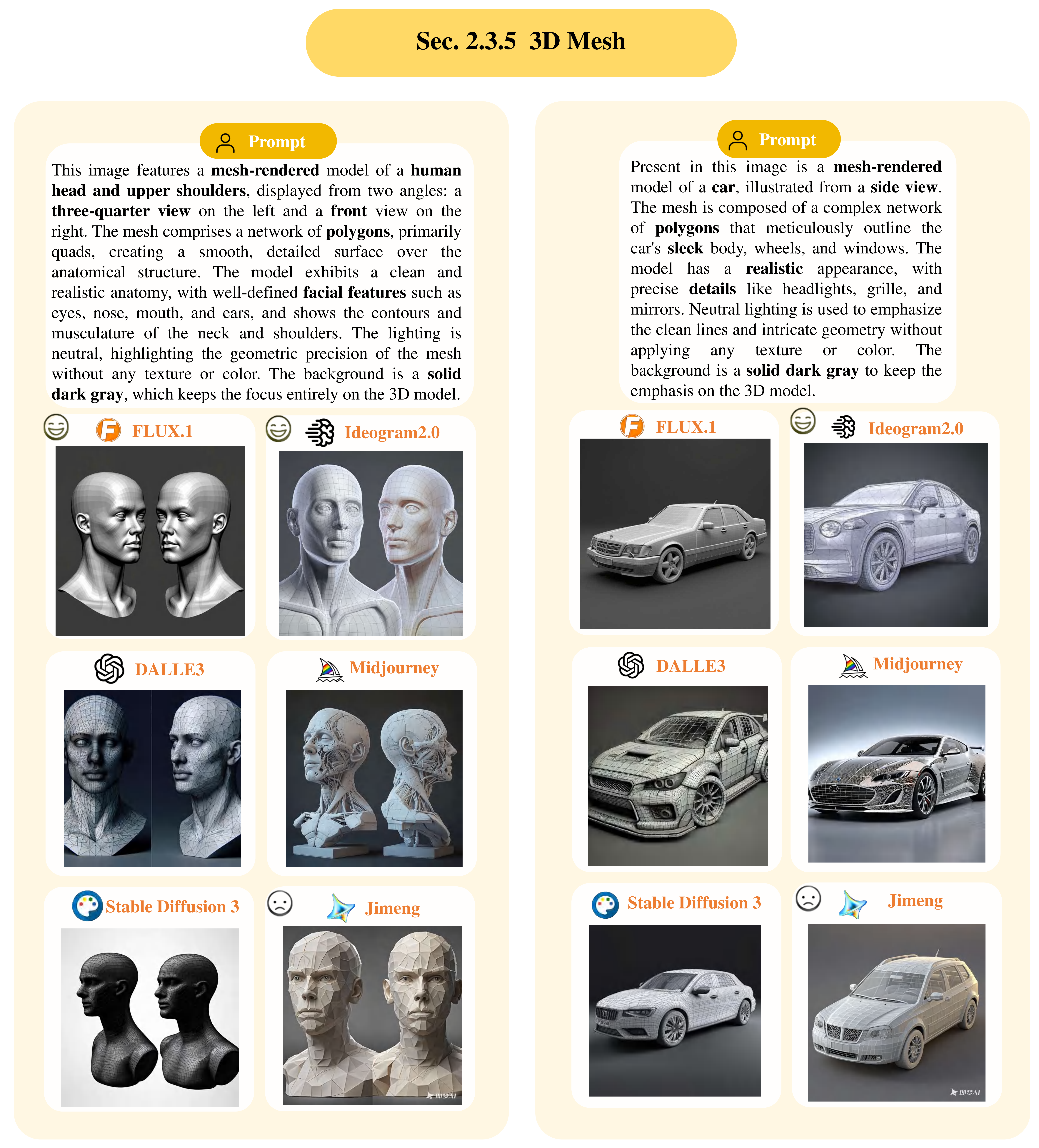}}
  \caption[Section~\ref{mesh}: 3D mesh.]{Results on 3D mesh task. Refer to Section \ref{mesh} for detailed discussions.}
  \label{mesh_image}
\end{figure*}

The synthesis of high-quality 3D assets from textual or visual inputs has become a central objective in modern generative modeling~\cite{Heo2025CaPaCS,Kim2025DAViDMD,Huang2025SPAR3DSP,Chen2024GraphicsDreamerIT}. In Figure \ref{mesh_image}, we test the models' ability to generate 3D mesh representations. It is observed that all models can generate images that seem like 3D mesh form. In the first example of a human head, FLUX.1, Ideogram2.0, Dall-E3 and Stable Diffusion 3 successfully generate 3D mesh-like representations. However, there are also mistakes that only Jimeng achieves "a three-quarter view on the left and a front view on the right" required by prompt. Additional errors including the curved edges of the plane in Midjourney and the textured eyes in Jimeng are incorrect in 3D mesh. In the second example of a car, FLUX.1 and Ideogram2.0 also generate correct 3D mesh representations. However, the outputs of Dall-E3, Midjourney and Stable Diffusion 3 are more like dividing a realistic car by lines, and Jimeng even breaks the car into pieces. Overall, FLUX.1 and Ideogram2.0 have the correct understanding of 3D mesh. However, there remains a distinction between the T2I models generating two-dimensional images and 3D generation models.

\textbf{Score. }
The results of this experiment are shown in the Table \ref{tab_mesh}. It can be seen that the model performances are fairly balanced. Overall, the Ideogram2.0 model achieved higher scores.
\begin{table}[h]
    \centering
    \caption[Section~\ref{mesh}: mesh.]{The scoring of generation results by six models on 3D mesh task under different evaluation systems. Refer to Section \ref{mesh} for detailed discussions.}
    \begin{tabular}{l|c|c|c|c|c}
        \midrule
        Model & CLIPScore & HPSv2 & Aesthetic Score & GPT-4o & Human \\
        \midrule
        FLUX.1      & 29.23 & 0.24 & \textbf{6.00} &  5.84 & 7.92 \\
        Ideogram2.0 & 32.64 & \textbf{0.27} & 5.96 &  5.84 & \textbf{8.33} \\
        Dall-E3     & \textbf{32.98} & 0.26 & 5.15 &  5.00 & 7.08 \\
        Midjourney  & 32.36 & 0.24 & 5.27 &  6.25 & 6.25 \\
        SD3         & 31.22 & 0.24 & 5.37 &  \textbf{7.50} & 7.92 \\
        Jimeng      & 29.87 & 0.25 & 5.55 &  6.66 & \textbf{8.33} \\
        \midrule
    \end{tabular}
    \label{tab_mesh}
\end{table}

\subsubsection{Chemistry}
\label{chemistry}
\begin{figure*}[!ht]
  \centering 
  \makebox[\textwidth][c]{\includegraphics[width=0.7\textwidth]{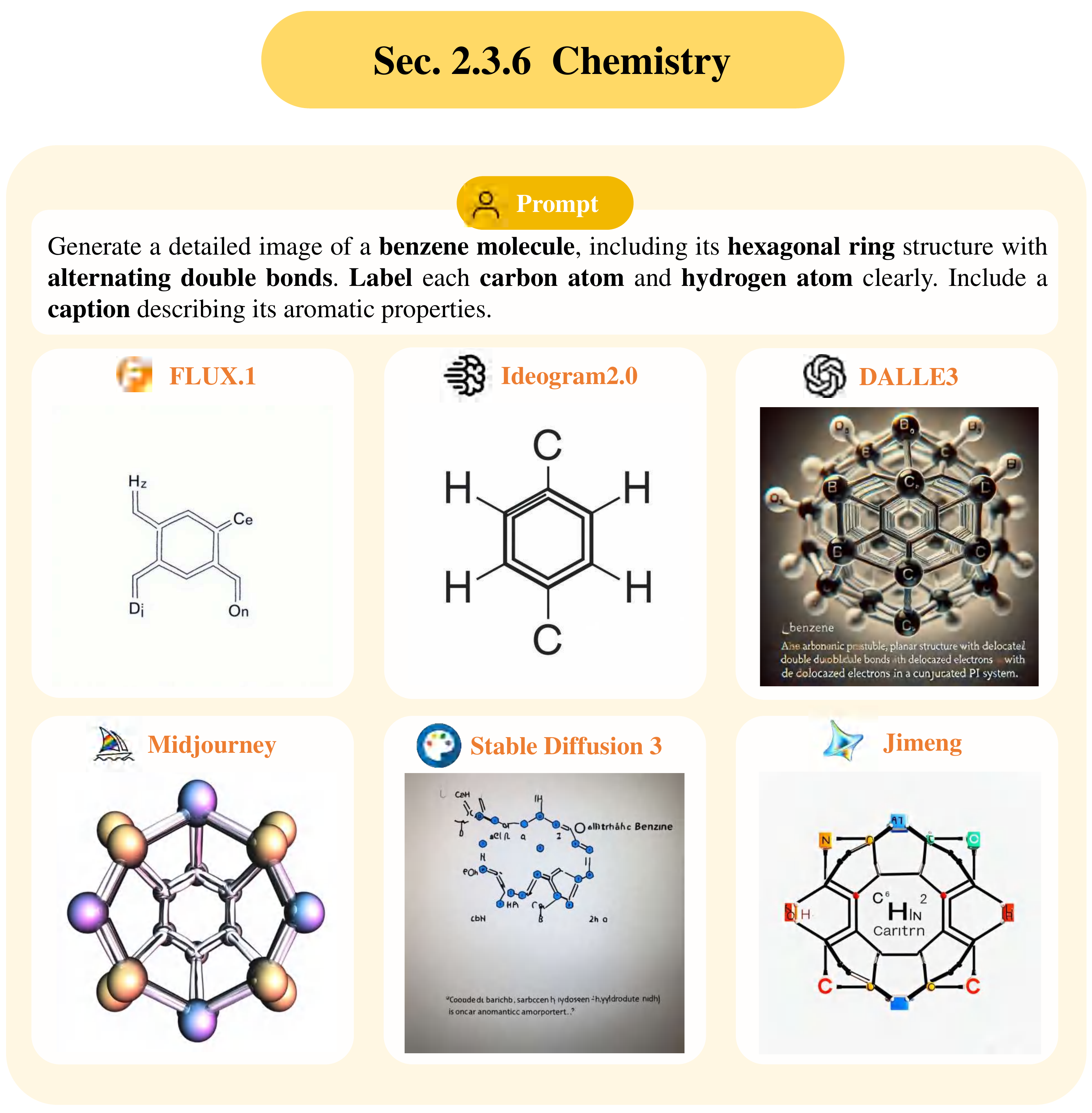}}
  \caption[Section~\ref{chemistry}: chemistry.]{Results on chemistry task. Refer to Section \ref{chemistry} for detailed discussions.}
  \label{chemistry1}
\end{figure*}

In Figures \ref{chemistry1}, we assess the models' capabilities in generating scientifically accurate images within the domain of chemistry. We evaluate the models' understanding of chemistry by prompting them to generate a correct representation of a benzene molecule, but none of the models succeed. Overall, the T2I models struggle to generate scientifically accurate images in the field of chemistry, failing to meet the necessary scientific standards.

\textbf{Score. }
The results of this task are shown in the Table \ref{tab_chemistry}. Ideogram2.0 produced the best outputs in this task. However, the scores from several evaluation systems differ from human intuition, likely because this task requires a certain level of chemistry knowledge, making accurate evaluation more demanding.

\begin{table}[h]
    \centering
    \caption[Section~\ref{chemistry}: chemistry.]{The scoring of generation results by six models on chemistry tasks under different evaluation systems. Refer to Section \ref{chemistry} for detailed discussions.}
    \begin{tabular}{l|c|c|c|c|c}
        \midrule
        Model & CLIPScore & HPSv2 & Aesthetic Score &  GPT-4o & Human \\
        \midrule
        FLUX.1      & \textbf{29.38} & \textbf{0.27} & \textbf{5.45} &   2.50 & 6.67 \\
        Ideogram2.0 & 27.42 & \textbf{0.27} & 5.09 &  5.00 & \textbf{10.00} \\
        Dall-E3     & 28.49 & 0.18 & 3.54 &   2.50 & 8.33 \\
        Midjourney  & 28.65 & 0.20 & 4.66 &   \textbf{7.50} & 6.67 \\
        SD3         & 25.52 & 0.16 & 4.13 &   5.83 & 5.00 \\
        Jimeng      & 26.53 & 0.21 & 5.04 &   2.50 & 4.17 \\
        \midrule
    \end{tabular}
    \label{tab_chemistry}
\end{table}

\subsubsection{Biology}
\label{bio}
\begin{figure*}[!ht]
  \centering 
  \makebox[\textwidth][c]{\includegraphics[width=1.0\textwidth]{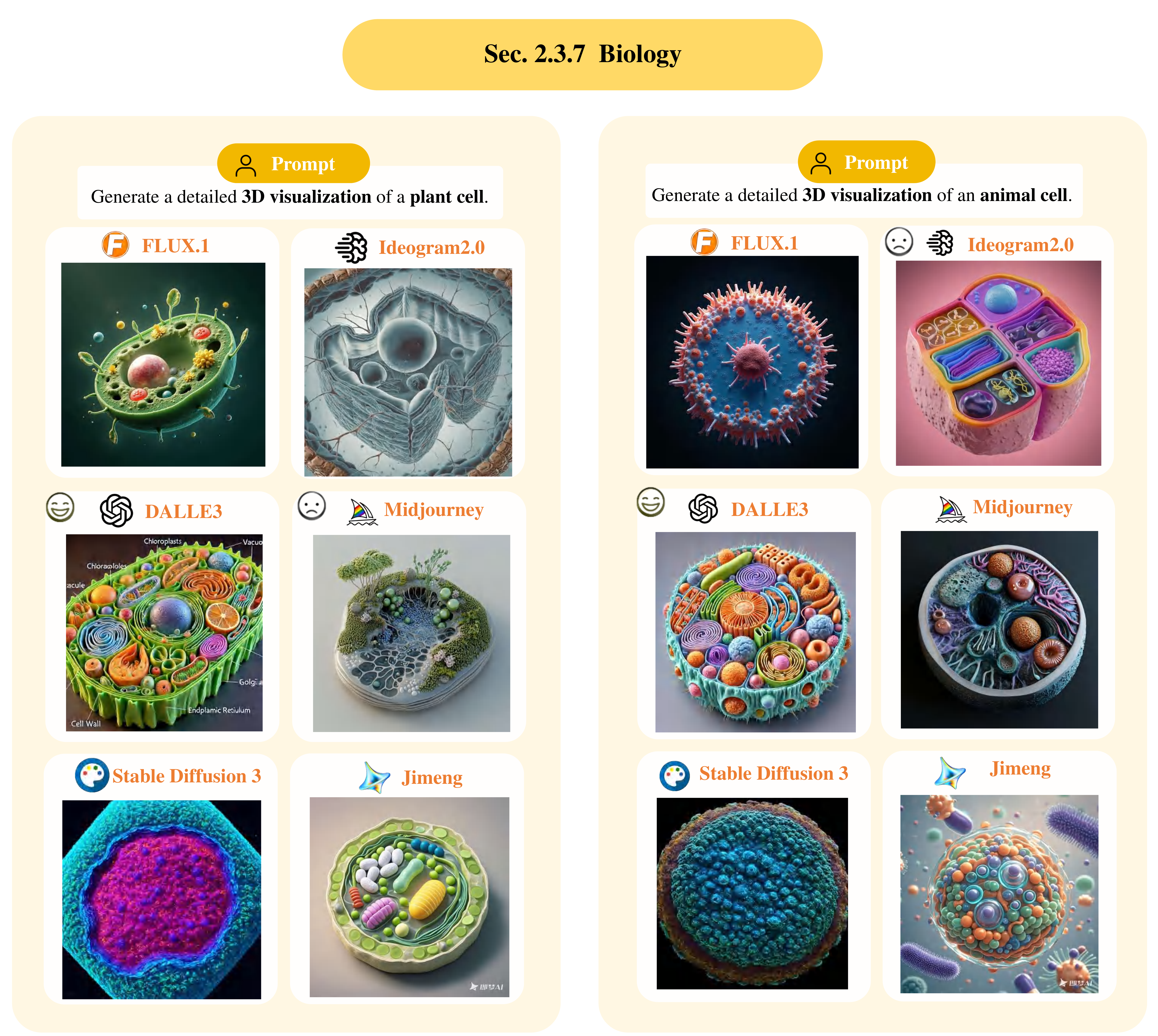}}
  \caption[Section~\ref{bio}: biology.]{Results on biology task. Refer to Section \ref{bio} for detailed discussions.}
  \label{biology1}
\end{figure*}

 Figure \ref{biology1} focuses on their grasp of biology, specifically through the generation of plant cell and animal cell images. In the first example, FLUX.1, Dall-E3, and Jimeng attempt to depict plant cell structures. However, FLUX.1 erroneously includes leaves inside the cell, while Dall-E3 incorporates an orange slice incorrectly. Meanwhile, Ideogram2.0 and Stable Diffusion 3 offer microscopic views of plant cells, but Midjourney seems to lack sufficient biological knowledge. The results of the second example are similarly unsatisfactory. 

\textbf{Score. }
The results of this experiment are shown in Table \ref{tab_bio}. FLUX.1 and Dall-E3 performed relatively well in this task; however, only the CLIPScore results aligned with human intuitive perception.
\begin{table}[h]
    \centering
    \caption[Section~\ref{bio}: bio.]{The scoring of generation results by six models on biology under different evaluation systems. Refer to Section \ref{bio} for detailed discussions.}
    \begin{tabular}{l|c|c|c|c|c}
        \midrule
        Model & CLIPScore & HPSv2 & Aesthetic Score &  GPT-4o & Human \\
        \midrule
        FLUX.1      & \textbf{30.54} & 0.27 & 5.77 &  4.58 & \textbf{9.17} \\
        Ideogram2.0 & 26.10 & 0.22 & 5.78 & 5.42 & 6.25 \\
        Dall-E3     & 27.58 & 0.26 & 5.56 &  6.25 & 7.50 \\
        Midjourney  & 21.99 & 0.23 & 5.79 &  7.08 & 5.00 \\
        SD3         & 26.85 & \textbf{0.28} & \textbf{5.99} & 3.75 & 5.84 \\
        Jimeng      & 23.14 & 0.25 & 5.82 & \textbf{7.50} & 7.92 \\
        \midrule
    \end{tabular}
    \label{tab_bio}
\end{table}
 
\subsubsection{Robotics and Simulation Tasks}
\label{robotics}
\begin{figure*}[!ht]
  \centering 
  \makebox[\textwidth][c]{\includegraphics[width=1\textwidth]{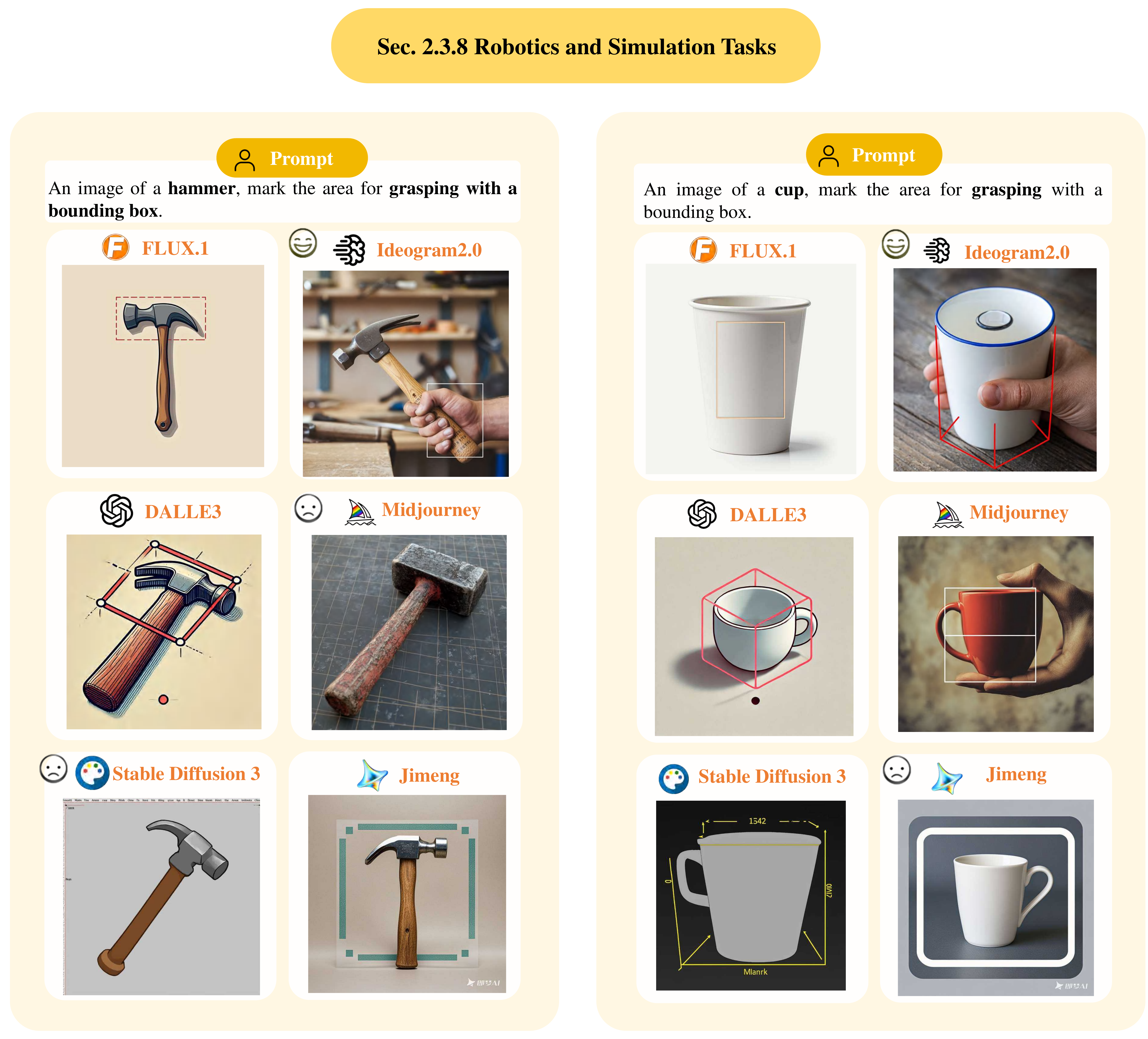}}
  \caption[Section~\ref{robotics}: robotics and simulation tasks.]{Results on robotics and simulation task. Refer to Section \ref{robotics} for detailed discussions.}
  \label{fig_toolusage}
\end{figure*}

\textbf{Tool-Usage Grounding}
In Robotics, the model needs to recognize the area to grasp or manipulate~\cite{Gu2023SeerLI}. To test whether T2I models have the tool-usage grounding ability, we prompt them using two examples. For the first prompt, as depicted in the left subplot of Figure \ref{fig_toolusage}, only Ideogram2.0 can generate the hammer image with the grasping area marked with a bounding box. Dall-E3 and FLUX.1 also generate the bounding box in the image, but it's placed in the wrong area. For the second example, Stable Diffusion 3, Midjourney, Dall-E3 and Ideagram2.0 all generate the bounding box in the image, but only Ideogram2.0 places it on the grasping area.

\begin{figure*}[!ht]
  \centering 
  \makebox[\textwidth][c]{\includegraphics[width=1\textwidth]{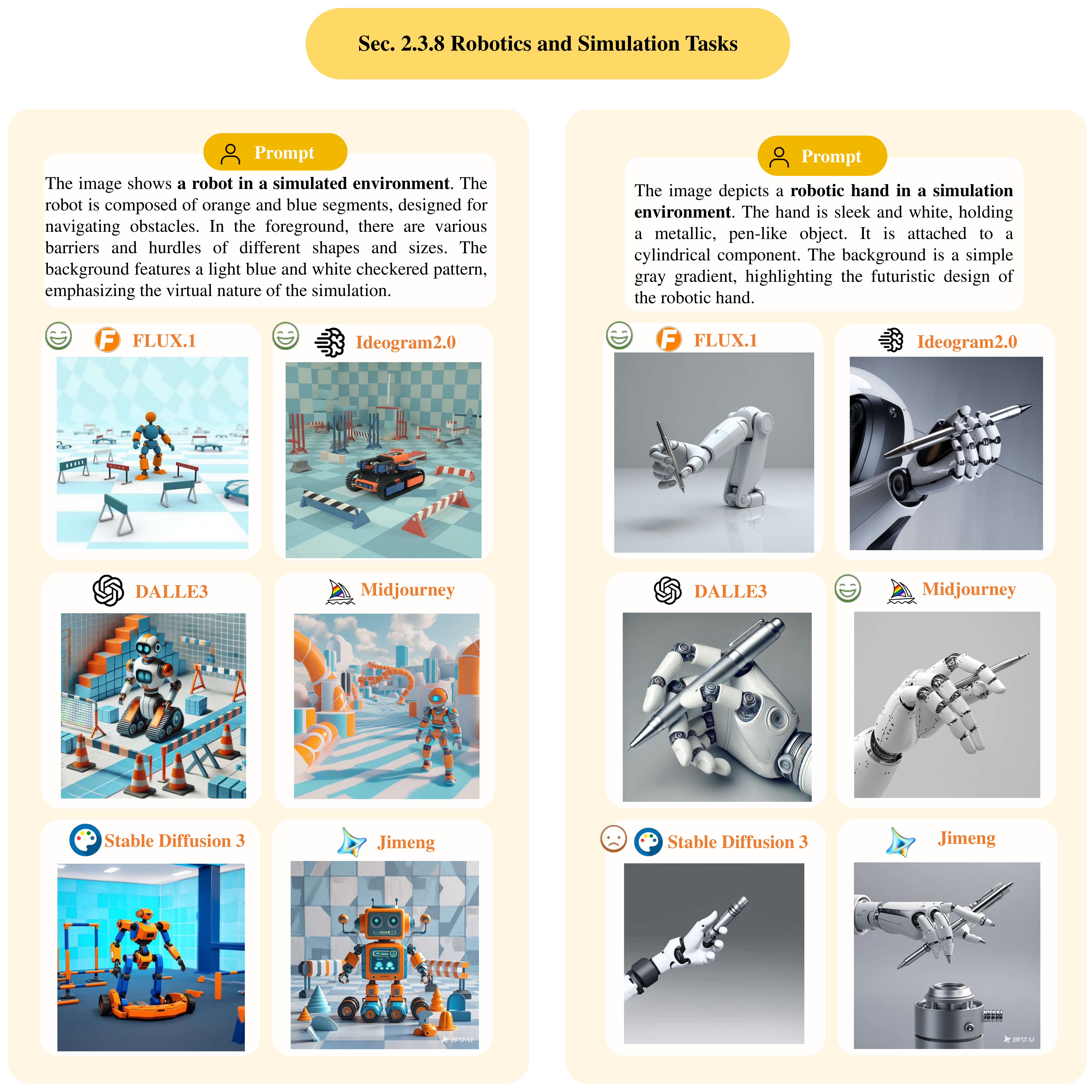}}
  \caption[Section~\ref{robotics}: robotics and simulation tasks. ]{Results on robotics and simulation task. Refer to Section \ref{robotics} for detailed discussions.}
  \label{fig_simulation}
\end{figure*}

\textbf{Simulation environment}
Lots of robotic research is conducted in the simulated environment. Thus it's important to test T2I models' ability to produce the images in the simulation environment domain. Here we give two simulations, one is a robot navigating, and the other is a dexterous hand. The results are shown in the right subplot of Figure \ref{fig_simulation}. For the first simulation, only the images from Ideogram2.0 and FLUX.1 look like the rendering image of the simulation environment. For the second simulation, only FLUX.1 correctly generated the dexterous hand.

\textbf{Score. }
The results of this experiment are shown in the Table \ref{tab_robotics}. In this experiment, FLUX.1 achieved a relatively high score, and GPT-4o's evaluation results were more consistent with human intuition.
\begin{table}[h]
    \centering
    \caption[Section~\ref{robotics}: robotics and simulation tasks.]{The scoring of generation results by six models on robotics and simulation tasks under different evaluation systems. Refer to Section \ref{robotics} for detailed discussions.}
    \begin{tabular}{l|c|c|c|c|c}
        \midrule
        Model & CLIPScore & HPSv2 & Aesthetic Score & GPT-4o & Human \\
        \midrule 
        FLUX.1      & 30.93 & 0.28 & 5.41 & \textbf{7.09} & \textbf{9.17} \\
        Ideogram2.0 & 30.59 & 0.27 & 5.31 & 5.83 & 8.75 \\
        Dall-E3     & \textbf{30.94} & \textbf{0.30} & 5.32 & 4.58 & 7.92 \\
        Midjourney  & 29.05 & 0.29 & 5.71 & 5.63 & 6.67 \\
        SD3         & 28.11 & 0.27 & 5.34 & 4.79 & 6.04 \\
        Jimeng      & 30.10 & 0.28 & \textbf{5.81} & 5.63 & 7.08 \\
        \midrule
    \end{tabular}
    \label{tab_robotics}
\end{table}

\subsubsection{Autonomous Driving}
\label{autodrive}
For autonomous driving model development, synthetic data plays an important role in solving the corner cases. For example, in the real world, it is hard to collect a large amount of autonomous driving data under extreme weather. Utilizing the T2I model, we can synthesize these hard-to-collect data efficiently. Here we prompt these T2I models to generate the road scene under rainy weather and foggy weather respectively.

\begin{figure*}[!ht]
\vspace{-5em}
  \centering 
  \makebox[\textwidth][c]{\includegraphics[width=1\textwidth]{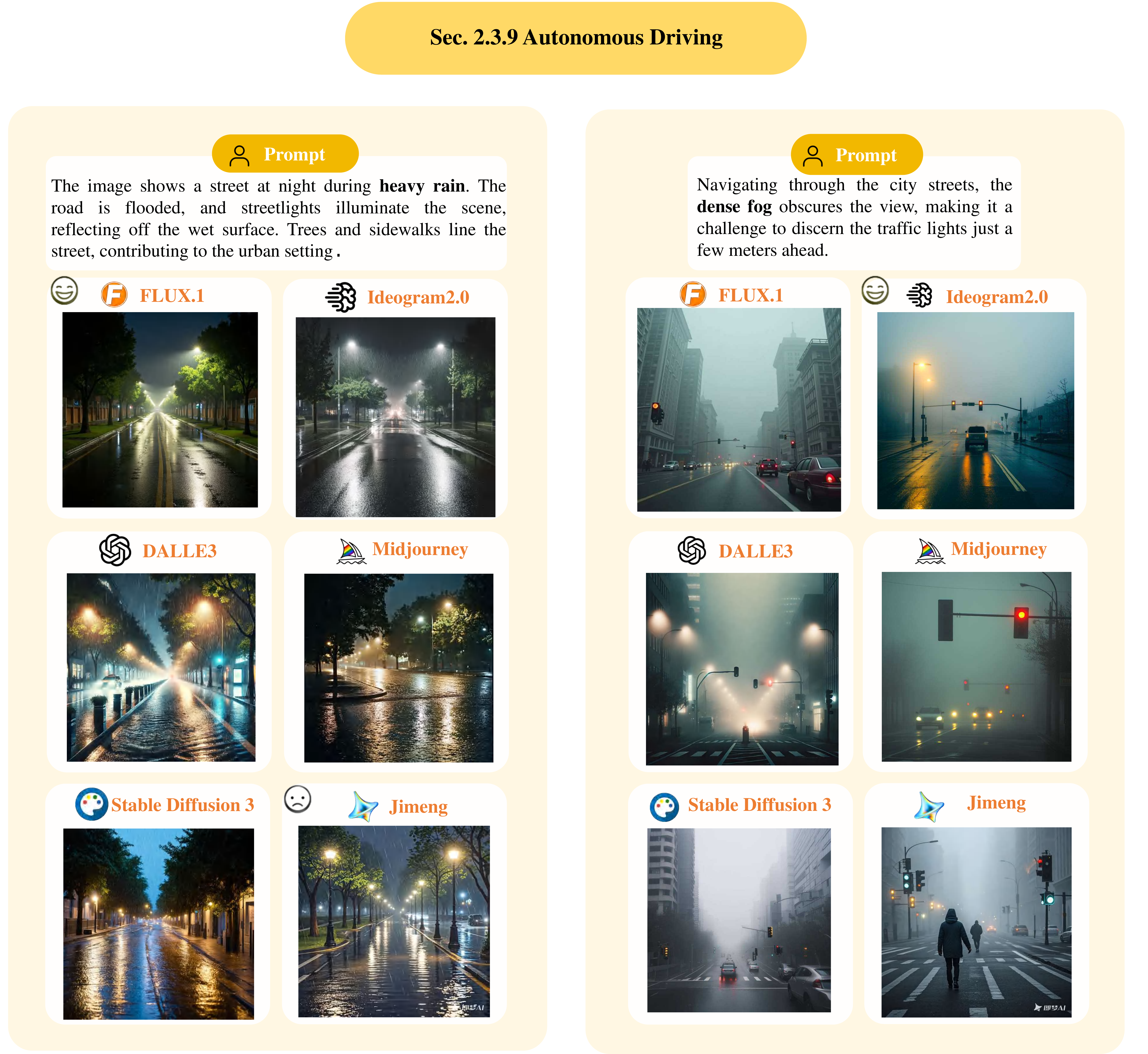}}
  \caption[Section~\ref{autodrive}: autonomous driving task.]{Results on autonomous drive corner cases task. Refer to Section \ref{autodrive} for detailed discussions.}
  \label{fig_autodrive}
\end{figure*}

The results are shown in Figure \ref{fig_autodrive}. For the road scene under rainy weather, all six models perform well, while images generated by Dall-E3 and Jimeng are not as realistic as the others. For the road scene under foggy weather, all six models perform well, except the Jimeng-generated image contains commonsense artifacts -- pedestrians won't walk in the middle of the road.

\textbf{Score. }
The results of this experiment are shown in Table \ref{tab_autodrive}. Stable Diffusion 3 and Jimeng scored higher in other evaluation systems, but there is a significant gap compared to human ratings. This may be due to the fact that the images in this experiment often lack realism in multiple details, which the evaluation systems fail to accurately recognize.

\begin{table}[h]
    \centering
    \caption[Section~\ref{autodrive}: autonomous drive corner cases.]{The scoring of generation results by six models on autonomous driving task under different evaluation systems. Refer to Section \ref{autodrive} for detailed discussions.}
    \begin{tabular}{l|c|c|c|c|c}
        \midrule
        Model & CLIPScore & HPSv2 & Aesthetic Score & GPT-4o & Human \\
        \midrule 
       FLUX.1      & 29.25 & 0.26 & 5.45 &    6.67& 8.34 \\
        Ideogram2.0 & 28.42 & 0.29 & 5.90 &   6.67 & 8.34 \\
        Dall-E3     & 26.75 & 0.26 & 5.35 &   6.67 & 8.33 \\
        Midjourney  & 28.01 & 0.26 & 5.27 &   6.25 & \textbf{9.17} \\
        SD3         & \textbf{30.21} & 0.26 & 5.89 &   \textbf{8.33}& 7.50 \\
        Jimeng      & 29.17 & \textbf{0.30} & \textbf{6.02} &   7.08 & 6.67 \\
        \midrule
    \end{tabular}
    \label{tab_autodrive}
\end{table}

\clearpage
\subsection{Challenging Scenario Generation}
\label{sec:05challenging}
The application of T2I models is expanding, particularly in generating images for more challenging scenarios. In this section, we carefully curate a set of complex prompts to evaluate the models' ability to handle intricate settings.

In Section \ref{mark}-\ref{setofmark}, we prompt the models to generate images based on traditional visual task settings. Section \ref{denseocr} examines the models' ability to generate images containing dense text and pictures. In Section \ref{multilingual}, we assess the models' understanding of different languages from various regions. Finally, in Section \ref{emoji2image}, we input emojis into the models to evaluate their ability to organize diverse elements and piece together the logical relationships between multiple emojis in an image. In Section \ref{irrational}, we explored the ability of T2I models to generate images in irrational scenarios. In Section \ref{qa}, we investigated whether T2I models could generate corresponding images based on answers when given simple questions as input. In Section \ref{watermark}, we experimented with the ability of T2I models to generate images with watermarks. In Section \ref{ood}, we explored the model's ability to generate low-quality images, which indicates whether the model's training data contains low-quality data. In Section \ref{multi-image}, we systematically explored the model's ability to generate images containing multiple sub-images. In Section \ref{text}, we specifically studied the model's ability to generate text within images. 

\subsubsection{Image with Mark}
\label{mark}
\begin{figure*}[!ht]
  \centering 
  \makebox[\textwidth][c]{\includegraphics[width=0.8\textwidth]{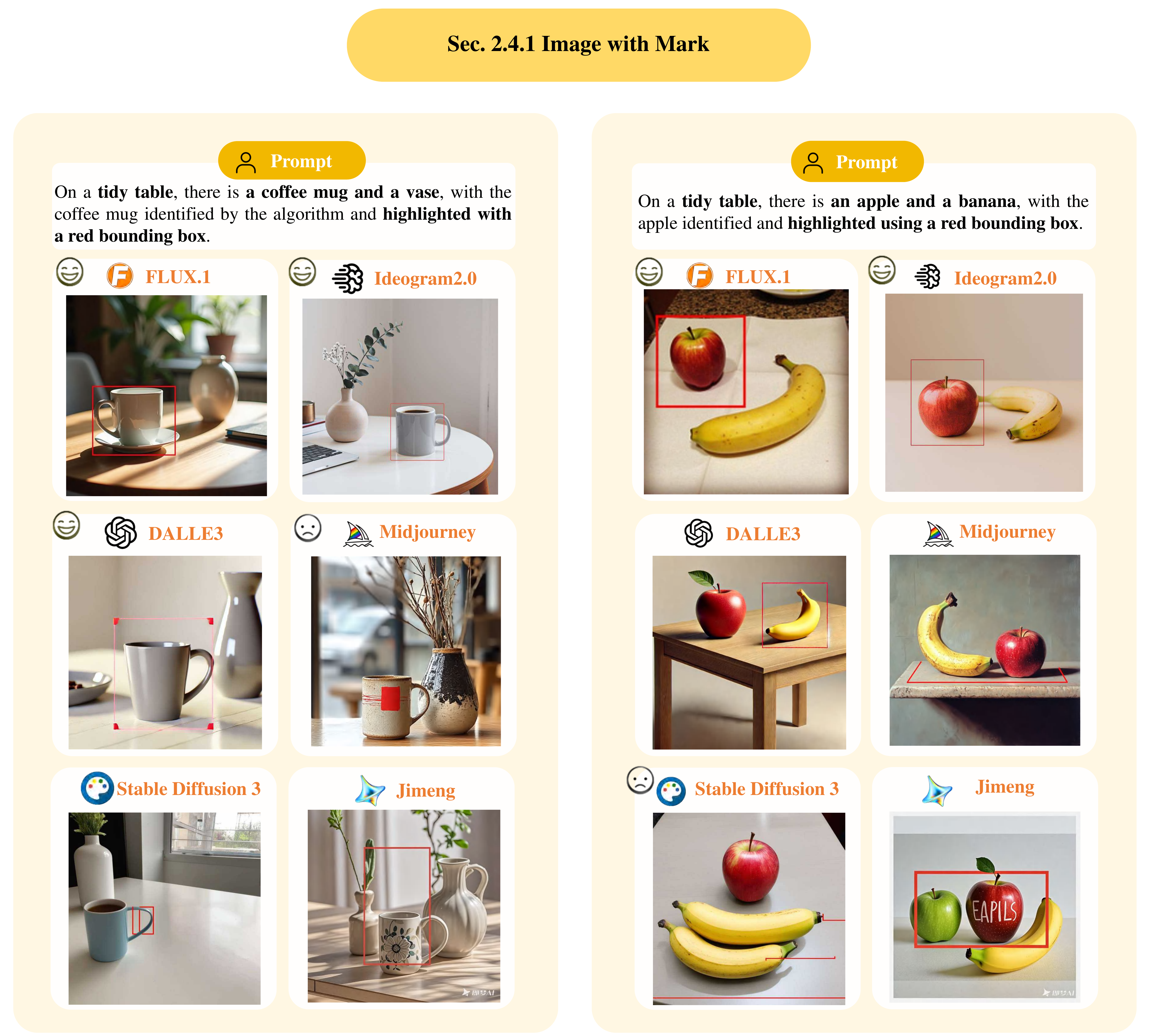}}
  \caption[Section~\ref{mark}: image with mark: single mark. ]{Results on image with mark task. Refer to Section \ref{mark} for detailed discussions.}
  \label{fig_mark-1}
\end{figure*}

Object detection is a common task in the field of computer vision~\cite{Liu2023ContinualDT,Chen2019DetNASBS,Peng2025MutualForceME,Mekhalfi2025LeveragingCI,Miwa2025OneDPieceIT}. In this paper, we design the image with mark task to explore whether text-to-image models can be applied to computer vision tasks.
In this task, we examine the models' ability to generate images resembling the output of computer vision algorithms. The results are presented in Figure \ref{fig_mark-1}. In the first prompt, we asked the models to highlight a coffee mug with a bounding box, and in the second prompt, to highlight an apple in the same manner. We observed that only FLUX.1~\cite{flux2024}, Ideogram2.0~\cite{ideogram2.0}, and Dall-E3~\cite{betker2023improving} successfully completed the first task, while only FLUX.1 and Ideogram2.0 correctly accomplished the second task.

\textbf{Score. }
The results of this experiment are shown in the Table \ref{tab_mark}. It can be observed that the outputs of FLUX.1 and Ideogram2.0 are relatively good. The same conclusion can be drawn from the other scores, except for the CLIPScore.

\begin{table}[h]
    \centering
    \caption[Section~\ref{mark}: image with mark.]{The scoring of generation results by six models on image with mark under different evaluation systems. Refer to Section \ref{mark} for detailed discussions.}
    \begin{tabular}{l|c|c|c|c|c}
        \midrule
        Model & CLIPScore & HPSv2 & Aesthetic Score & GPT-4o & Human \\
        \midrule
        FLUX.1      & 29.66 & 0.25 & 5.13 &   \textbf{8.74} & \textbf{10.00} \\
        Ideogram2.0 & 32.27 & \textbf{0.27} & \textbf{5.94} &   4.59 & \textbf{10.00} \\
        Dall-E3     & \textbf{36.37} & \textbf{0.27} & 4.94 &   7.90 & 9.17 \\
        Midjourney  & 31.12 & \textbf{0.27} & 4.90 &   7.07 & 6.67 \\
        SD3         & 34.43 & 0.26 & 5.19 &   4.59 & 7.50 \\
        Jimeng      & 32.04 & 0.26 & 4.71 &   5.00 & 8.33 \\
        \midrule
    \end{tabular}
    \label{tab_mark}
\end{table}

\subsubsection{Set of Mark}
\label{setofmark}
\begin{figure*}[!ht]
  \centering 
  \makebox[\textwidth][c]{\includegraphics[width=0.9\textwidth]{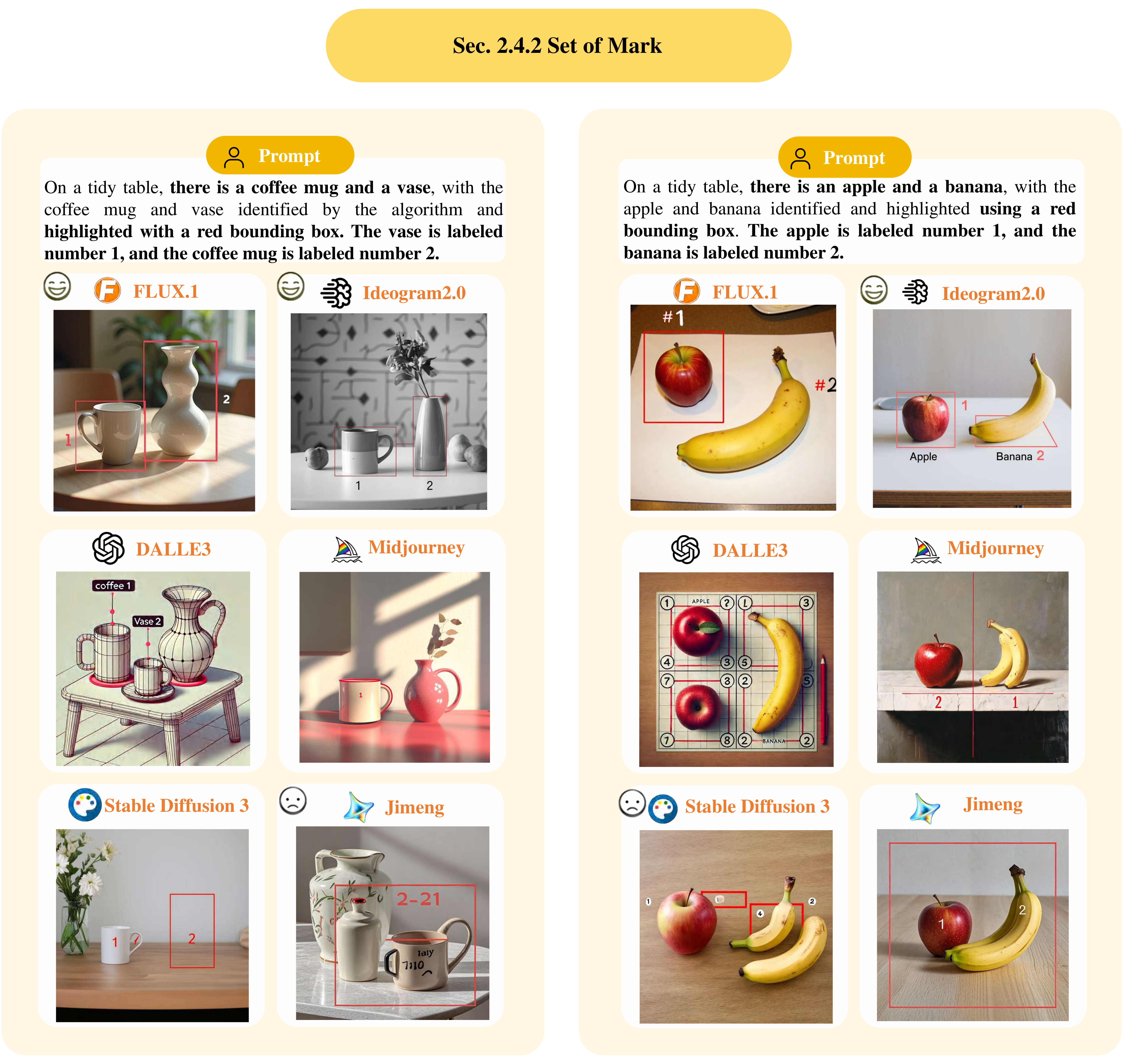}}
  \caption[Section~\ref{setofmark}: image with mark: set of mark. ]{Results on set of mark task. Refer to Section \ref{setofmark} for detailed discussions.}
  \label{fig_mark-2}
\end{figure*}

Set of mark is also a traditional task in computer vision~\cite{Yang2023SetofMarkPU}.
As depicted in the figure \ref{fig_mark-2}, FLUX accurately generates red bounding boxes around the specified coffee cup and vase, although the serial numbers are incorrect. In contrast, Midjourney fails to produce regular rectangles, while Stable Diffusion 3~\cite{rombach2022high} misplaces the bounding box, entirely missing the vase, and also assigns incorrect serial numbers. Jimeng, however, successfully frames all required objects. These findings indicate that while FLUX.1 exhibits minor labeling inaccuracies, it holds promise for improving object detection and grounding tasks in subsequent processes.

\textbf{Score. }
The results of this experiment are shown in the Table \ref{tab_setofmark}. From a human intuitive perspective, the output of FLUX.1 is the best. Only the results from HPSv2 align closely with human intuition, while the scores from the other metrics show some discrepancies.
\begin{table}[h]
    \centering
    \caption[Section~\ref{setofmark}: set of mark.]{The scoring of generation results by six models on set of mark task under different evaluation systems. Refer to Section \ref{setofmark} for detailed discussions.}
    \begin{tabular}{l|c|c|c|c|c}
        \midrule
        Model & CLIPScore & HPSv2 & Aesthetic Score &  GPT-4o & Human \\
        \midrule
        FLUX.1      & 33.72 & \textbf{0.27} & 5.24 &   6.94 & \textbf{9.22} \\
        Ideogram2.0 & 34.93 & 0.25 & 4.87 &   6.94 & 8.89 \\
        Dall-E3     & 33.30 & 0.24 & 5.20 &   3.61 & 7.50 \\
        Midjourney  & 32.04 & 0.25 & \textbf{5.80} &   \textbf{6.95} & 5.56 \\
        SD3         & \textbf{37.93} & 0.25 & 5.18 &   5.83 & 7.22 \\
        Jimeng      & 35.27 & 0.24 & 5.19 &   3.89 & 5.83 \\
        \midrule
    \end{tabular}
    \label{tab_setofmark}
\end{table}

\subsubsection{Multilingual}
\label{multilingual}
\begin{figure*}[!ht]
  \centering 
  \makebox[\textwidth][c]{\includegraphics[width=0.8\textwidth]{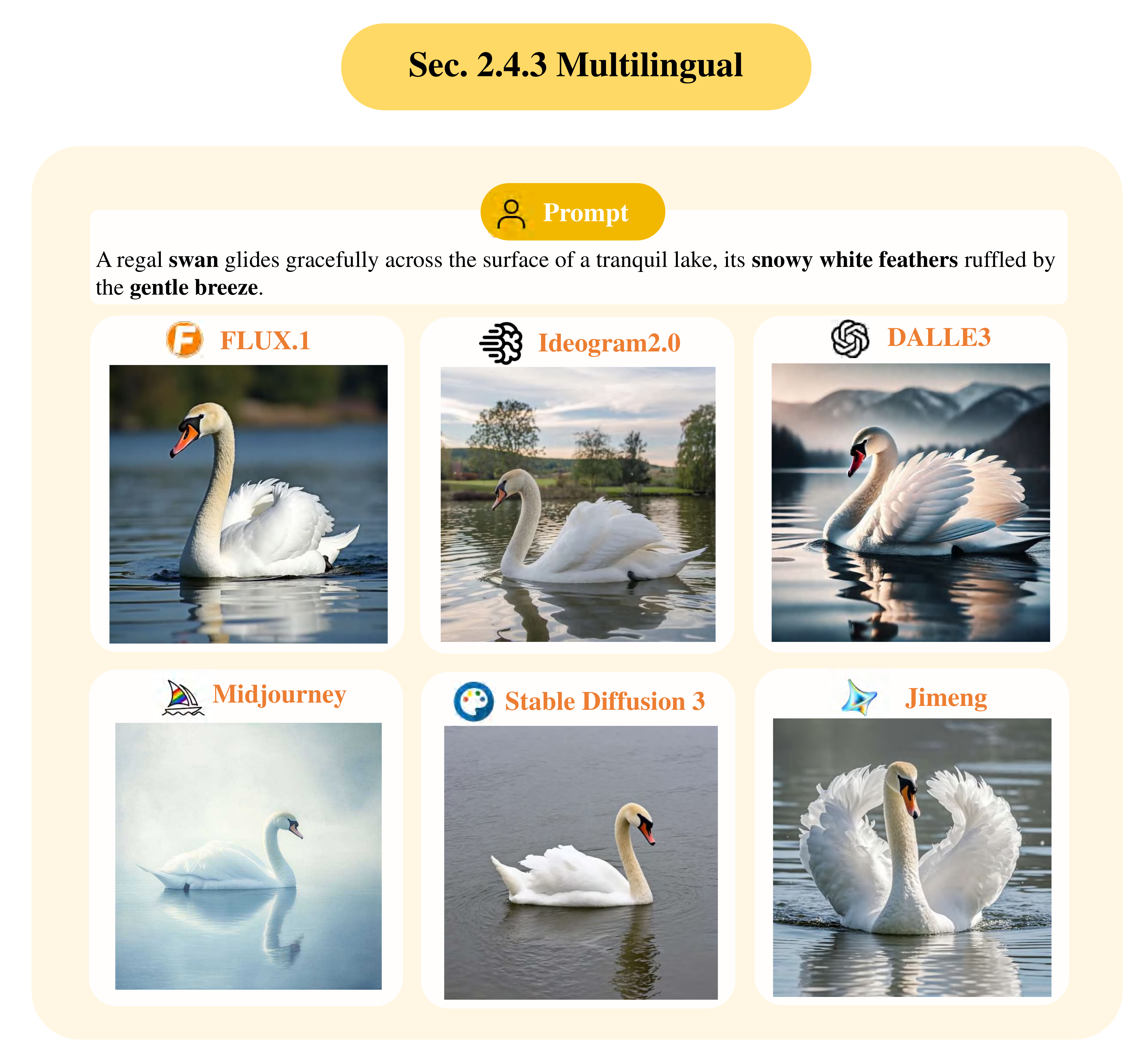}}
  \caption[Section~\ref{multilingual}: multilingual]{Results on multilingual task. Refer to Section \ref{multilingual} for detailed discussions.}
  \label{fig_multilingual_1}
\end{figure*}

\begin{figure*}[!ht]
  \centering 
  \makebox[\textwidth][c]{\includegraphics[width=0.75\textwidth]{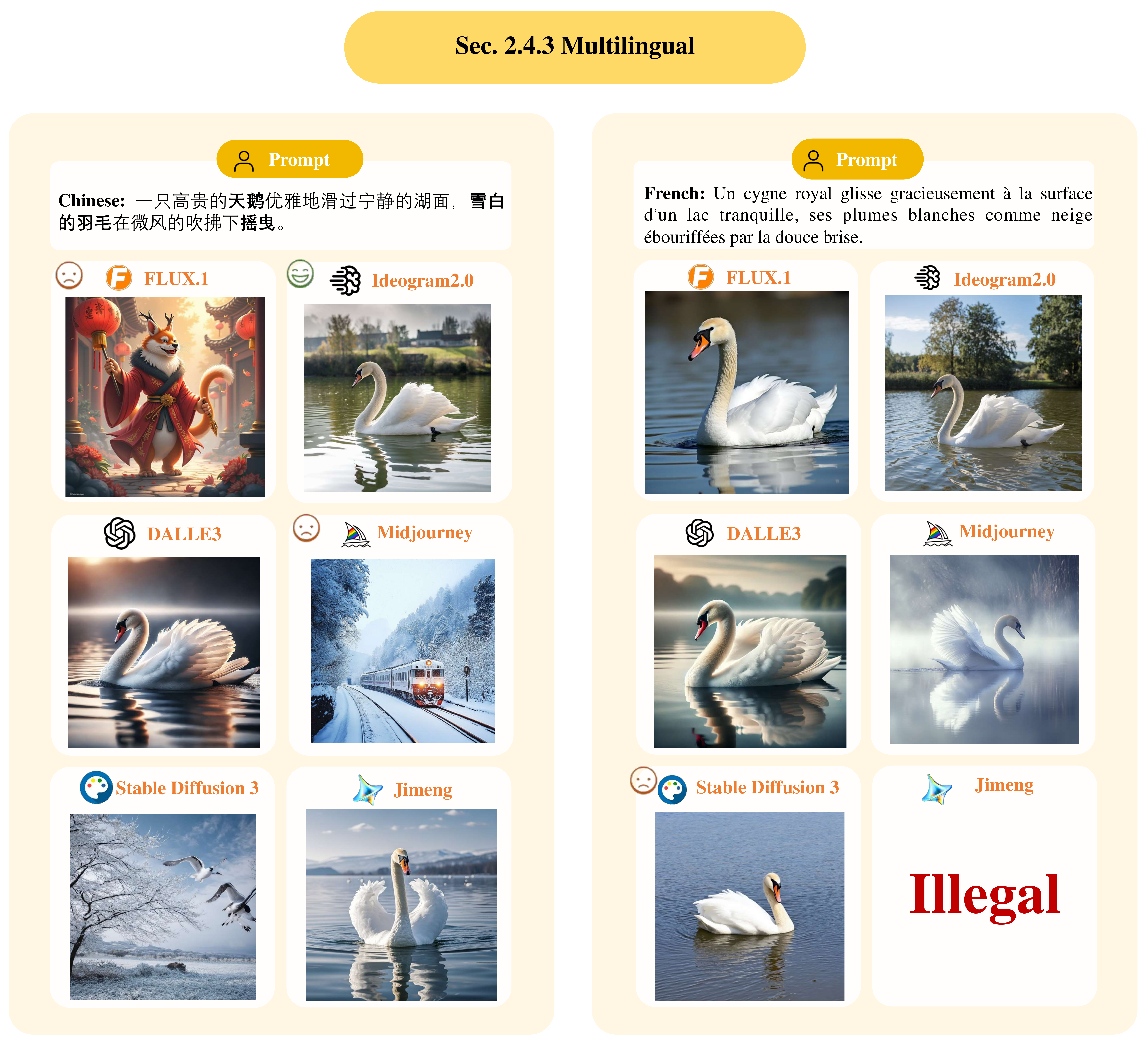}}
  \caption[Section~\ref{multilingual}: multilingual]{Results on multilingual task. Refer to Section \ref{multilingual} for detailed discussions.}
  \label{fig_multilingual_2}
\end{figure*}

\begin{figure*}[!ht]
  \centering 
  \makebox[\textwidth][c]{\includegraphics[width=0.75\textwidth]{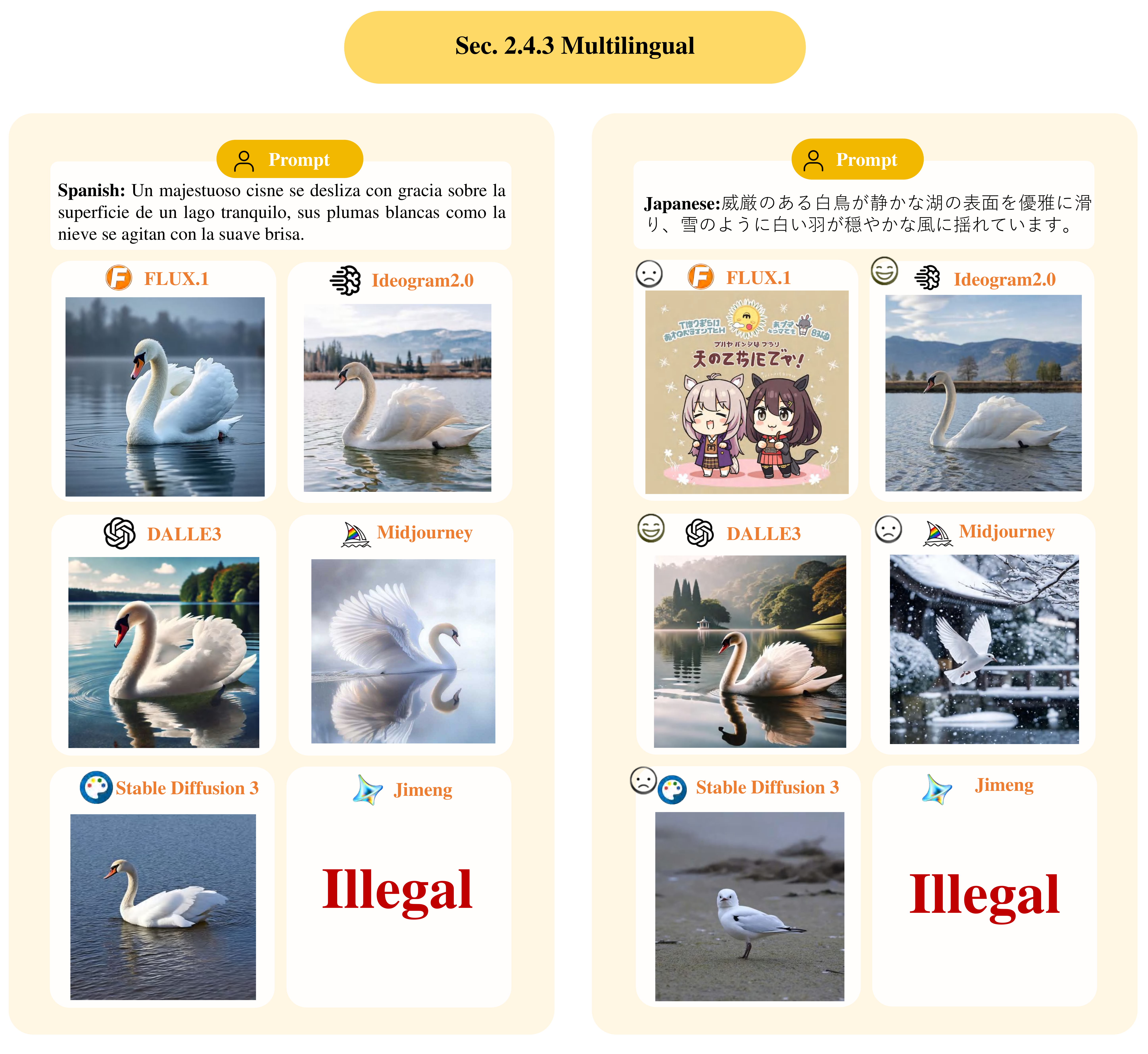}}
  \caption[Section~\ref{multilingual}: multilingual]{Results on multilingual task. Refer to Section \ref{multilingual} for detailed discussions.}
  \label{fig_multilingual_3}
\end{figure*}

As shown in Figures~\ref{fig_multilingual_1}-\ref{fig_multilingual_3}, the multilingual capabilities of T2I models exhibit significant variations across different languages~\cite{Mu2025BoostingTG,Xing2024MuLanAM}. Models like Ideogram2.0 and Dall-E3 demonstrate strong performance when processing prompts in English, Spanish, and French. However, a notable limitation remains: FLUX.1 performs poorly with Chinese prompts, while FLUX.1, Midjourney, and Stable Diffusion 3 show subpar results with Japanese prompts. This may be attributed to their use of text encoders that support only English, highlighting a crucial area for improvement in the development of more universally robust multilingual T2I models.

\textbf{Score. }
The results of this experiment are shown in the Table \ref{tab_multilingual}. The results for CLIPScore and Aesthetic Score are relatively consistent with human ratings.
\begin{table}[h]
    \centering
    \caption[Section~\ref{multilingual}: multilingual.]{The scoring of generation results by six models on multilingual under different evaluation systems. Refer to Section \ref{multilingual} for detailed discussions.}
    \begin{tabular}{l|c|c|c|c|c}
        \midrule
        Model & CLIPScore & HPSv2 & Aesthetic Score &GPT-4o & Human \\
        \midrule
        FLUX.1      & 23.40 & 0.28 & 5.62 &   6.94 & 8.33 \\
        Ideogram2.0 & \textbf{25.75} & 0.29 & \textbf{6.06} &  7.50 & \textbf{10.00} \\
        Dall-E3     & 23.29 & \textbf{0.30} & 5.98 &   8.05 & 8.33 \\
        Midjourney  & 23.26 & \textbf{0.30} & 6.16 &   6.94 & 7.78 \\
        SD3         & 25.19 & 0.27 & 5.93 &   5.55 & 8.33 \\
        Jimeng      & 21.77 & 0.29 & 5.88 &   \textbf{8.75} & 9.17 \\
        \midrule
    \end{tabular}
    \label{tab_multilingual}
\end{table}

\subsubsection{Dense OCR}
\label{denseocr}
\begin{figure*}[!ht]
  \centering 
  \makebox[\textwidth][c]{\includegraphics[width=0.75\textwidth]{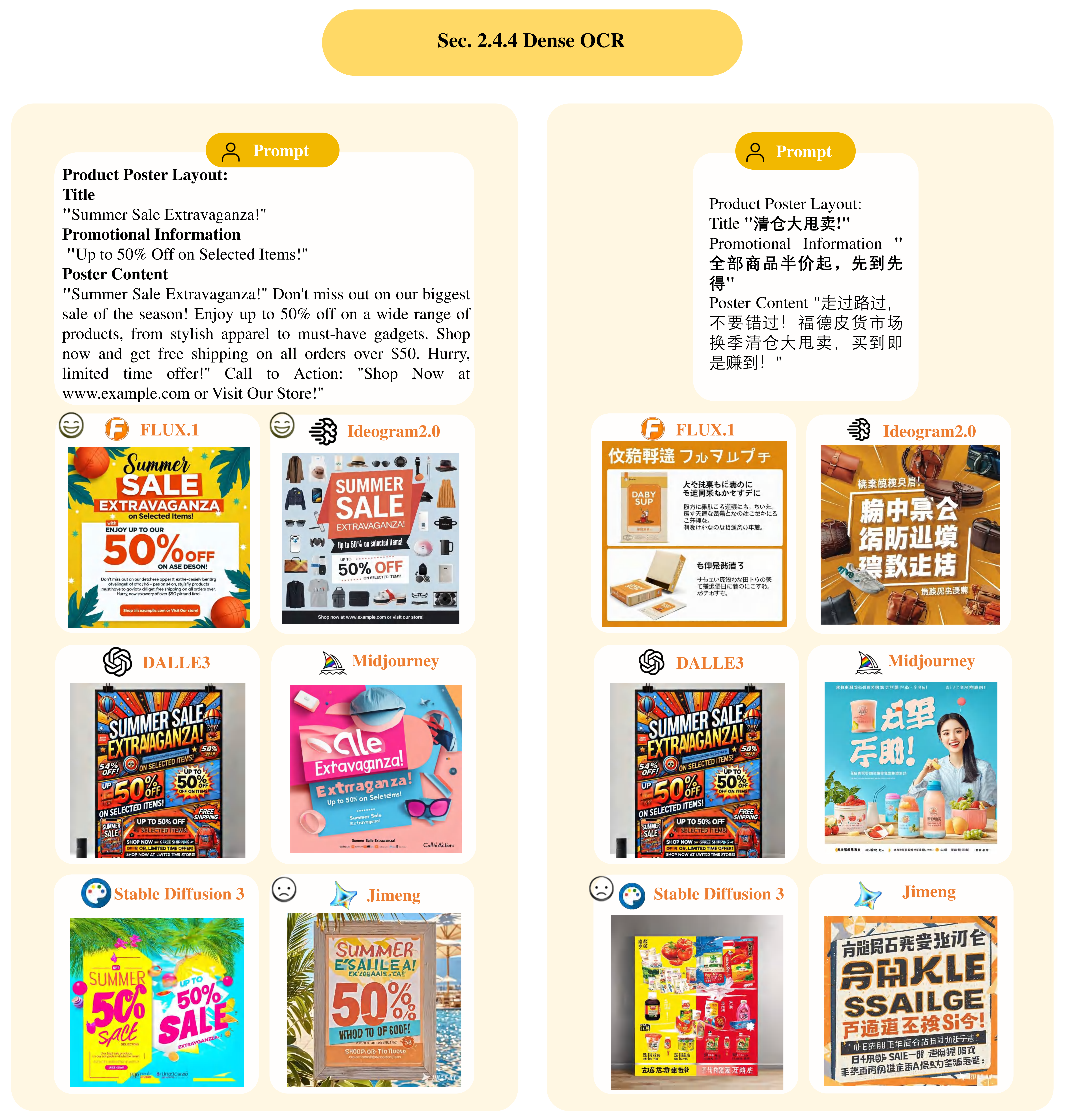}}
  \caption[Section~\ref{denseocr}: dense OCR]{Results on denseocr task. Refer to Section \ref{denseocr} for detailed discussions.}
  \label{fig_denseocr1}
\end{figure*}

\begin{figure*}[!ht]
  \centering 
  \makebox[\textwidth][c]{\includegraphics[width=0.8\textwidth]{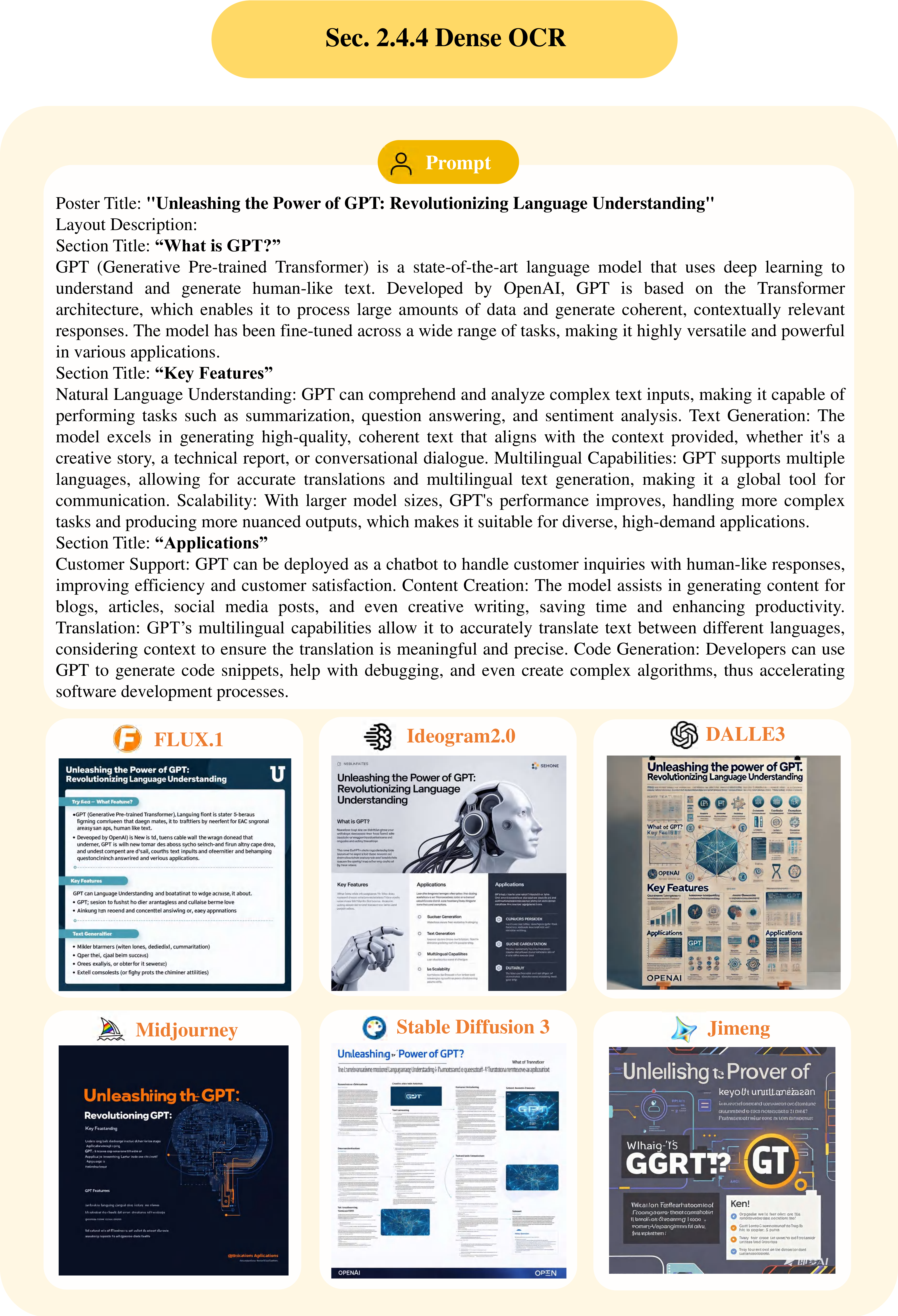}}
  \caption[Section~\ref{denseocr}: dense OCR]{Results on denseocr task. Refer to Section \ref{denseocr} for detailed discussions.}
  \label{fig_denseocr2}
\end{figure*}

Figure~\ref{fig_denseocr1} and Figure~\ref{fig_denseocr2} present an evaluation of the dense OCR capabilities of various T2I models. When generating posters with an English corpus, FLUX.1 successfully captures the overall content based on the given requirements but exhibits some spelling errors in the generated text. In contrast, Jimeng, Dall-E3, Ideogram2.0, and Stable Diffusion 3 focus primarily on the title, failing to generate additional textual content from the provided prompts. Notably, Stable Diffusion 3 introduces considerable spelling errors. Furthermore, none of these T2I models effectively recognize or generate Chinese text when tasked with poster generation using a Chinese corpus, highlighting a significant limitation in handling Chinese OCR.
For academic paper poster generation, FLUX.1 and Ideogram2.0 demonstrate the ability to generate most of the textual content with a clear and legible appearance. However, Dall-E3, Stable Diffusion 3, Jimeng, and Midjourney struggle with text clarity and exhibit prominent spelling errors, indicating limitations in generating accurate and coherent textual content in this context.

\textbf{Score. }
The results of this experiment are shown in the Table \ref{tab_denseocr}. From a human intuitive perspective, Ideogram2.0 performed the best in this experiment, with only the GPT-4o results aligning closely with human perception.

\begin{table}[h]
    \centering
    \caption[Section~\ref{denseocr}: dense OCR.]{The scoring of generation results by six models on dense OCR under different evaluation systems. Refer to Section \ref{denseocr} for detailed discussions.}
    \begin{tabular}{l|c|c|c|c|c}
        \midrule
        Model & CLIPScore & HPSv2 & Aesthetic Score  & GPT-4o & Human \\
        \midrule
        FLUX.1      & 32.22 & \textbf{0.25} & 4.22 &   1.67 & 7.22 \\
        Ideogram2.0 & 29.27 & \textbf{0.25} & 4.64 &   \textbf{5.83} & \textbf{7.78} \\
        Dall-E3     & 31.07 & \textbf{0.25} & \textbf{4.73} &   5.56 & 5.83 \\
        Midjourney  & 28.98 & 0.22 & 4.10 &   3.33 & 6.94 \\
        SD3         & \textbf{33.01} & \textbf{0.25} & 4.60 &   3.89 & 5.83 \\
        Jimeng      & 23.89 & 0.21 & 3.86 &   4.17 & 5.56 \\
        \midrule
    \end{tabular}
    \label{tab_denseocr}
\end{table}

\subsubsection{Emoji}
\label{emoji2image}
\begin{figure*}[!ht]
\vspace{1em}
  \centering 
  \makebox[\textwidth][c]{\includegraphics[width=1\textwidth]{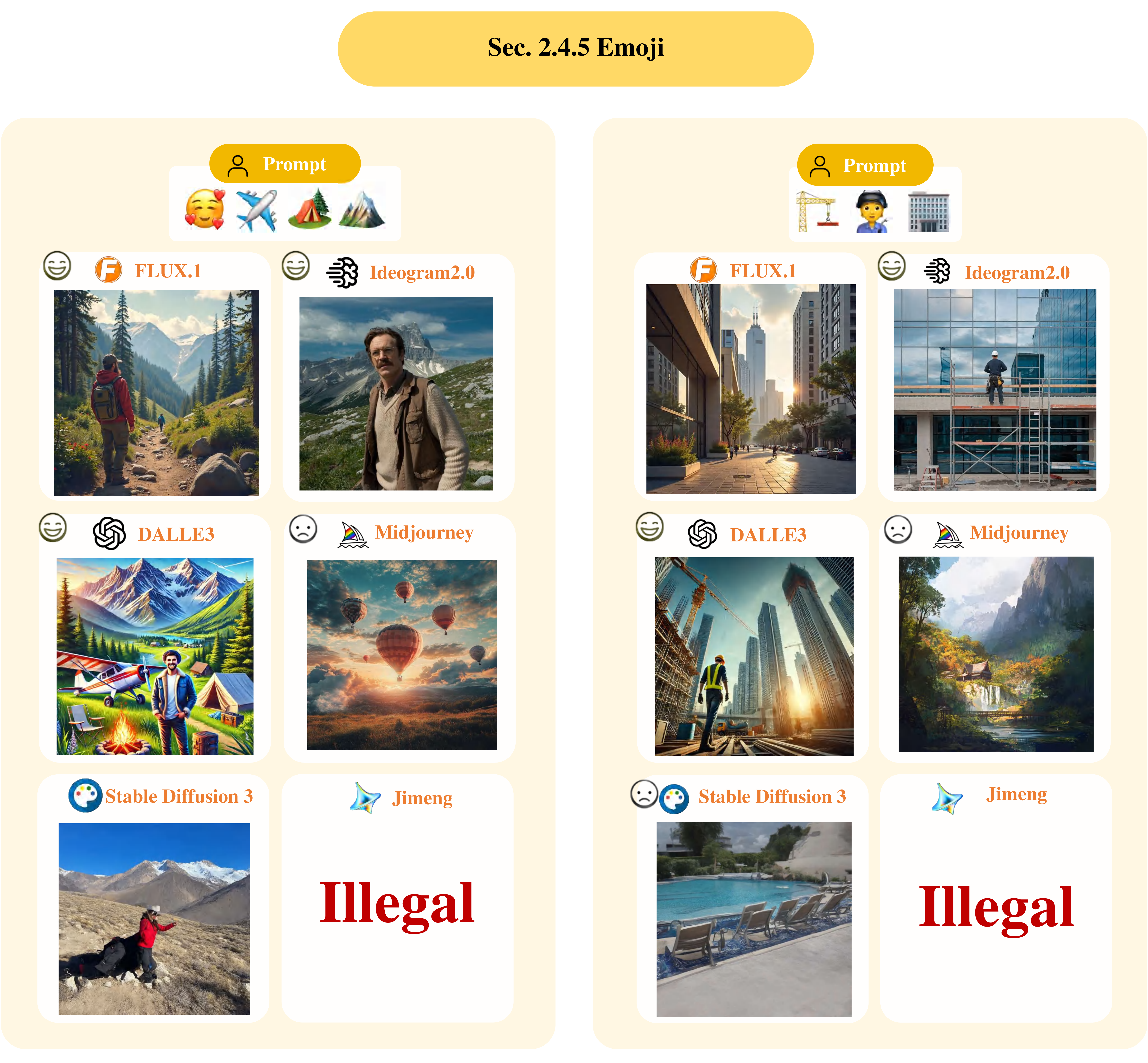}}
  \caption[Section~\ref{emoji2image}: emoji.]{Results on emoji task. Refer to Section \ref{emoji2image} for detailed discussions.}
  \label{fig_emojis1}
\end{figure*}

\begin{figure*}[!ht]
  \centering 
  \makebox[\textwidth][c]{\includegraphics[width=1\textwidth]{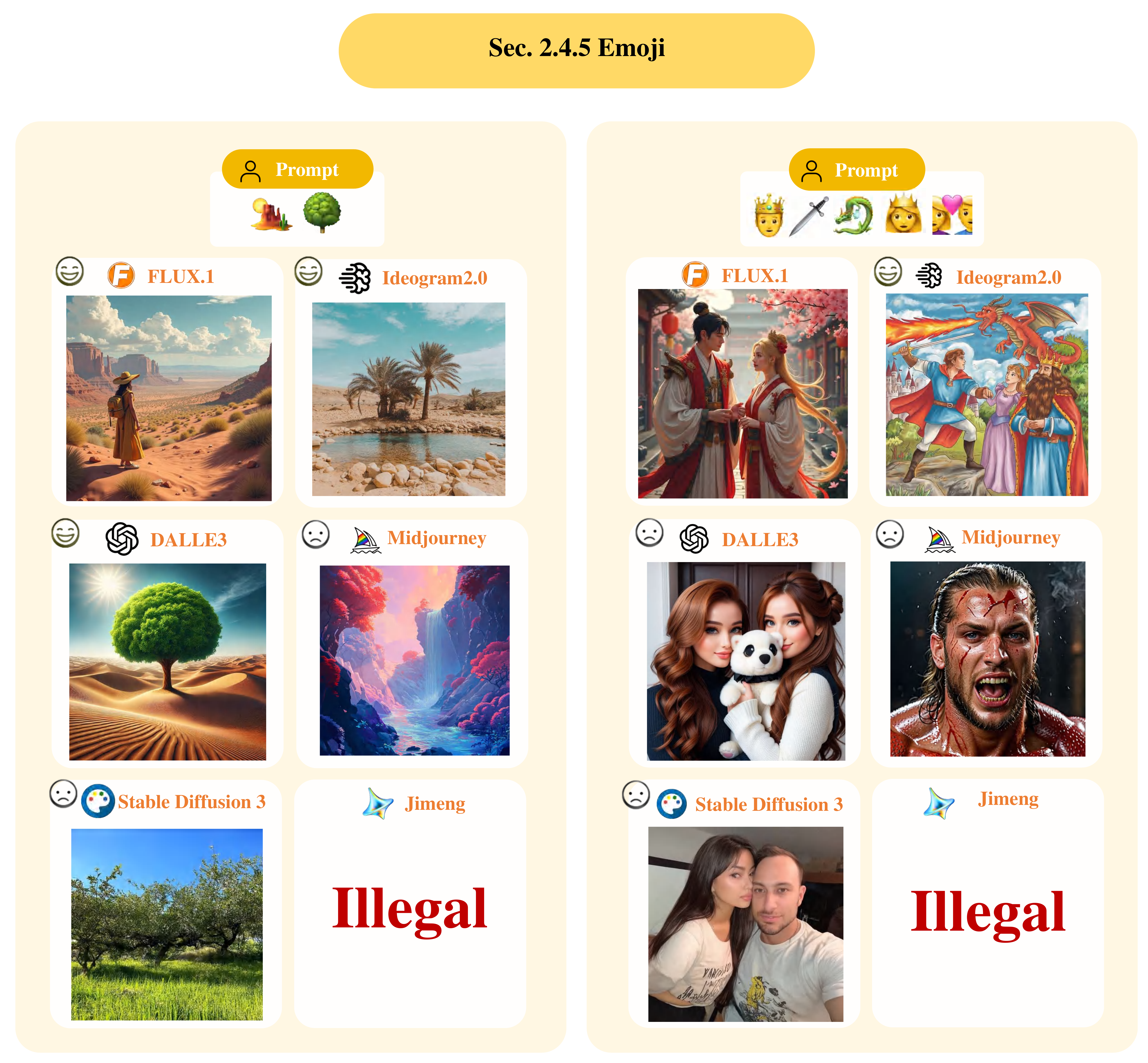}}
  \caption[Section~\ref{emoji2image}: emoji.]{Results on emoji task. Refer to Section \ref{emoji2image} for detailed discussions.}
  \label{fig_emojis2}
\end{figure*}

In Figures \ref{fig_emojis1}-\ref{fig_emojis2}, we investigate the models' ability to comprehend emojis. We observe that FLUX.1 and Ideogram2.0 attempt to construct stories from combinations of emojis, but they tend to focus on certain emojis while ignoring others. For example, FLUX.1 disregards the construction site emoji in the second prompt of Figure \ref{fig_emojis1} and the tree emoji in the first prompt of Figure \ref{fig_emojis2}. Ideogram2.0 performs better than FLUX.1 by considering nearly all emojis and their logical relationships. For instance, as shown in the left subplot of Figure \ref{fig_emojis2}, Ideogram2.0 integrates the desert and tree emojis into an oasis in the first example and also accurately understands the story conveyed by the emojis in the second example. Dall-E3 approaches the inputs differently, merging all emojis into its output and rendering it in a comic style. However, it sometimes loses logical coherence, as seen in the first example of the left subplot in Figure \ref{fig_emojis2}, where it depicts a tree standing alone in the desert. Midjourney, Jimeng, and Stable Diffusion 3 struggle to handle this type of task correctly, with Jimeng even considering this type of prompt illegal. Notably, Ideogram2.0 outperforms the others across all models, particularly in managing complex, multi-emoji inputs and accurately interpreting the stories expressed by the emojis.

\textbf{Score. }
The results of this experiment are shown in the Table \ref{tab_emoji}. In this task, the outputs of FLUX.1, Ideogram2.0, and Dall-E3 are all relatively good.
\begin{table}[h]
    \centering
    \caption[Section~\ref{emoji2image}: emoji.]{The scoring of generation results by six models on emoji task under different evaluation systems. Refer to Section \ref{emoji2image} for detailed discussions.}
    \begin{tabular}{l|c|c|c|c|c}
        \midrule
        Model & CLIPScore & HPSv2 & Aesthetic Score &GPT-4o & Human \\
        \midrule 
        FLUX.1      & 19.51 & \textbf{0.23} & \textbf{6.98} &  5.83 & 8.89 \\
        Ideogram2.0 & 19.45 & \textbf{0.23} & 6.18 &   6.39 & \textbf{9.72} \\
        Dall-E3     & \textbf{21.19} & 0.21 & 6.15 &   \textbf{6.94} & 8.05 \\
        Midjourney  & 15.47 & 0.20 & 5.52 &  3.89 & 5.00 \\
        SD3         & 23.82 & \textbf{0.23} & 5.89 &   4.17 & 5.83 \\
        \midrule
    \end{tabular}
    \label{tab_emoji}
\end{table}

\subsubsection{Irrational Scene Generation}
\label{irrational}
\begin{figure*}[!ht]
  \centering 
  \makebox[\textwidth][c]{\includegraphics[width=1\textwidth]{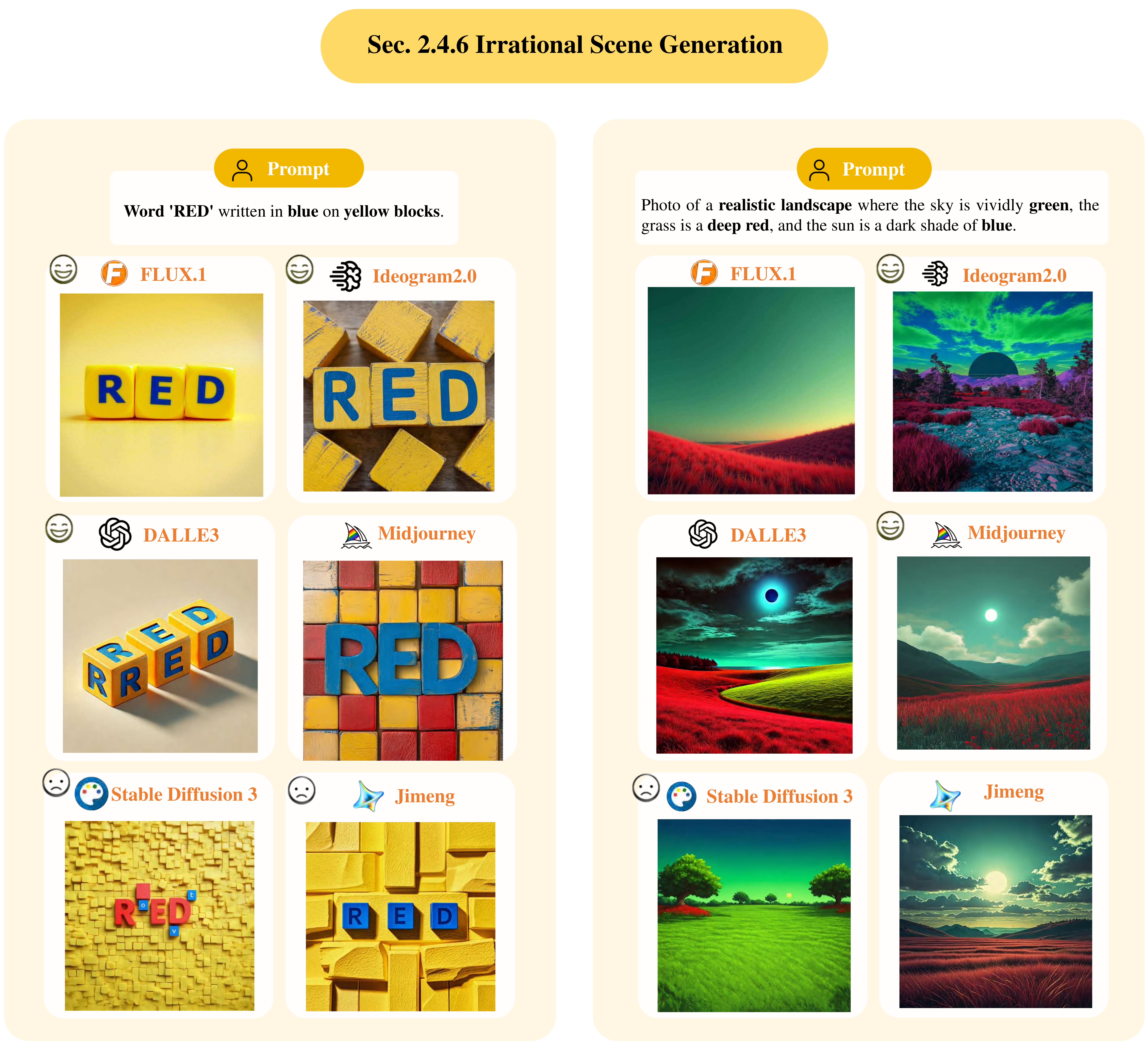}}
  \caption[Section~\ref{irrational}: Irrational Scene Generation]{Results on irrational scene generation task. Refer to Section \ref{irrational} for detailed discussions.}
  \label{fig_irrational1}
\end{figure*}

\begin{figure*}[!ht]
  \centering 
  \makebox[\textwidth][c]{\includegraphics[width=0.7\textwidth]{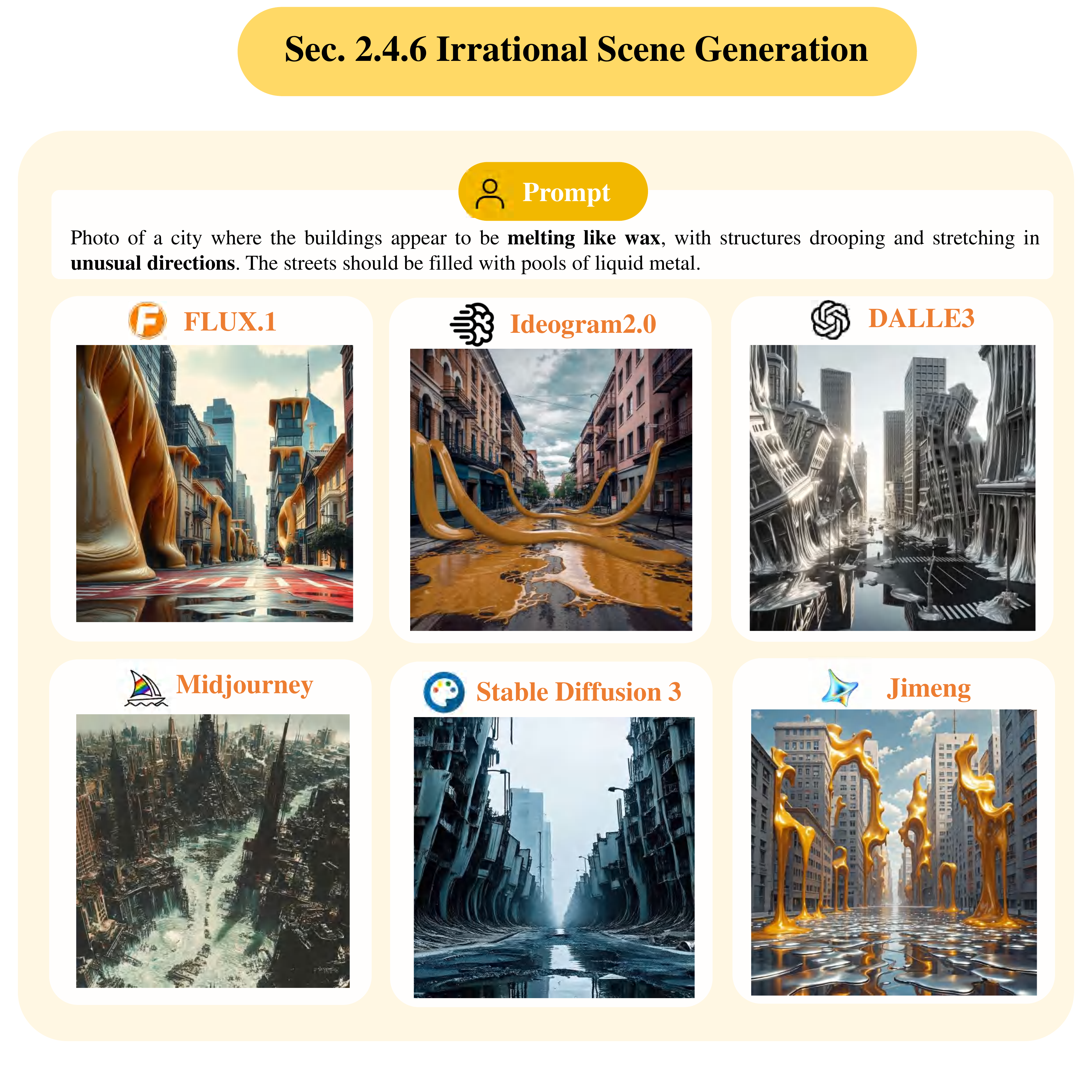}}
  \caption[Section~\ref{irrational}: Irrational Scene Generation]{Results on irrational scene generation task. Refer to Section \ref{irrational} for detailed discussions.}
  \label{fig_irrational2}
\end{figure*}

In Figure \ref{fig_irrational1}-\ref{fig_irrational2}, we evaluated the ability of these models to generate irrational scenes. In the first instance, we instructed the models to generate the word "RED" written in blue on yellow blocks. FLUX.1, Ideogram2.0, and Dall-E3 completed the task perfectly. Midjourney generated a red background, Jimeng used black text, and Stable Diffusion 3 used red text, each with some flaws. In the second and third instances, we had the models generate anomalous scenes, including objects with unusual colors and materials. Among all the models, Dall-E3 performed the best, perfectly generating scenes according to the text description. When generating objects with anomalous colors, FLUX.1, Stable Diffusion 3, and Jimeng missed parts of the text description, resulting in flawed outputs. Additionally, FLUX.1, Ideogram2.0, and Jimeng failed to correctly interpret the metaphor in the text when generating objects with anomalous materials, leading to incorrect outputs.

\textbf{Score. }
The results of this experiment are shown in the Table \ref{tab_irrational}. In this task, the outputs of FLUX.1, Ideogram2.0, and Dall-E3 are relatively good; however, the results for CLIPScore and GPT-4o differ significantly from human evaluations.
\begin{table}[h]
    \centering
    \caption[Section~\ref{irrational}: irrational scene generation.]{The scoring of generation results by six models on irrational scene generation under different evaluation systems. Refer to Section \ref{irrational} for detailed discussions.}
    \begin{tabular}{l|c|c|c|c|c}
        \midrule
        Model & CLIPScore & HPSv2 & Aesthetic Score &  GPT-4o & Human \\
        \midrule 
        FLUX.1      & 30.06 & \textbf{0.28} & 6.02 &  6.39 & 8.33 \\
        Ideogram2.0 & 28.47 & \textbf{0.28} & 5.93 &  6.11 & 8.06 \\
        Dall-E3     & 31.89 & 0.26 & \textbf{6.17} &  5.83 & \textbf{8.61} \\
        Midjourney  & 30.57 & 0.27 & 5.90 &  4.45 & 7.22 \\
        SD3         & 30.66 & 0.27 & 6.08 &  \textbf{6.94} & 6.94 \\
        Jimeng      & \textbf{32.47} & 0.27 & 5.88 &  \textbf{6.94} & 6.94 \\
        \midrule
    \end{tabular}
    \label{tab_irrational}
\end{table}

\subsubsection{LLM QA}
\label{qa}
\begin{figure*}[!ht]
  \centering 
  \makebox[\textwidth][c]{\includegraphics[width=1\textwidth]{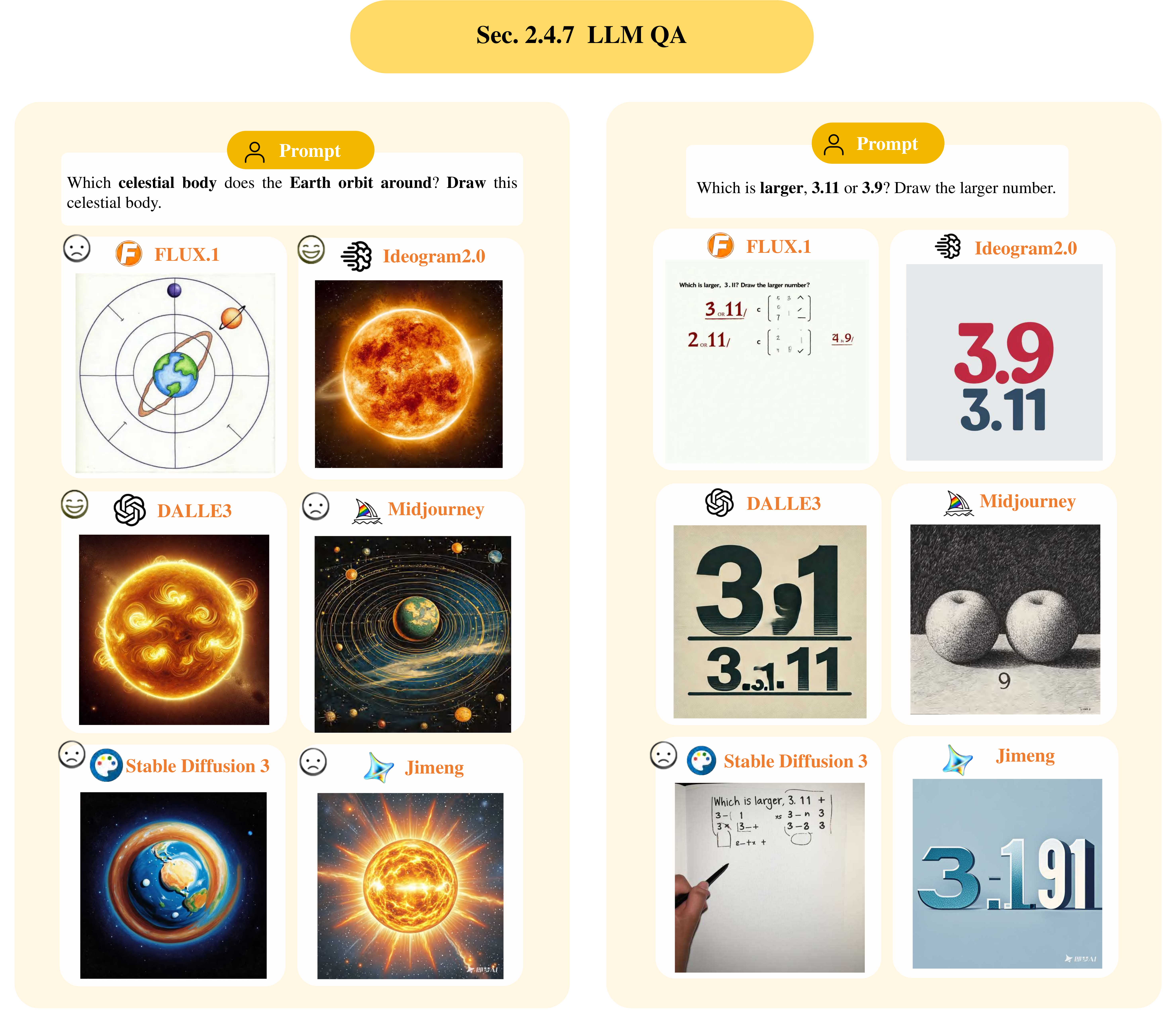}}
  \caption[Section~\ref{qa}: LLM QA.]{Results on LLM QA task. Refer to Section \ref{qa} for detailed discussions.}
  \label{qa_image}
\end{figure*}

As the T2I models get stronger and more versatile, it seems that they have the ability to develop towards fundamental models. In Figure \ref{qa_image}, we illustrate two examples to assess the LLM question-answering (QA) capabilities of models. This evaluation involves both understanding the questions and generating accurate visual representations of the answers. In the first example, the correct answer to the question about the celestial body orbiting the Earth is the Sun. Models such as Ideogram2.0, Dall-E3, and Jimeng correctly interpret the question and generate images representing the Sun. Although these images may not be scientifically precise representations of the Sun, they are easily recognizable. Conversely, models such as FLUX.1, Midjourney, and Stable Diffusion 3 misunderstand the prompt and produce irrelevant images. In the second example, which involves a simple mathematical query, the desired output is an image featuring the number 3.11. Among the models, only Dall-E3 generates an image that approximates the correct answer. The other models produce images that do not correspond to the required representation.

\textbf{Score. }
The results of this experiment are shown in the Table \ref{tab_qa}. Due to the evaluation system of this experiment requiring both an understanding of the images and a certain level of mathematical and physical knowledge, the results obtained may differ significantly from human intuitive perception.

\begin{table}[h]
    \centering
    \caption[Section~\ref{qa}: LLM QA.]{The scoring of generation results by six models on LLM QA under different evaluation systems. Refer to Section \ref{qa} for detailed discussions.}
    \begin{tabular}{l|c|c|c|c|c}
        \midrule
        Model & CLIPScore & HPSv2 & Aesthetic Score & GPT-4o & Human \\
        \midrule 
        FLUX.1      & 24.49 & 0.23 & 5.13 &  5.00 & \textbf{10.00} \\
        Ideogram2.0 & 20.72 & 0.23 & 5.32 &  4.59 & 7.92 \\
        Dall-E3     & \textbf{26.51} & 0.24 & \textbf{5.71} &  5.42 & 7.92 \\
        Midjourney  & 23.53 & 0.23 & 5.66 &  \textbf{6.67} & 6.83 \\
        SD3         & 21.75 & 0.21 & 5.09 &  5.71 & 2.92 \\
        Jimeng      & 23.67 & \textbf{0.25} & 5.66 &  5.42 & 6.67 \\
        \midrule
    \end{tabular}
    \label{tab_qa}
\end{table}

\subsubsection{Watermark}
\label{watermark}
\begin{figure*}[!ht]
  \centering 
  \makebox[\textwidth][c]{\includegraphics[width=1\textwidth]{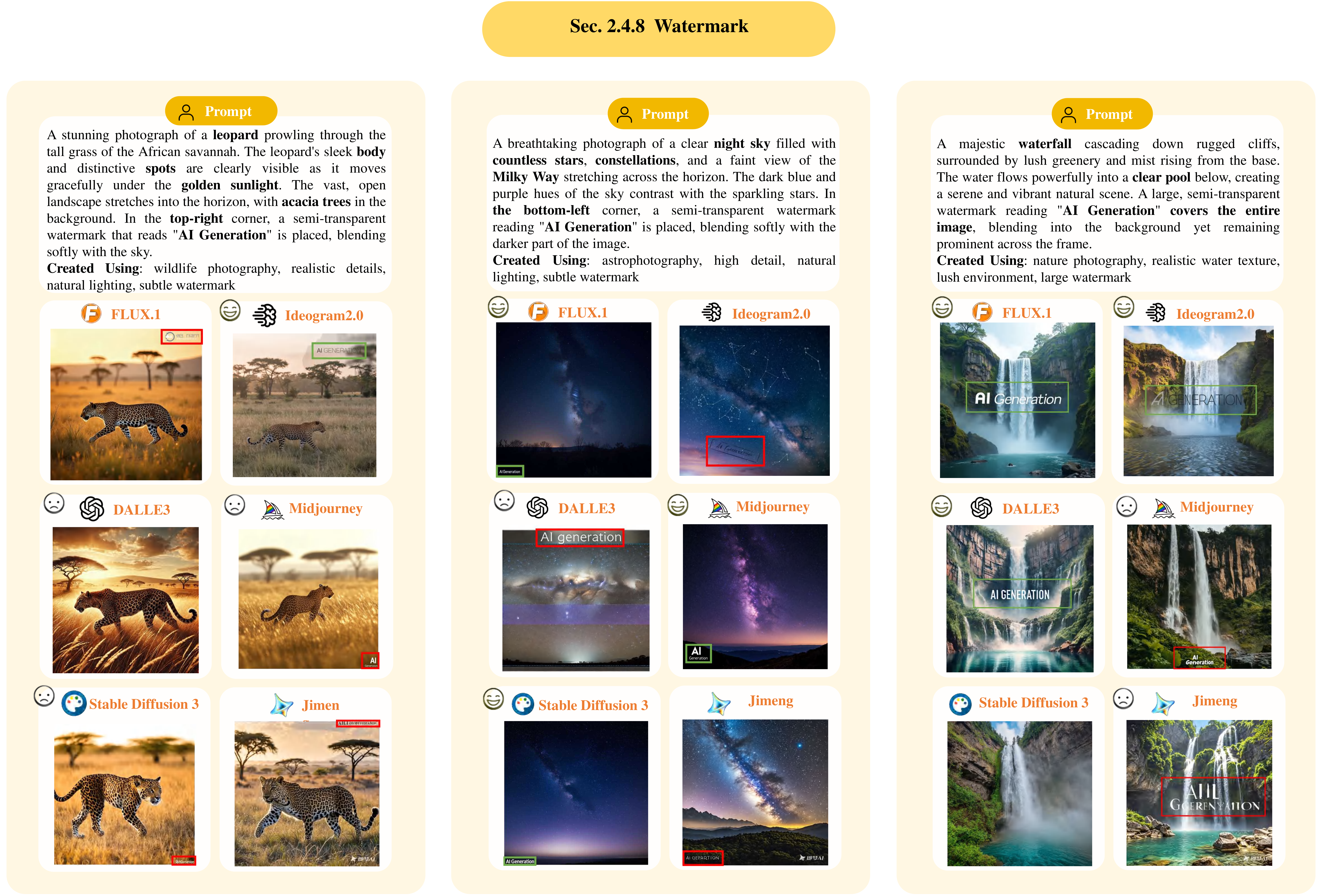}}
  \caption[Section~\ref{watermark}: watermark.]{Results on watermark task, with green frames to highlight the right watermarks and red frames to denote watermarks that are either misplaced, contain incorrect text, or are entirely omitted. Refer to Section \ref{watermark} for detailed discussions.}
  \label{watermark2}
\end{figure*}

In Figures \ref{watermark2}, we evaluate the models' capability to correctly generate watermarks in images, which remind that the images are AI-generated for safety purposes. We prompt the models to create images with watermarks placed in different locations and avoid destroying the integrity of the pictures. We use green frames to highlight the right watermarks and red frames to denote watermarks that are either misplaced, contain incorrect text, or are entirely omitted. The results show that none of the models consistently produced correct watermarks across all examples. Notably, the Jimeng model failed to generate accurate watermarks in any task, potentially due to its inherent watermarking system.

\textbf{Score. }
The results of this experiment are shown in the Table \ref{tab_watermark}. The evaluation in this experiment focuses on the accuracy of watermark placement, which requires precise attention to image details. As a result, there is some difference between other evaluation systems and human scoring.
\begin{table}[h]
    \centering
    \caption[Section~\ref{watermark}: watermark.]{The scoring of generation results by six models on watermark task under different evaluation systems. Refer to Section \ref{watermark} for detailed discussions.}
    \begin{tabular}{l|c|c|c|c|c}
        \midrule
        Model & CLIPScore & HPSv2 & Aesthetic Score & GPT-4o & Human \\
        \midrule 
        FLUX.1      & 27.37 & 0.25 & 5.17 &   6.67 & \textbf{9.17} \\
        Ideogram2.0 & 27.10 & \textbf{0.28} & \textbf{6.10} &   6.67 & 8.33 \\
        Dall-E3     & 27.42 & 0.26 & 5.83 &   5.83 & 5.42 \\
        Midjourney  & 29.01 & 0.24 & 5.66 &   \textbf{8.34} & 8.34 \\
        SD3         & \textbf{30.26} & 0.25 & 5.73 &   7.50 & 8.33 \\
        Jimeng      & 29.34 & \textbf{0.30} & \textbf{6.10} &   5.00 & 8.33 \\
        \midrule
    \end{tabular}
    \label{tab_watermark}
\end{table}

\subsubsection{Low Quality}
\label{ood}
\begin{figure*}[!ht]
\vspace{-1em}
  \centering 
  \makebox[\textwidth][c]{\includegraphics[width=0.9\textwidth]{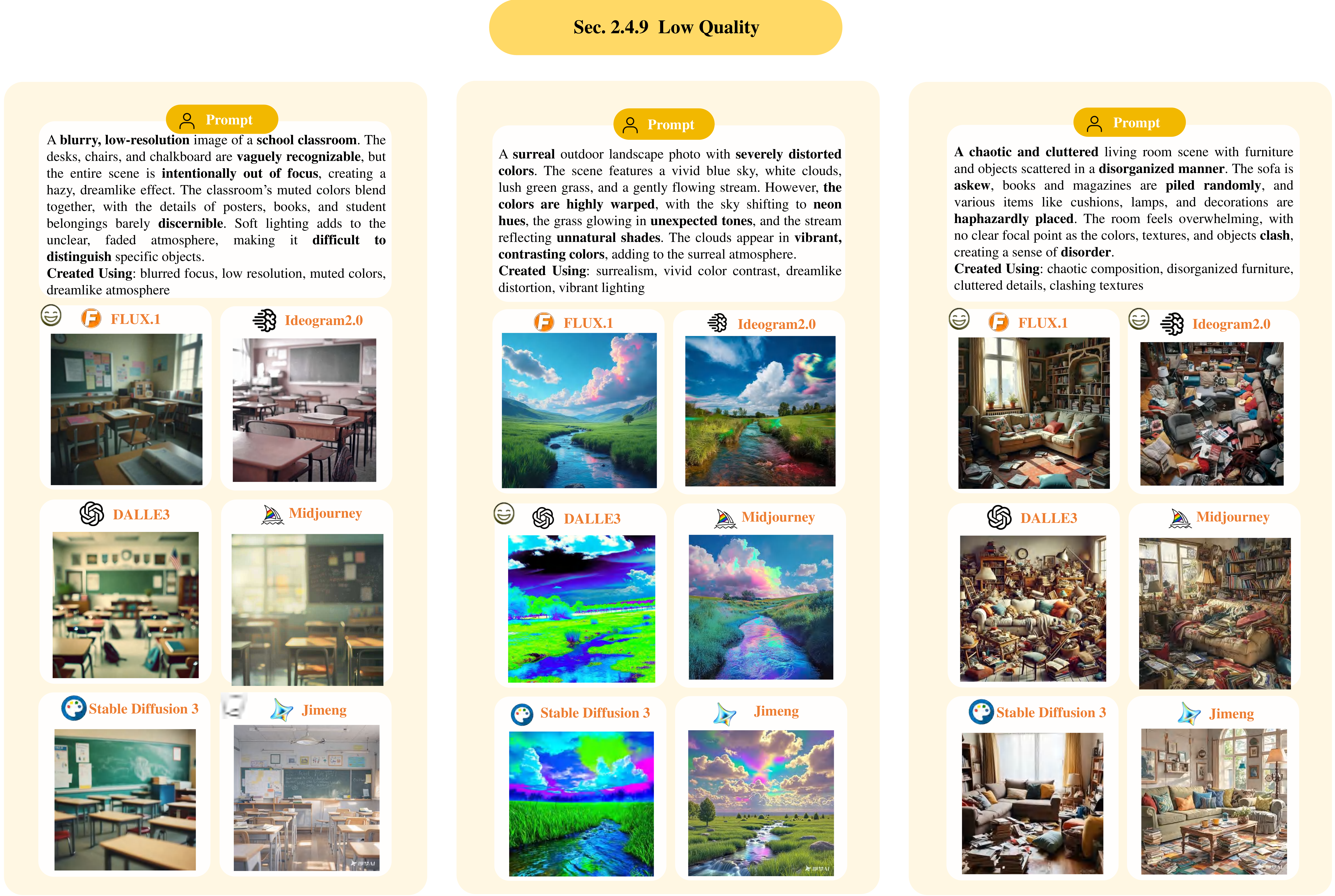}}
  \caption[Section~\ref{ood}: low quality.]{Results on low quality task. Refer to Section \ref{ood} for detailed discussions.}
  \label{OOD1}
\end{figure*}

\begin{figure*}[!ht]
\vspace{1em}
  \centering 
  \makebox[\textwidth][c]{\includegraphics[width=1\textwidth]{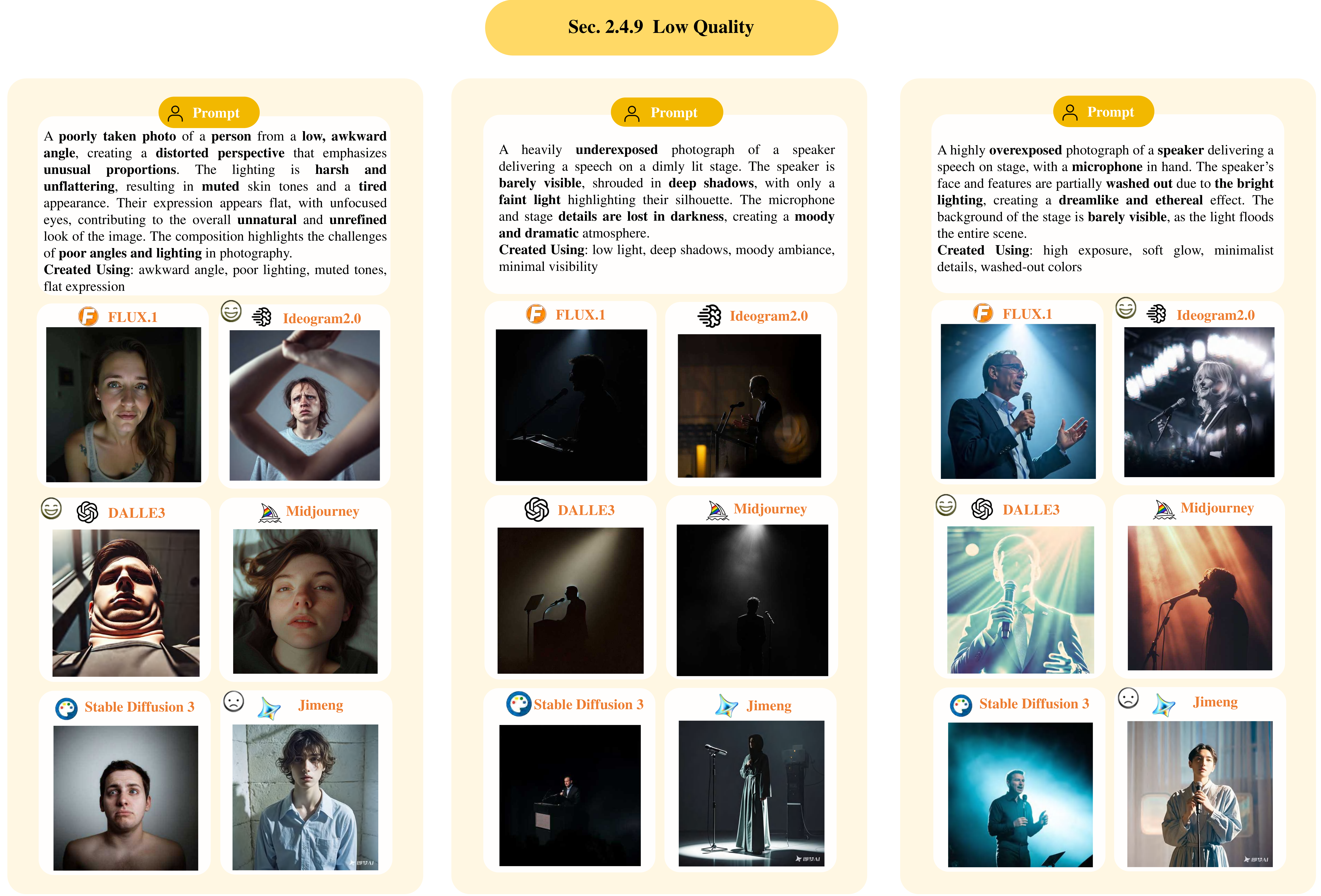}}
  \caption[Section~\ref{ood}: low quality.]{Results on low quality task. Refer to Section \ref{ood} for detailed discussions.}
  \label{OOD2}
\end{figure*}

\begin{figure*}[!ht]
  \centering 
  \makebox[\textwidth][c]{\includegraphics[width=1\textwidth]{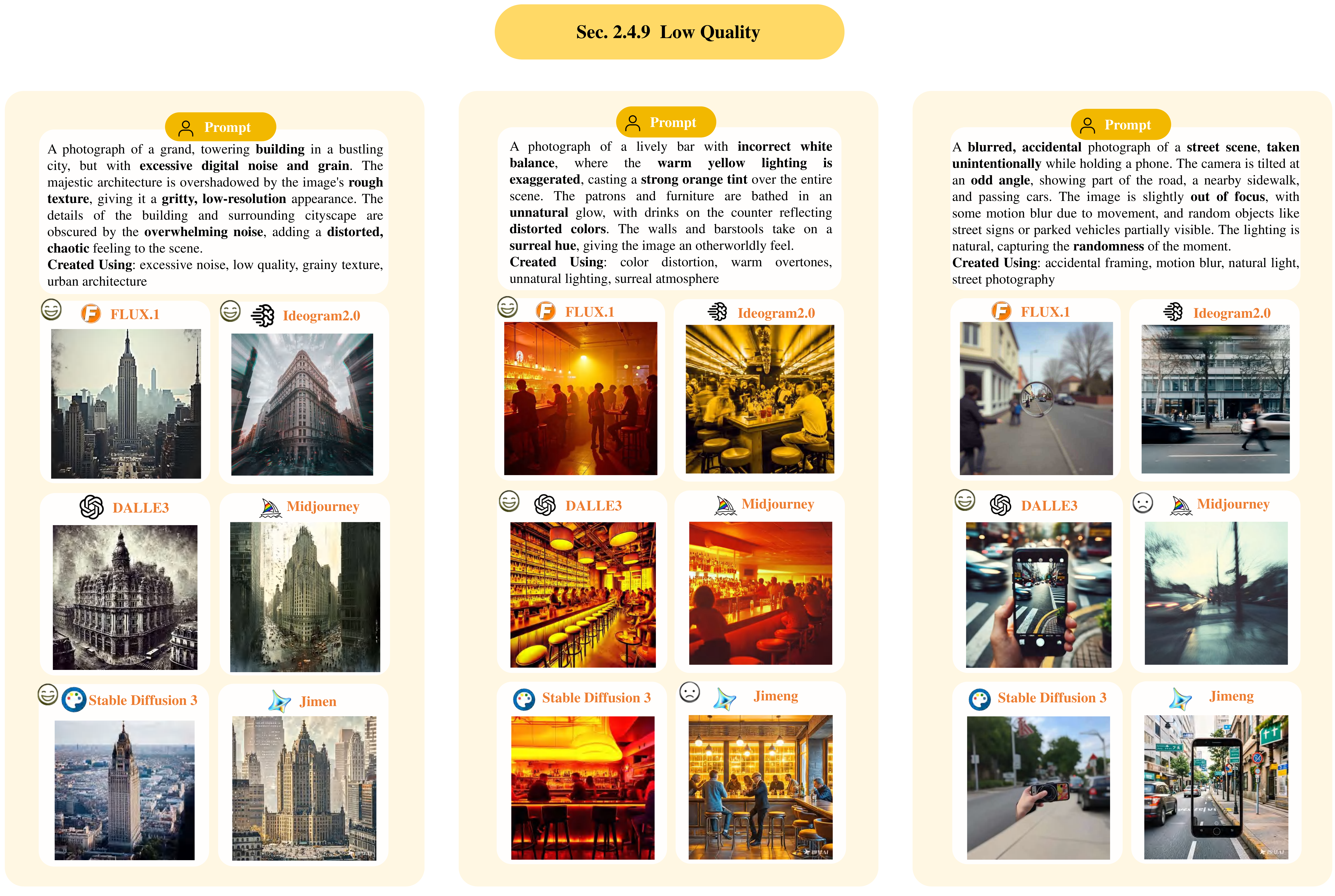}}
  \caption[Section~\ref{ood}: low quality.]{Results on low quality task. Refer to Section \ref{ood} for detailed discussions.}
  \label{OOD3}
\end{figure*}

In Figures \ref{OOD1}-\ref{OOD3}, we investigate the models' ability to generate low-quality images. During training, some models may discard low-quality images, while others may retain them and apply corresponding labels. This task tends to infer the nature of the datasets used by these models. Specifically, if a model can accurately generate low-quality images, it suggests that labeled low-quality data was incorporated during training. We provide prompts including low resolution, distorted colors, disorganization, ugly figures, underexposure, overexposure, excessive noise, incorrect white balance, and accidental photography. Notably, Dall-E3 successfully generates accurate representations for all prompts, except for the example of accidental photograph, where Dall-E3 might misunderstand the prompt "while holding a phone." This suggests that Dall-E3 may have been trained on datasets including low-quality data with corresponding labels. In contrast, other models consistently produced high-quality images to satisfy the key information of prompts.

The following provides a more detailed analysis of each example.

\textbf{Low Resolution.} In the left subplot of Figures \ref{OOD1}, FLUX.1, Dall-E3, Midjourney, and Stable Diffusion 3 successfully generate low-resolution images as specified in the prompt. However, Ideogram2.0 produces an image resembling an out-of-focus photograph, while Jimeng generates a high-resolution image, deviating from the intended requirement.

\textbf{Distorted Colors.} In the middle subplot of Figures \ref{OOD1}, Dall-E3 and Stable Diffusion 3 produce images with notably distorted colors, while the outputs from the other models exhibit an appealing use of multiple colors which are inconsistent with the prompt of istorted colors.

\textbf{Disorganization.} Among the models, Dall-E3 and Midjourney produce images that best meet the prompt's criteria in the right subplot of Figures \ref{OOD1},. The images generated by FLUX.1, Stable Diffusion 3, and Jimeng, although present a level of disorganization but are more like well-laid.

\textbf{Ugly Figures.} In the left subplot of Figures \ref{OOD2}, Dall-E3 generates the image that most closely aligns with the prompt's specifications. While Ideogram2.0 also creates an unattractive figure, it includes an unreasonable foreground element. FLUX.1, Midjourney, and Stable Diffusion 3 generate detailed portraits, and Jimeng produces even more intricate and refined images.

\textbf{Underexposure.} In the middle subplot of Figures \ref{OOD2}, all models generate images that exhibit underexposure, matching the prompt's expectations.

\textbf{Overexposure.} Only Dall-E3 successfully produces an overexposed photograph, in the right subplot of Figures \ref{OOD2}. The other models fail to accurately catch the concept of overexposure.

\textbf{Excessive Noise.} In the left subplot of Figures \ref{OOD3}, the image generated by Dall-E3 contains excessive digital noise and grain. The other models also produce lower-quality images, but Ideogram2.0 stands out for incorporating elaborate special effects into its output.

\textbf{Incorrect White Balance.} In the middle subplot of Figures \ref{OOD3}, Jimeng is the only model that misinterprets the concept of \textit{incorrect white balance}, producing an image with a warm lighting scene rather than the expected effect.

\textbf{Accidental Photography.} In the right subplot of Figures \ref{OOD3}, Ideogram2.0 and Midjourney accurately adhere to the prompt, generating images that depict the concept of accidental photography. Dall-E3 and Jimeng, however, place too much emphasis on the detail of "while holding a phone," resulting in images that are inconsistent with the intended outcome.

\textbf{Score. }
The results of this experiment are shown in the Table \ref{tab_ood}. Midjourney achieved the best generation performance in this experiment. Since several models produced similar generation results, the differences in evaluation scores are not significant.
\begin{table}[h]
    \centering
    \caption[Section~\ref{ood}: low quality.]{The scoring of generation results by six models on low quality under different evaluation systems. Refer to Section \ref{ood} for detailed discussions.}
    \begin{tabular}{l|c|c|c|c|c}
        \midrule
        Model & CLIPScore & HPSv2 & Aesthetic Score & GPT-4o & Human \\
        \midrule 
        FLUX.1      & 26.99 & 0.25 & 5.69 &  \textbf{6.30} & 9.07 \\
        Ideogram2.0 & 24.15 & 0.25 & 5.65 &  6.11 & 9.35 \\
        Dall-E3     & 26.92 & 0.26 & 5.68 &  6.02 & \textbf{9.63} \\
        Midjourney  & 24.08 & \textbf{0.28} & 5.83 &  5.46 & 8.98 \\
        SD3         & 25.38 & 0.27 & 5.65 &  5.83 & 8.98 \\
        Jimeng      & \textbf{27.71} & 0.26 & \textbf{5.90} &  5.37 & 8.24 \\
        \midrule
    \end{tabular}
    \label{tab_ood}
\end{table}

\subsubsection{Multi-image}
\label{multi-image}
Recent studies~\cite{Huang2024InContextLF, mao2025ace++, tan2024ominicontrol, shin2024large, cai2024diffusion} have demonstrated that advanced image generation models can produce grid-based compositions containing multiple images within a single output. 
These grid-based images not only maintain a certain level of contextual consistency across different cells but also extend the applicability of these models to various tasks, such as subject/style consistency, storyboard generation, and logical reasoning.
This observation suggests two key insights: first, these models can leverage a substantial amount of contextually coherent training data from existing internet sources during training; second, they exhibit a certain degree of holistic generation capability, conditioned on prompts and contextual information.
Motivated by these findings, we further assess the ability of state-of-the-art image generation models in producing contextually consistent grid images, paving the way for the evolution of image generation models toward more general-purpose intelligent agents~\cite{radford2018improving, radford2019language, brown2020language}.

\begin{figure*}[!ht]
\vspace{-1em}
  \centering 
  \makebox[\textwidth][c]{\includegraphics[width=0.95\textwidth]{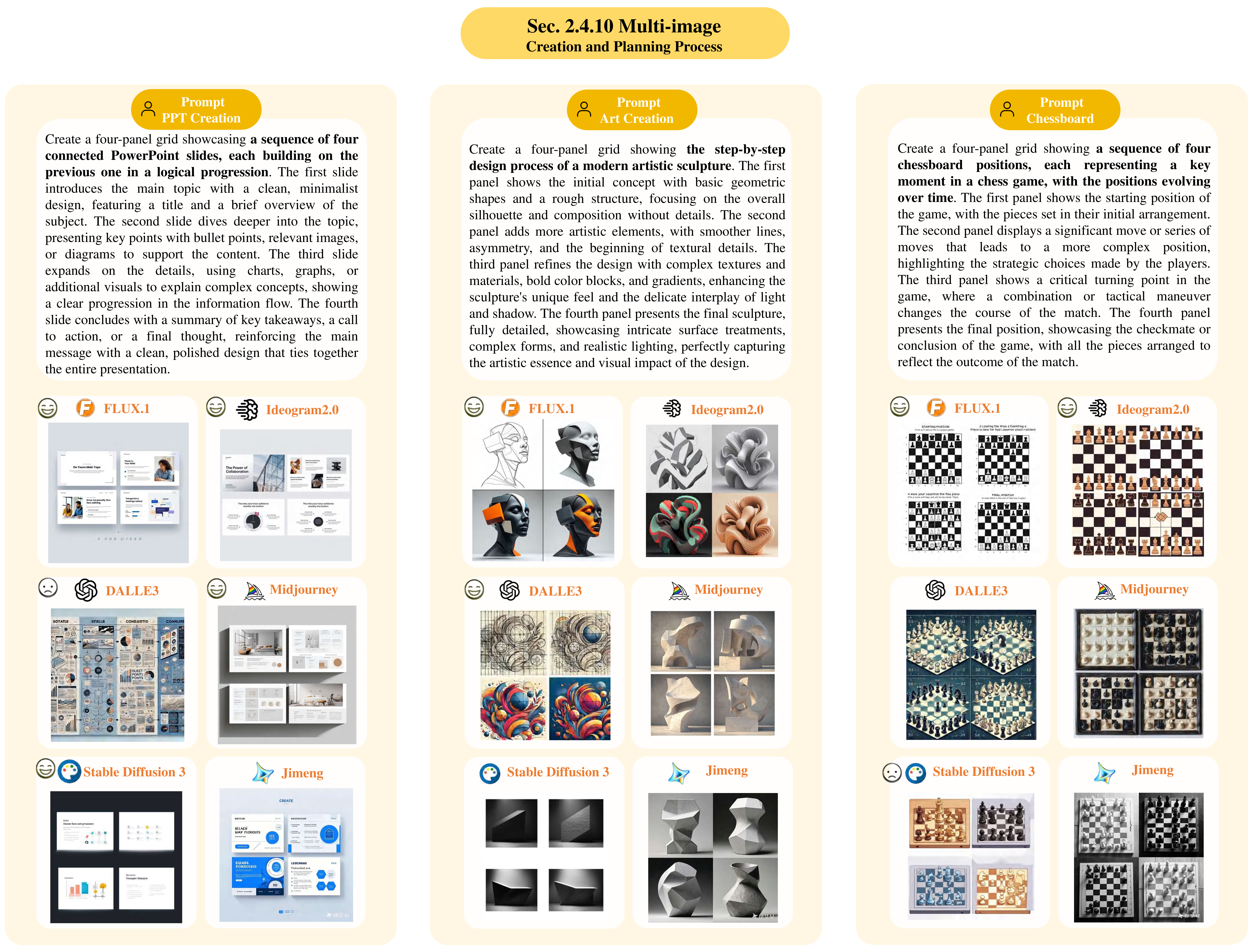}}
  \caption[Section~\ref{multi-image}: multi-image.]{Results on creation and planning process task. Refer to Section \ref{multi-image} for detailed discussions.}
  \label{fig_create}
\end{figure*}

\textbf{Creation and Planning process. }
In Figure \ref{fig_create}, we explore the models' ability to generate creation and planning processes. It is evident that FLUX.1 and Ideogram2.0 can generate reasonable PPT creation processes and deduce plausible chess game strategies. Dall-E3, on the other hand, can accurately generate the process of creating artwork sketches based on our requirements. These three models effectively present the creative planning process in a grid format.

\begin{figure*}[!ht]
\vspace{-1em}
  \centering 
  \makebox[\textwidth][c]{\includegraphics[width=0.8\textwidth]{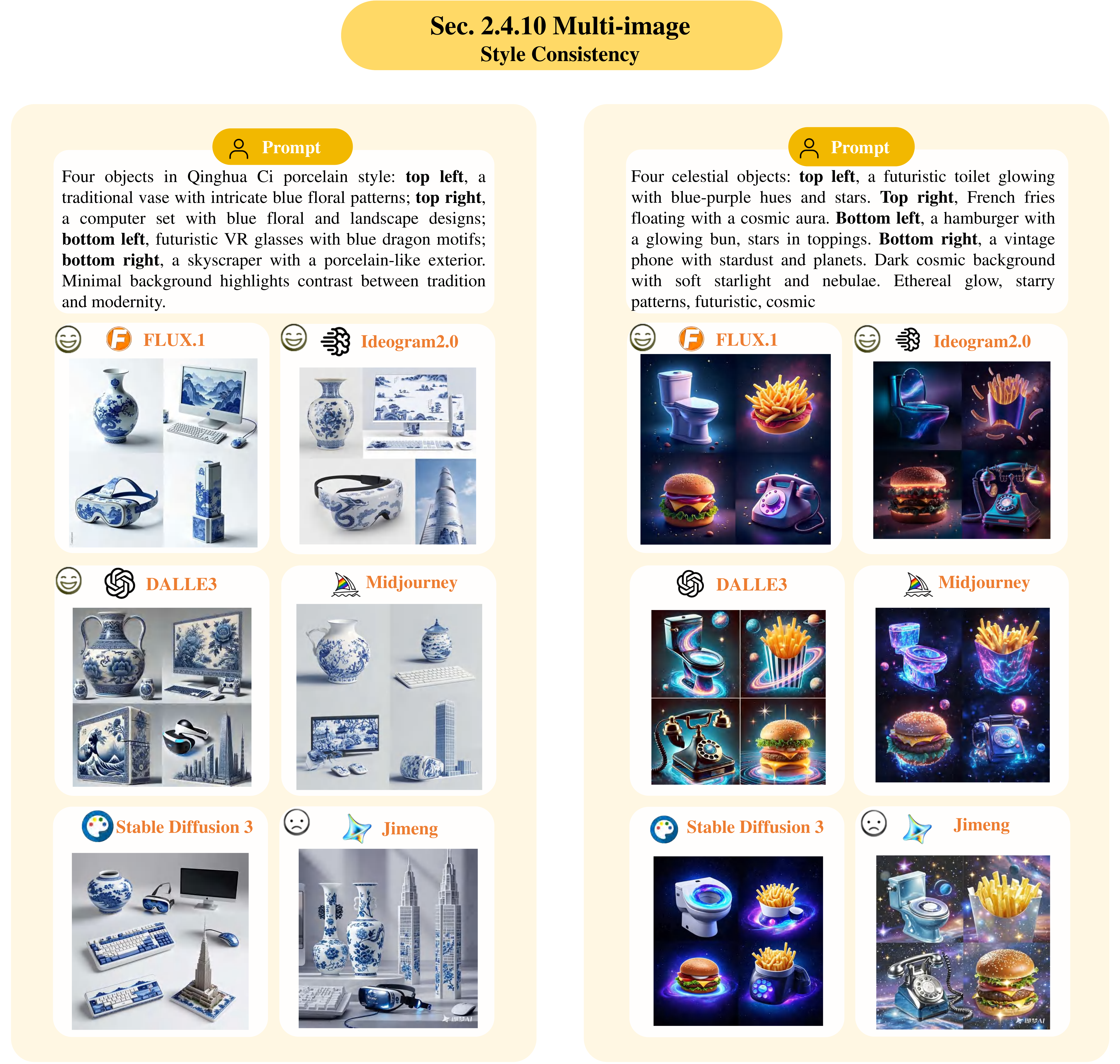}}
  \caption[Section~\ref{multi-image}: multi-image.]{Results on style consistency task. Refer to Section \ref{multi-image} for detailed discussions.}
  \label{fig_style_consistency}
\end{figure*}

\begin{figure*}[!ht]
  \centering 
  \makebox[\textwidth][c]{\includegraphics[width=0.7\textwidth]{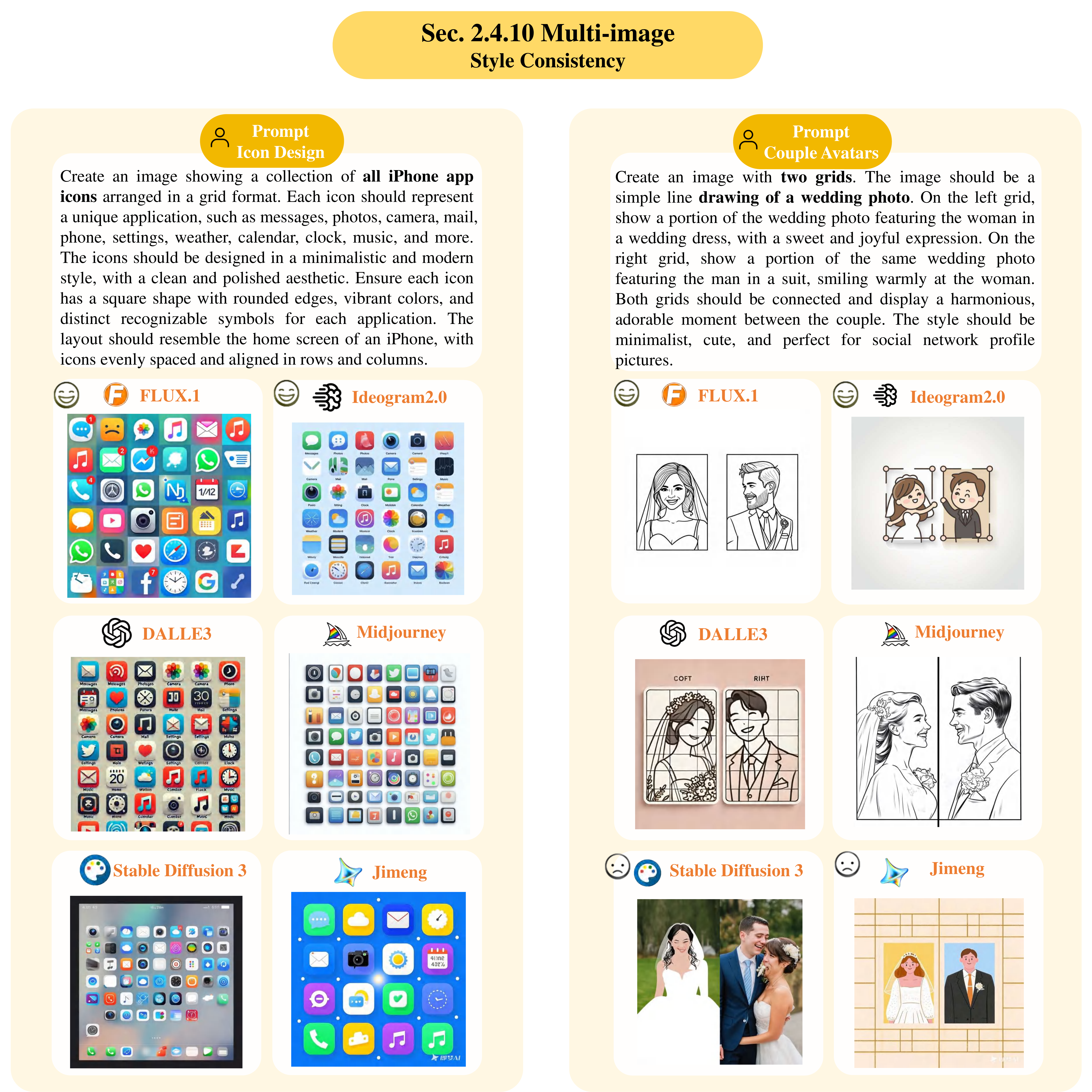}}
  \caption[Section~\ref{multi-image}: multi-image.]{Results on style consistency task. Refer to Section \ref{multi-image} for detailed discussions.}
  \label{fig_style_consistency_2}
\end{figure*}

\textbf{Style Consistency. }
In Figure \ref{fig_style_consistency}, we tested the capability of model style consistency. The model was tasked with generating four sub-images within a single image, each containing different objects, and converting them to the same style. FLUX.1 and Ideogram2.0 were able to follow the prompt relatively well. Dall-E3 and Midjourney could distinguish the sub-images and generate parts of the objects but with flaws. The capabilities of Stable Diffusion 3 and Jimeng were weaker, struggling to produce multi-image outputs and correctly render the objects.
Additionally, in Figure \ref{fig_style_consistency_2}, we designed tasks to generate app icons in a unified style and to create couple avatars. These tasks tested the models' ability to generate images with consistent styles. The results show that while the models can create images with a consistent style, some of the icons in the first task lack realism, with Ideogram2.0 performing the best. In the second task, all models successfully generated two grids, with FLUX.1, Ideogram2.0, and Midjourney producing the highest-quality images.

\begin{figure*}[!ht]
  \centering 
  \makebox[\textwidth][c]{\includegraphics[width=1\textwidth]{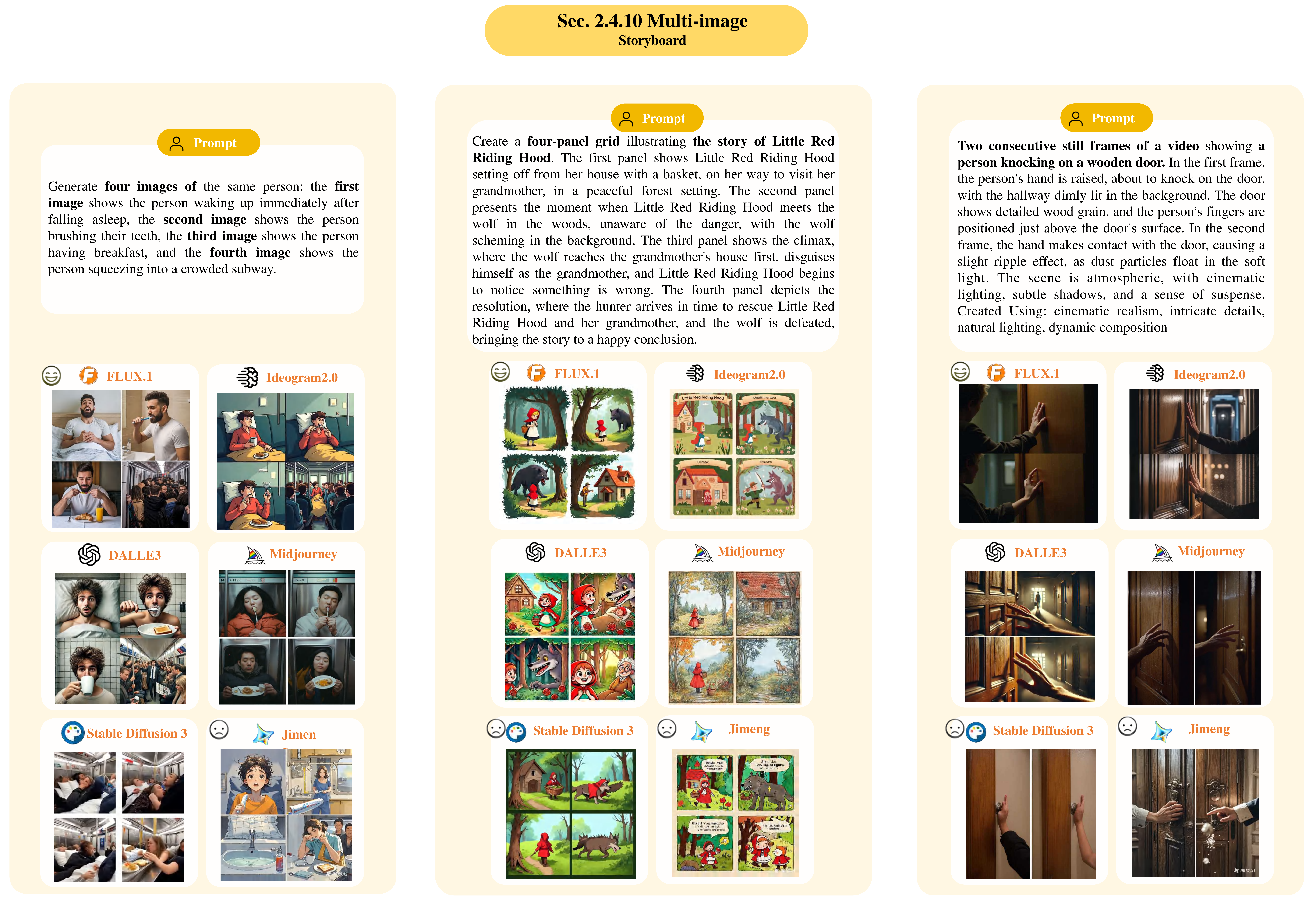}}
  \caption[Section~\ref{multi-image}: multi-image.]{Results on storyboard task. Refer to Section \ref{multi-image} for detailed discussions.}
  \label{fig_story}
\end{figure*}

\textbf{Storyboard. }
In Figure \ref{fig_story}, we tested the models for storytelling. The models were tasked with generating multiple images that exhibit logical relationships, with consistent styles, matching backgrounds and subjects, and a clear sense of temporal sequence. The models performed better when handling classic storylines, such as Little Red Riding Hood, compared to fabricated ones. Overall, FLUX.1 and Ideogram2.0 performed the best across all aspects, though there were still some logical inconsistencies, such as a person brushing their teeth and having breakfast in bed. In the prompt designed for a movie storyboard, FLUX.1 delivered the best results, while the other models exhibited issues such as background inconsistency.

\begin{figure*}[!ht]
  \centering 
  \makebox[\textwidth][c]{\includegraphics[width=1\textwidth]{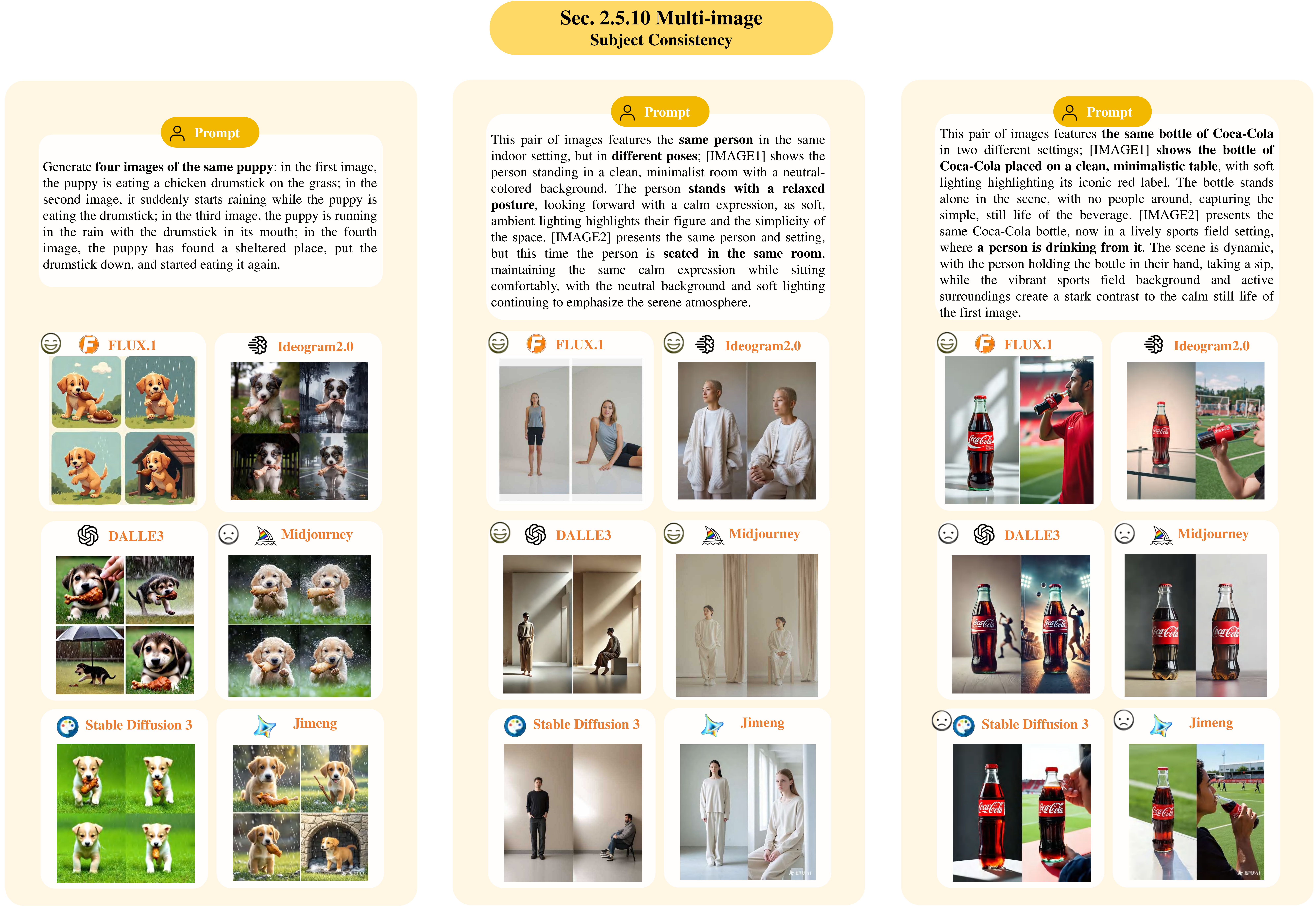}}
  \caption[Section~\ref{multi-image}: multi-image.]{Results on subject consistency task. Refer to Section \ref{multi-image} for detailed discussions.}
  \label{fig_subject}
\end{figure*}

\begin{figure*}[!ht]
  \centering 
  \makebox[\textwidth][c]{\includegraphics[width=0.75\textwidth]{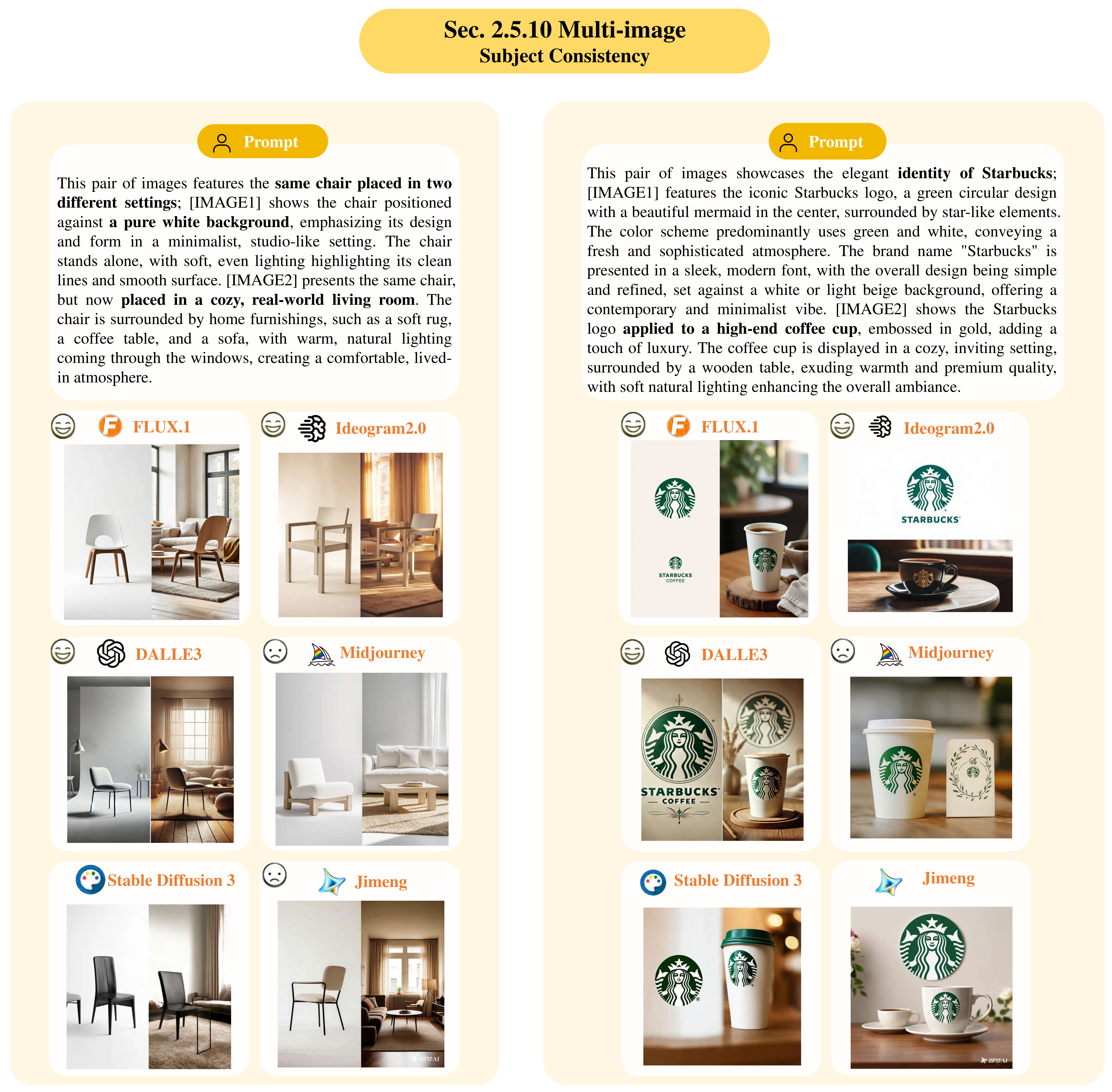}}
  \caption[Section~\ref{multi-image}: multi-image.]{Results on subject consistency task. Refer to Section \ref{multi-image} for detailed discussions.}
  \label{fig_subject_2}
\end{figure*}

\begin{figure*}[!ht]
  \centering 
  \makebox[\textwidth][c]{\includegraphics[width=0.7\textwidth]{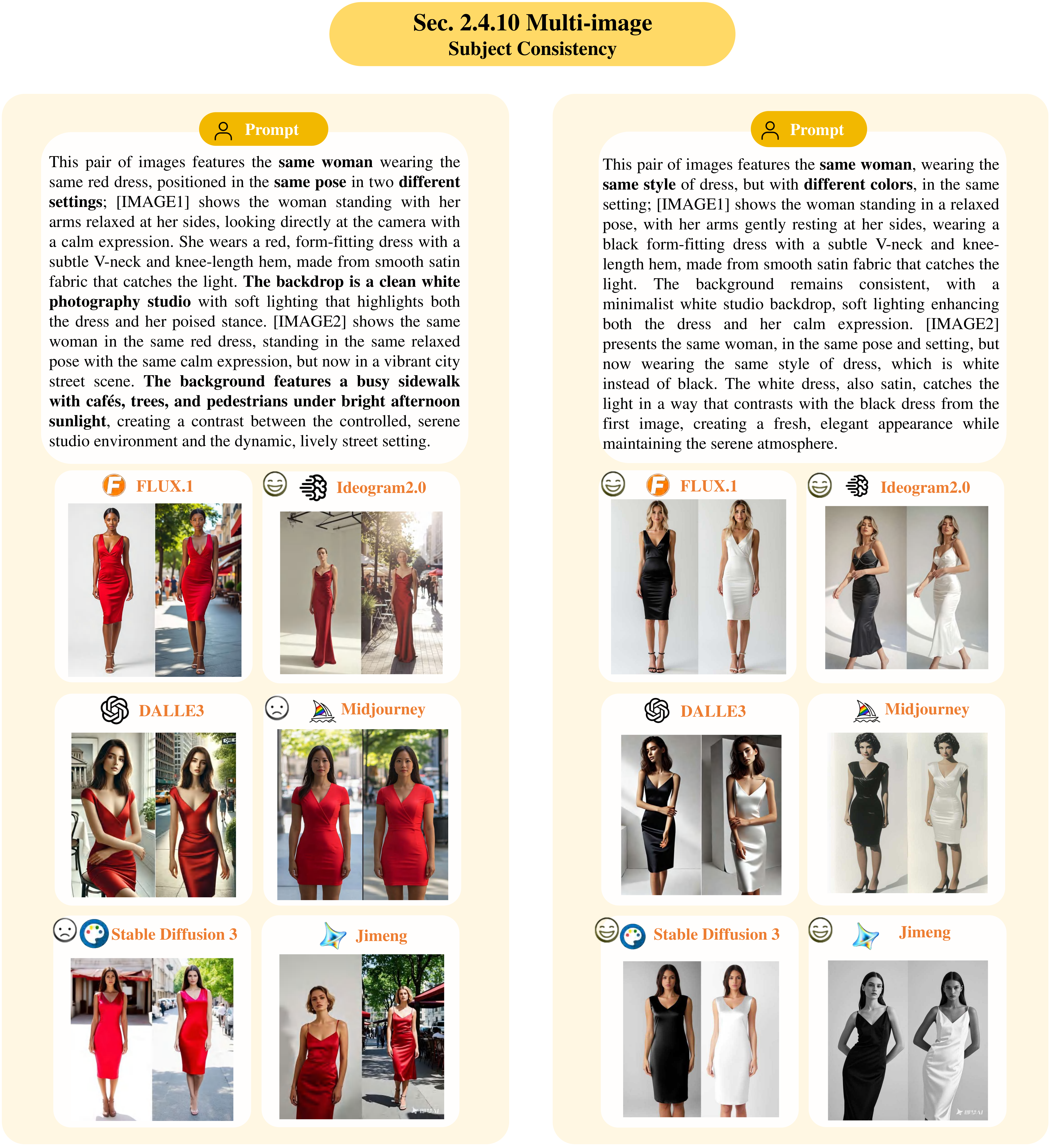}}
  \caption[Section~\ref{multi-image}: multi-image.]{Results on subject consistency task. Refer to Section \ref{multi-image} for detailed discussions.}
  \label{fig_subject_3}
\end{figure*}

\begin{figure*}[!ht]
  \centering 
  \makebox[\textwidth][c]{\includegraphics[width=0.7\textwidth]{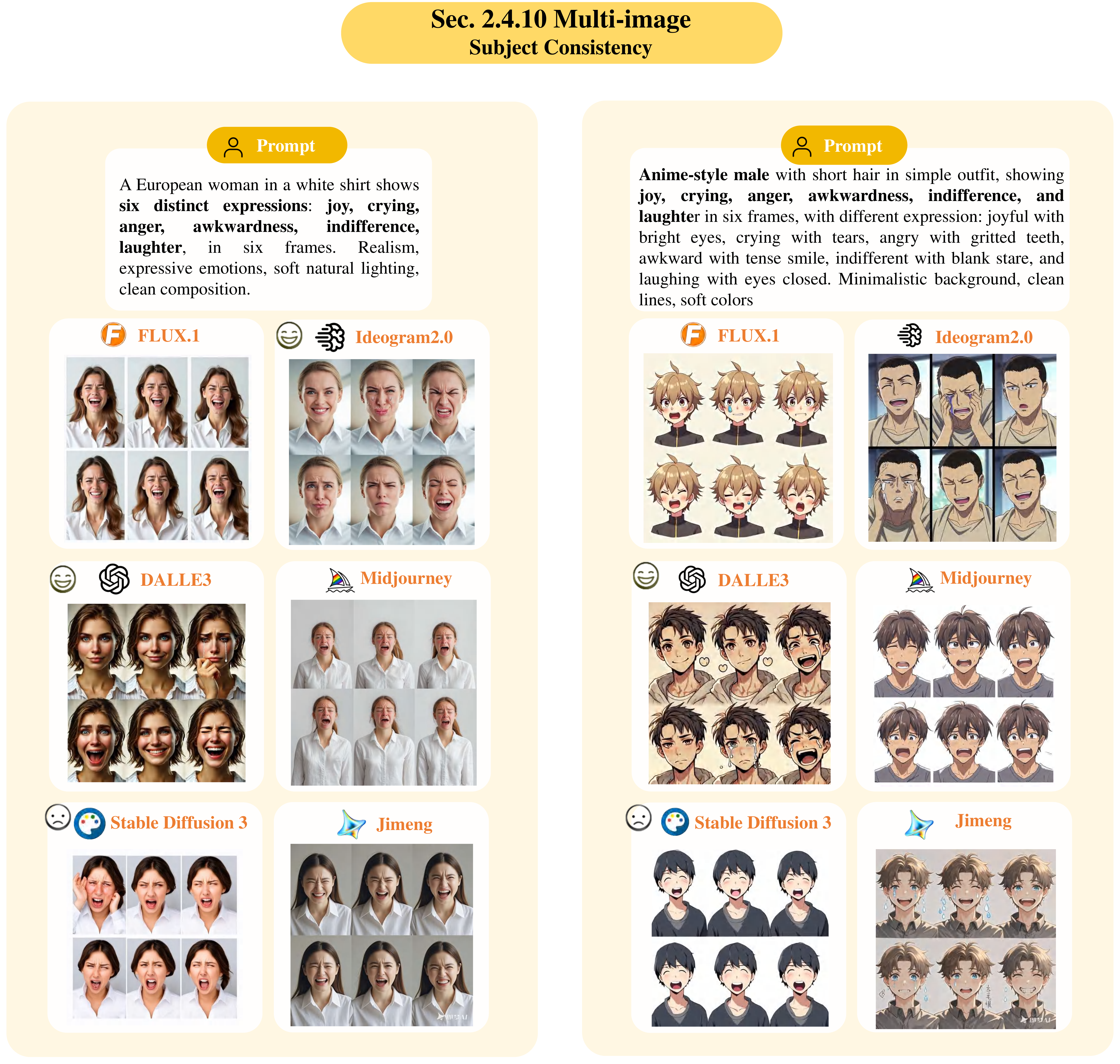}}
  \caption[Section~\ref{multi-image}: multi-image.]{Results on subject consistency task. Refer to Section \ref{multi-image} for detailed discussions.}
  \label{fig_facial-expression}
\end{figure*}

\begin{figure*}[!ht]
  \centering 
  \makebox[\textwidth][c]{\includegraphics[width=0.75\textwidth]{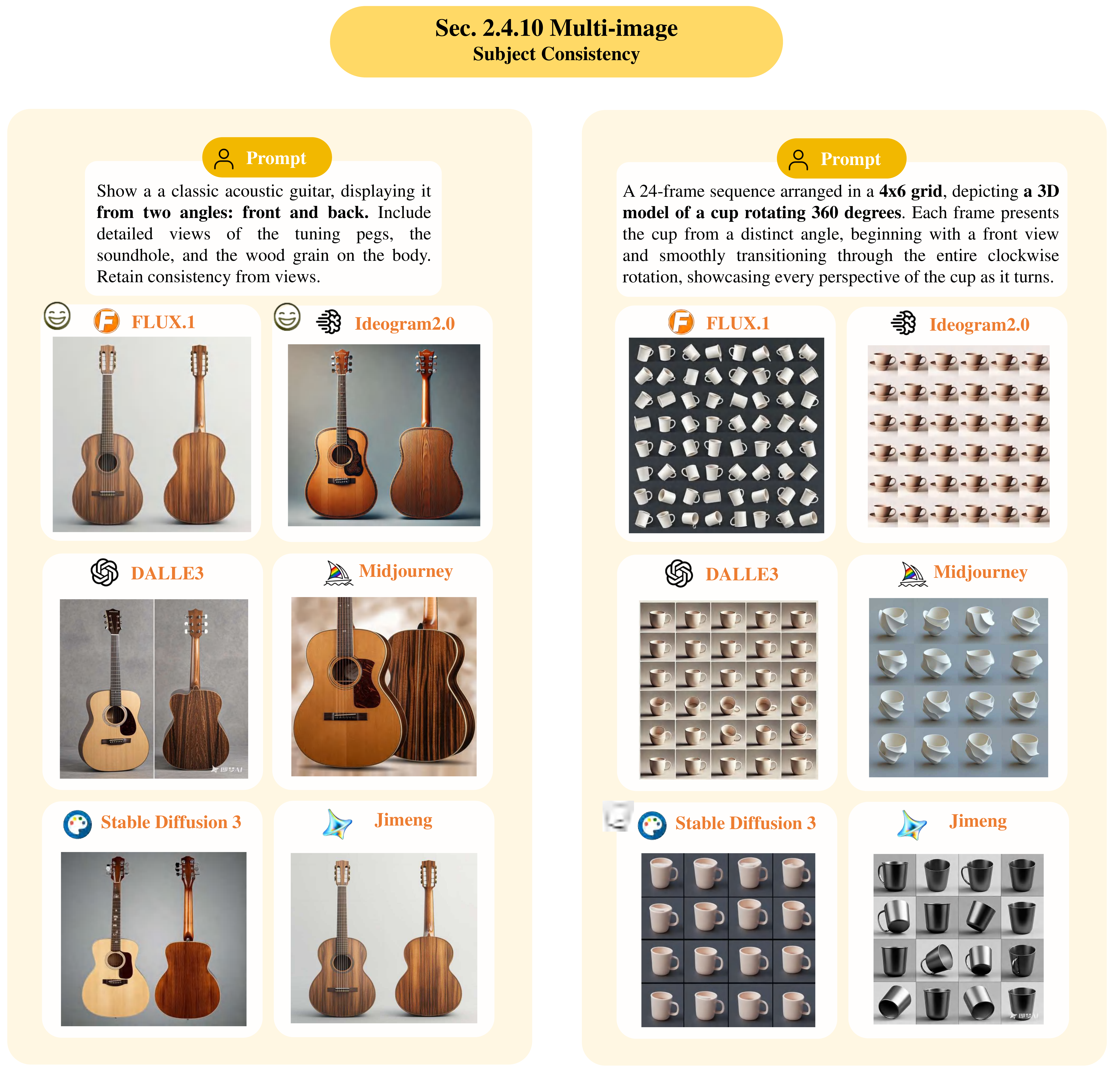}}
  \caption[Section~\ref{multi-image}: multi-image.]{Results on subject consistency task. Refer to Section \ref{multi-image} for detailed discussions.}
  \label{fig_multiview}
\end{figure*}

\textbf{Subject Consistency. }
In Figures \ref{fig_subject} to \ref{fig_multiview}, we tested the models' ability to generate multiple images with a consistent subject. In the first prompt (Figure \ref{fig_subject}), all models successfully generated a consistent subject—a dog—but only FLUX.1 was able to follow the prompt and generate detailed images in four grids. In the second prompt, all models performed well. In the third prompt, only FLUX.1 and Ideogram2.0 were able to follow the prompt correctly. 
In Figure \ref{fig_subject_2}, the subject is a chair and a logo. Most models were able to maintain subject consistency, with the exception of Stable Diffusion 3, Midjourney, and Jimeng, which produced images with inconsistencies.
In Figure \ref{fig_subject_3}, the subject is a woman. We changed both the background and the color of the woman's dress. Most models maintained subject consistency, with only Stable Diffusion 3 and Jimeng showing slight differences between the subjects in the two grids.

In Figure\ref{fig_facial-expression}, we tested the models to generate different facial expressions of the same person. Regardless of whether the person was real or anime-style, all models were able to maintain facial consistency relatively well. However, only Ideogram2.0 was able to generate the required expressions in the correct order with accuracy.

Regarding multi-view generation capabilities, as illustrated in the Figure~\ref{fig_multiview}, most models demonstrate a reasonable understanding and consistency when generating two perspectives. Regarding multi-view generation capabilities, as shown in Figure \ref{fig_multiview}, most models demonstrate reasonable understanding and consistency when generating two perspectives. However, challenges arise when generating multiple perspectives. With the exception of FLUX.1, other models frequently generate repeated perspectives.

\begin{figure*}[!ht]
\vspace{5em}
  \centering 
  \makebox[\textwidth][c]{\includegraphics[width=1\textwidth]{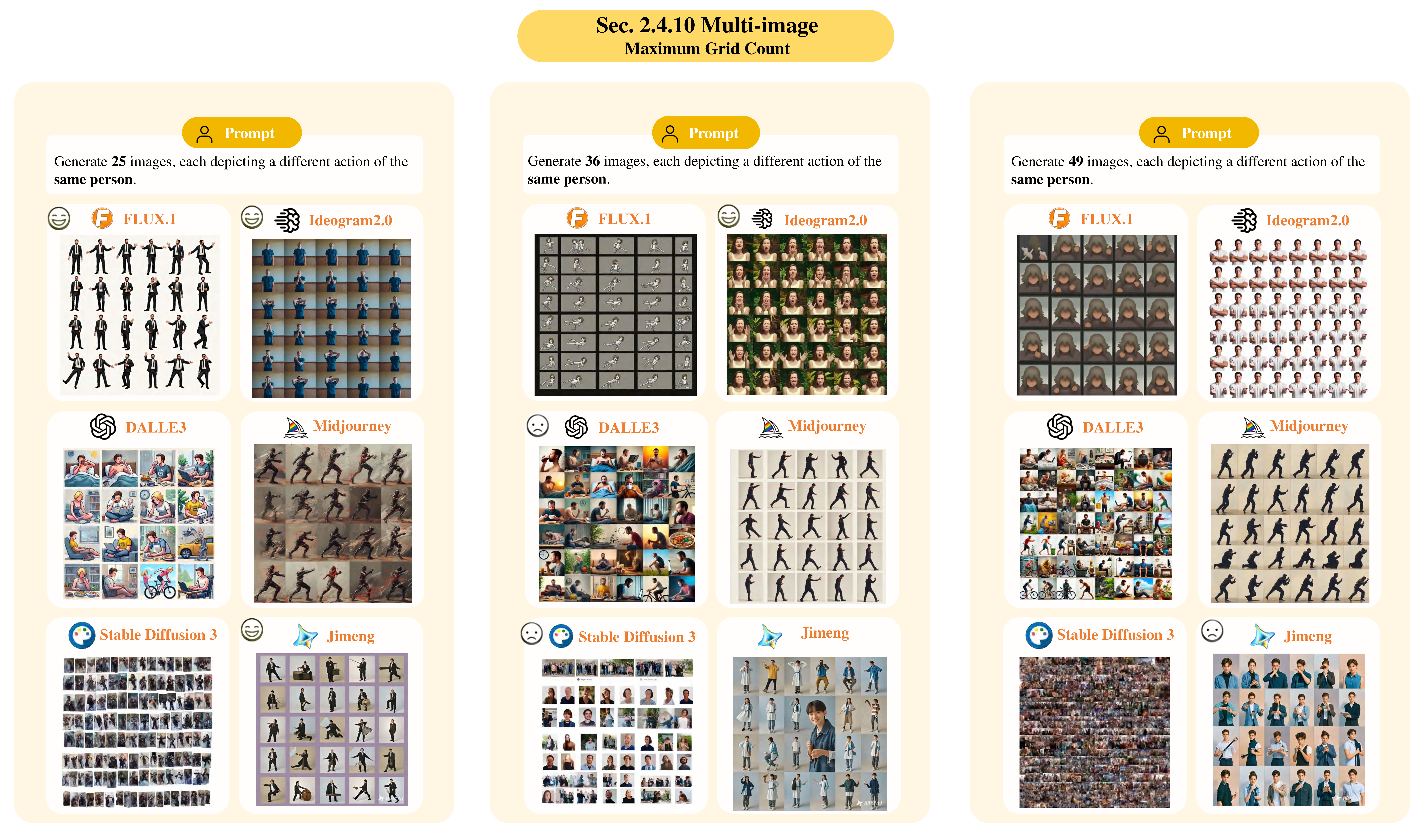}}
  \caption[Section~\ref{multi-image}: multi-image.]{Results on maximum grid count task. Refer to Section \ref{multi-image} for detailed discussions.}
  \label{fig_maxgrid}
\end{figure*}

\begin{figure*}[!ht]
\vspace{5em}
  \centering 
  \makebox[\textwidth][c]{\includegraphics[width=1\textwidth]{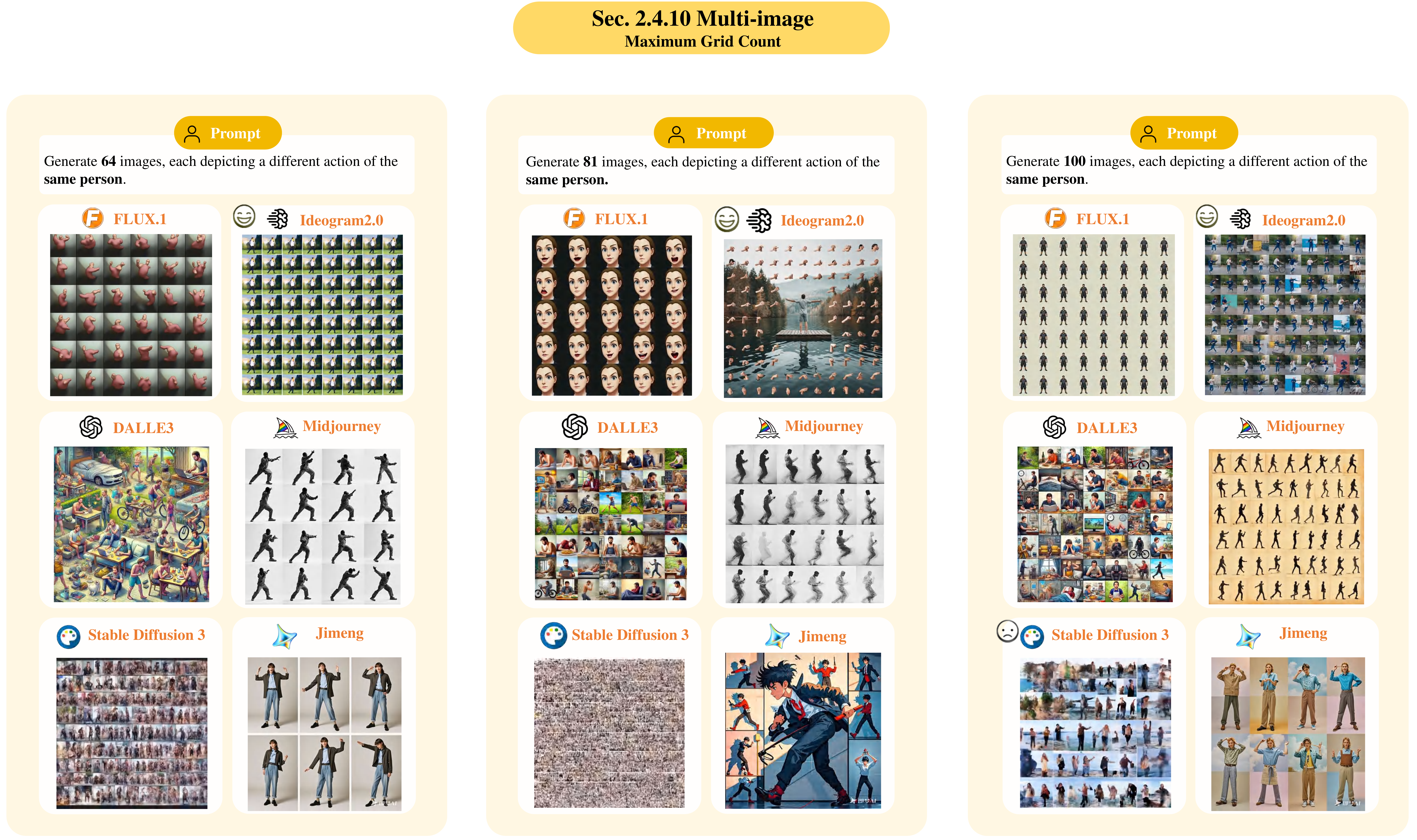}}
  \caption[Section~\ref{multi-image}: multi-image.]{Results on maximum grid count task. Refer to Section \ref{multi-image} for detailed discussions.}
  \label{fig_maxgrid_2}
\end{figure*}

\textbf{Macimum Grid Count. }
In Figure \ref{fig_maxgrid} and \ref{fig_maxgrid_2}, we tested the models to generate as many sub-images as possible within a single image. When the number of sub-images reached 25, Stable Diffusion 3 first exhibited confusion regarding the quantity. At 36 sub-images, FLUX.1 and Midjourney were also unable to generate the correct number of sub-images. At 49 and 64 sub-images, only Ideogram2.0 was able to produce results that were close to accurate. In experiments with more than 64 sub-images, none of the models could generate the correct number of sub-images.

\begin{figure*}[!ht]
  \centering 
  \makebox[\textwidth][c]{\includegraphics[width=1\textwidth]{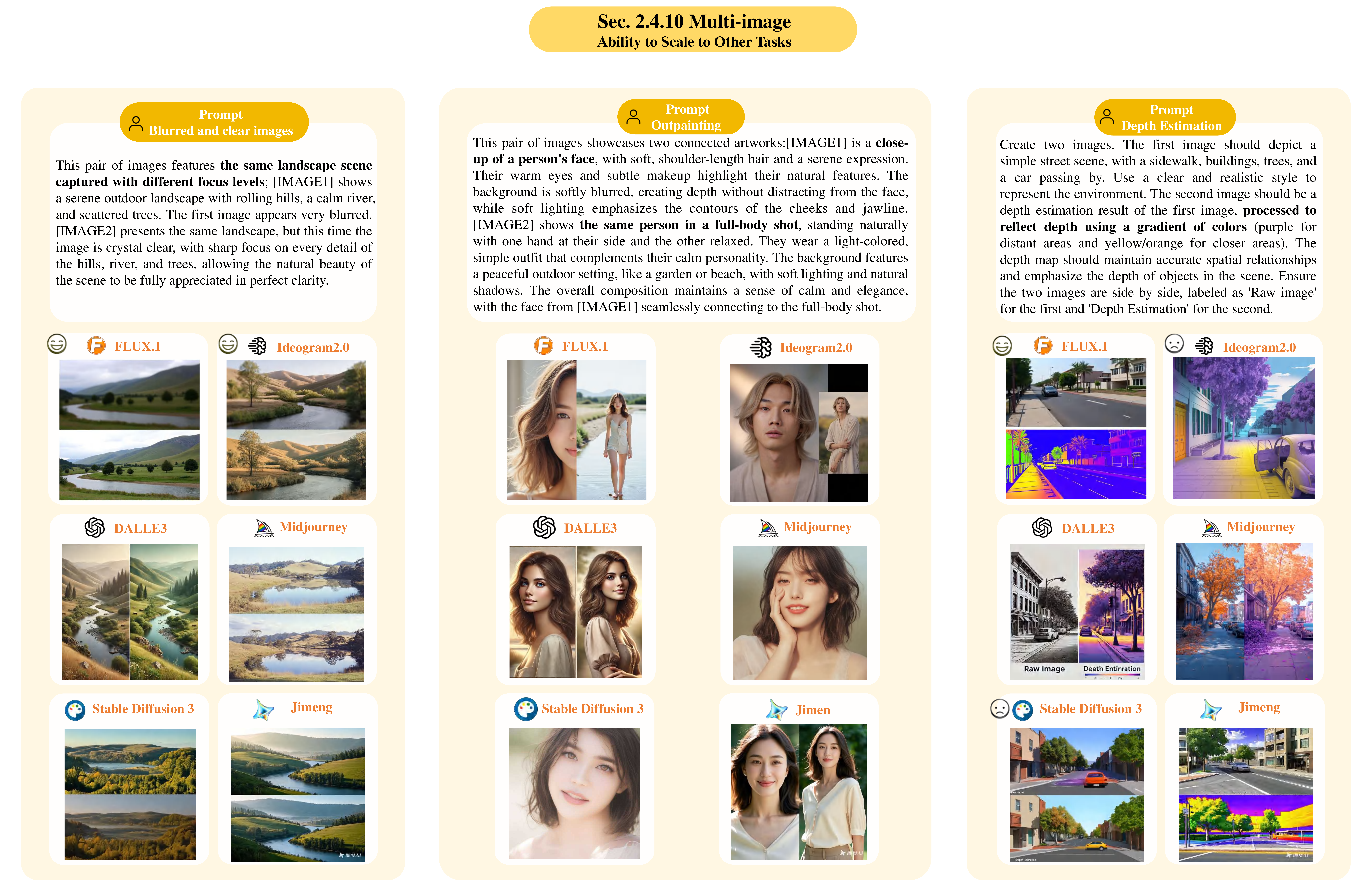}}
  \caption[Section~\ref{multi-image}: multi-image.]{Results on ability to scale to other tasks. Refer to Section \ref{multi-image} for detailed discussions.}
  \label{fig_othertask}
\end{figure*}

\begin{figure*}[!ht]
  \centering 
  \makebox[\textwidth][c]{\includegraphics[width=1\textwidth]{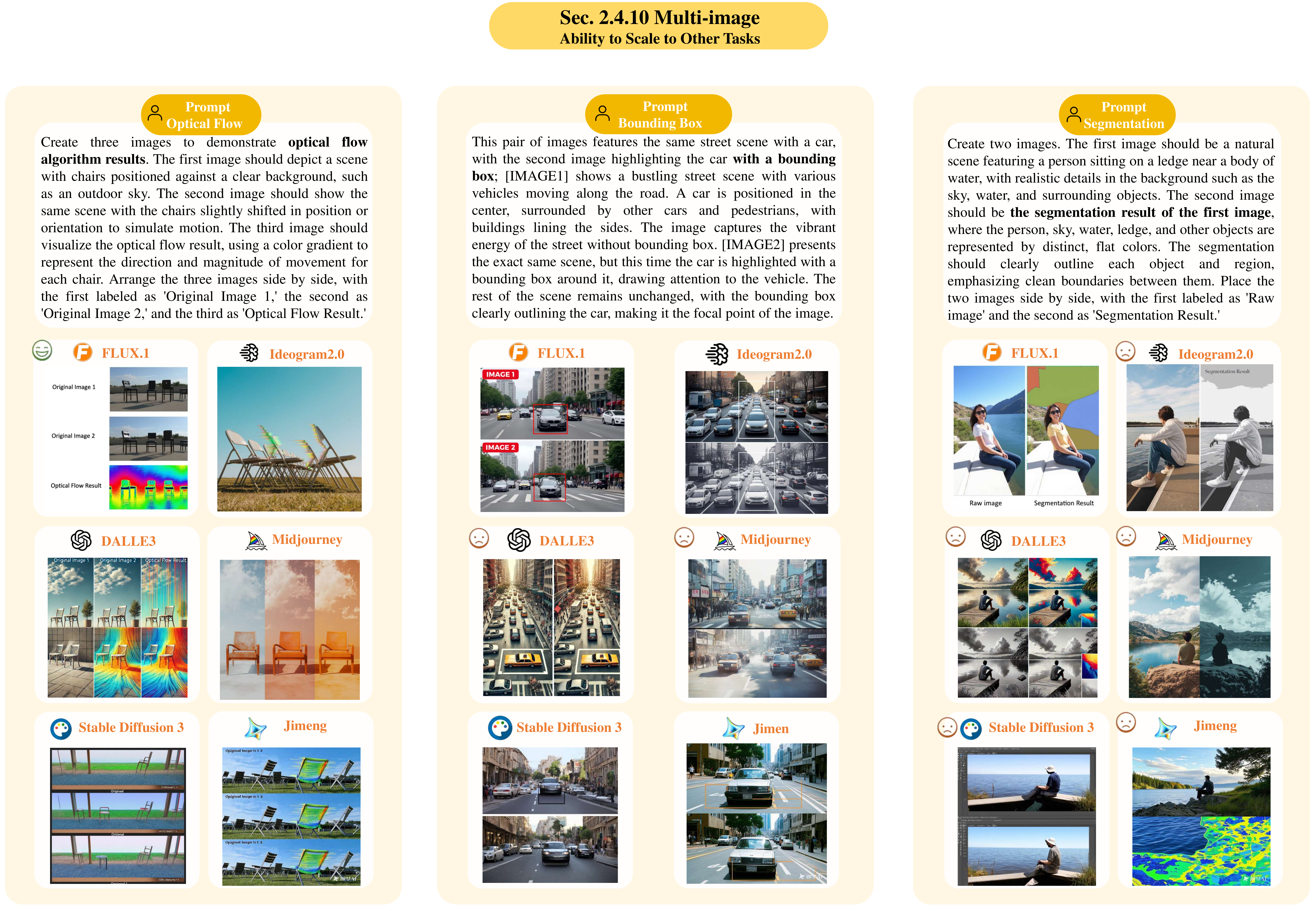}}
  \caption[Section~\ref{multi-image}: multi-image.]{Results on ability to scale to other tasks. Refer to Section \ref{multi-image} for detailed discussions.}
  \label{fig_othertask_2}
\end{figure*}

\textbf{Ability to Scale to Other Tasks. }
In this task, we explore the models' in-context generation ability when scaled to other tasks, as shown in Figures \ref{fig_othertask} and \ref{fig_othertask_2}. The models were tasked with generating two grids: the first one blurred and the second one clear. We found that FLUX.1 and Ideogram2.0 were able to accurately meet our requirements. We also extended the task to some computer vision tasks, such as depth estimation, optical flow, detection, and segmentation. However, the models struggled to achieve satisfactory results in these tasks.

\textbf{Score. }
In multi-image tasks, we found that FLUX.1 and Ideogram 2.0 performed exceptionally well across multiple tasks, demonstrating the potential of text-to-image models for in-context generation.
\begin{table}[h]
    \centering
    \caption[Section~\ref{multi-image}: Multi-image.]{The scoring of generation results by six models on multi-image under different evaluation systems. Refer to Section \ref{multi-image} for detailed discussions.}
    \begin{tabular}{l|c|c|c|c|c}
        \midrule
        Model & CLIPScore & HPSv2 & Aesthetic Score & GPT-4o & Human \\
        \midrule 
        FLUX.1      & 26.99 & 0.25 & 5.69 &  \textbf{6.30} & \textbf{9.63} \\
        Ideogram2.0 & 27.15 & 0.25 & 5.65 &  6.02 & 9.35 \\
        Dall-E3     & 26.92 & 0.26 & 5.68 &  5.83 & 9.07 \\
        Midjourney  & 22.08 & \textbf{0.28} & 5.68 &  5.46 & 8.98 \\
        SD3         & 25.38 & \textbf{0.28} & 5.65 &  5.83 & 8.24 \\
        Jimeng      & \textbf{27.71} & 0.26 & \textbf{5.90} &  5.37 & 8.24 \\
        \midrule
    \end{tabular}
    \label{tab_ood}
\end{table}

\subsubsection{Text Writing}
\label{text}
\begin{figure*}[!ht]
  \centering 
  \makebox[\textwidth][c]{\includegraphics[width=1\textwidth]{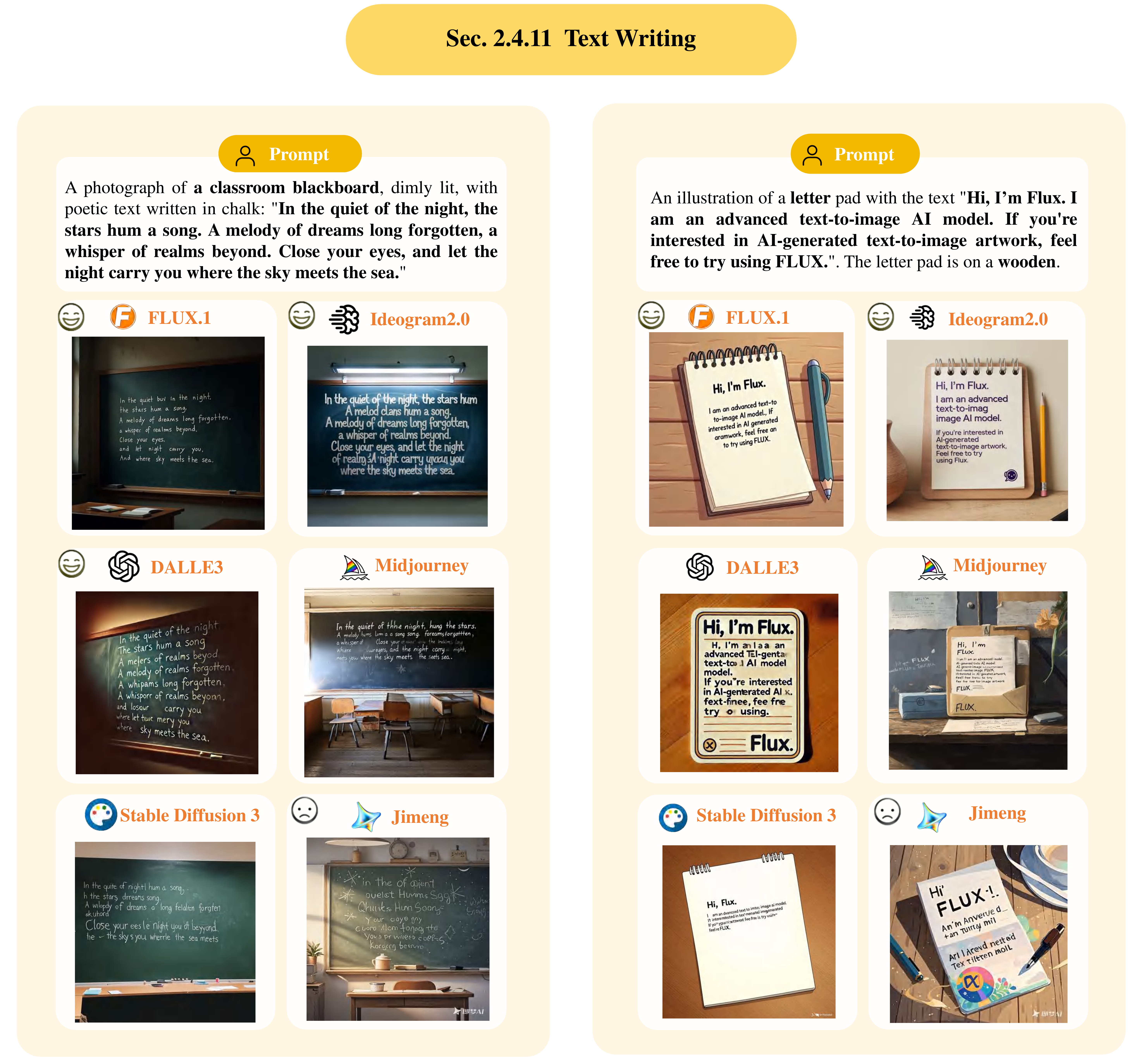}}
  \caption[Section~\ref{text}: text writing.]{Results on text writing task. Refer to Section \ref{text} for detailed discussions.}
  \label{fig_text}
\end{figure*}

In Figures \ref{fig_text}, we examine the models' ability to generate images containing accurate text. The results demonstrate that none of the models consistently produce correct text within images. Specifically, Jimeng fails to generate correct words, instead producing distorted or illegible characters. Dall-E3, Midjourney, and Stable Diffusion 3 occasionally include text with unreadable characters. FLUX.1 and Ideogram2.0 show better performance, though they still generate text with misspelled words or incomplete phrases.

\textbf{Score. }
The results of this experiment are shown in the Table \ref{tab_text}, where FLUX.1 and Ideogram2.0 performed the best. Only the results of HPSv2 aligned closely with human perception, possibly because the evaluation of this task requires a certain understanding of the text information in the images.
\begin{table}[h]
    \centering
    \caption[Section~\ref{text}: text.]{The scoring of generation results by six models on text writing under different evaluation systems. Refer to Section \ref{text} for detailed discussions.}
    \begin{tabular}{l|c|c|c|c|c}
        \midrule
        Model & CLIPScore & HPSv2 & Aesthetic Score &  GPT-4o & Human \\
        \midrule 
        FLUX.1      & 37.01 & \textbf{0.29} & 4.34 &  5.83 & \textbf{8.33} \\
        Ideogram2.0 & 35.33 & 0.28 & 4.94 &  5.00 & \textbf{8.33} \\
        Dall-E3     & 35.18 & 0.25 & 3.94 &  4.17 & 6.67 \\
        Midjourney  & 42.27 & 0.28 & 4.19 &  \textbf{6.67} & 5.00 \\
        SD3         & \textbf{46.14} & 0.28 & 4.23 &  3.33 & 3.33 \\
        Jimeng      & 36.03 & 0.25 & \textbf{5.81} &  5.00 & 3.33 \\
        \midrule
    \end{tabular}
    \label{tab_text}
\end{table}

\clearpage
\subsection{Multi-style Creation Task}
\label{style}
\begin{figure*}[!ht]
\vspace{-1em}
  \centering 
  \makebox[\textwidth][c]{\includegraphics[width=0.8\textwidth]{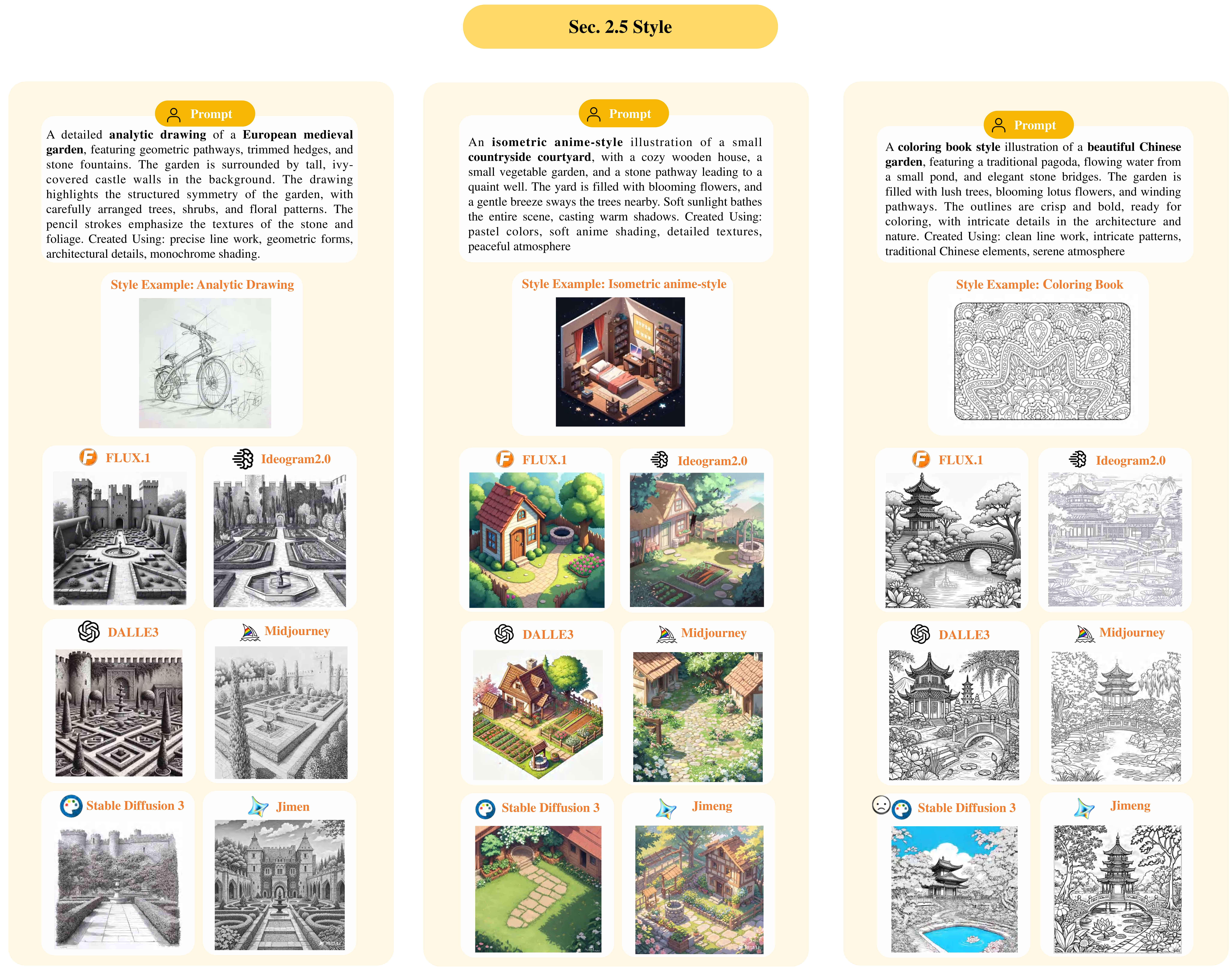}}
  \caption[Section~\ref{style}: style.]{Results on different style image generation task. Refer to Section \ref{style} for detailed discussions.}
  \label{fig_style_1}
\end{figure*}

\begin{figure*}[!ht]
  \centering 
  \makebox[\textwidth][c]{\includegraphics[width=0.9\textwidth]{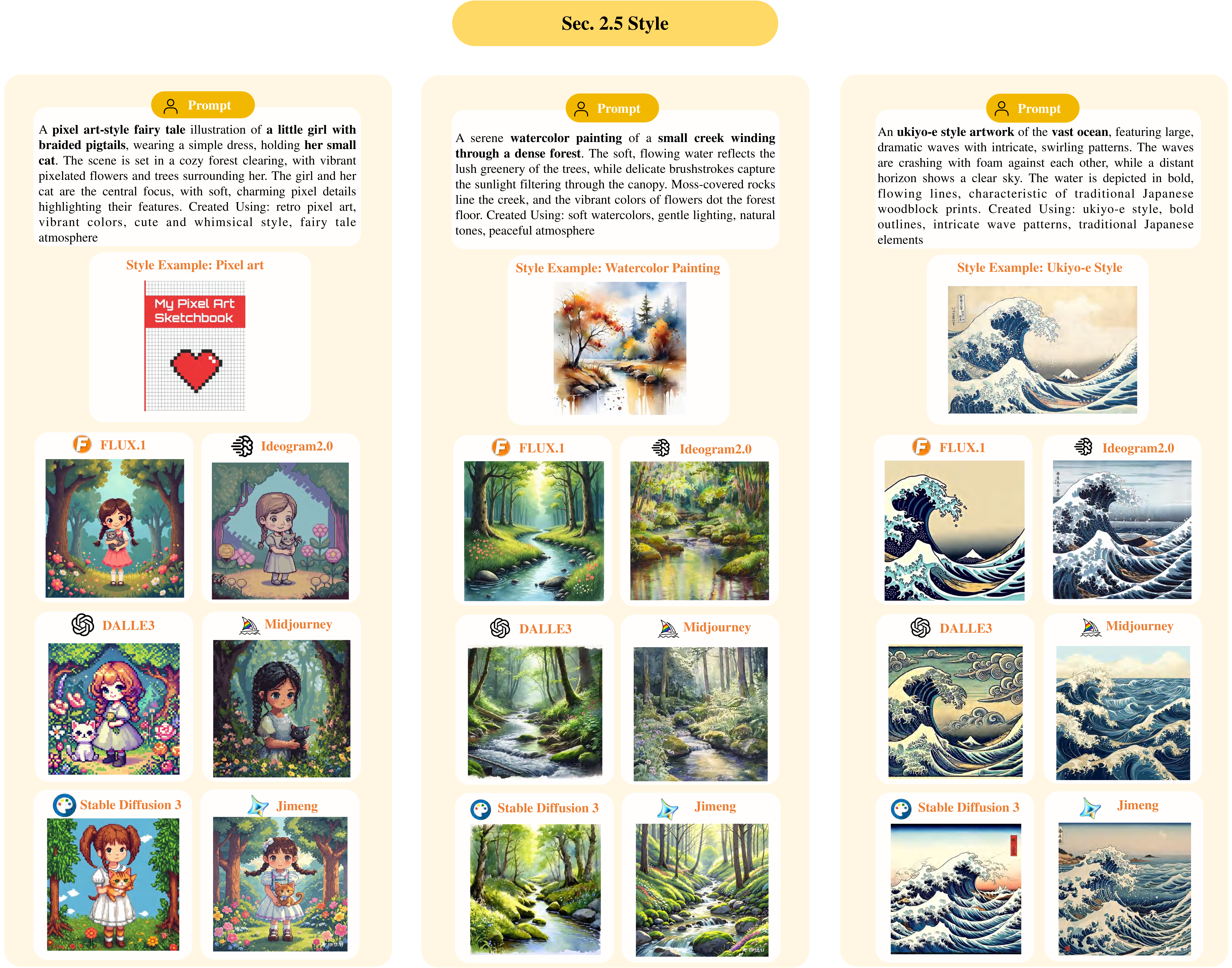}}
  \caption[Section~\ref{style}: style.]{Results on different style image generation task. Refer to Section \ref{style} for detailed discussions.}
  \label{fig_style_2}
\end{figure*}

\begin{figure*}[!ht]
  \centering 
  \makebox[\textwidth][c]{\includegraphics[width=0.9\textwidth]{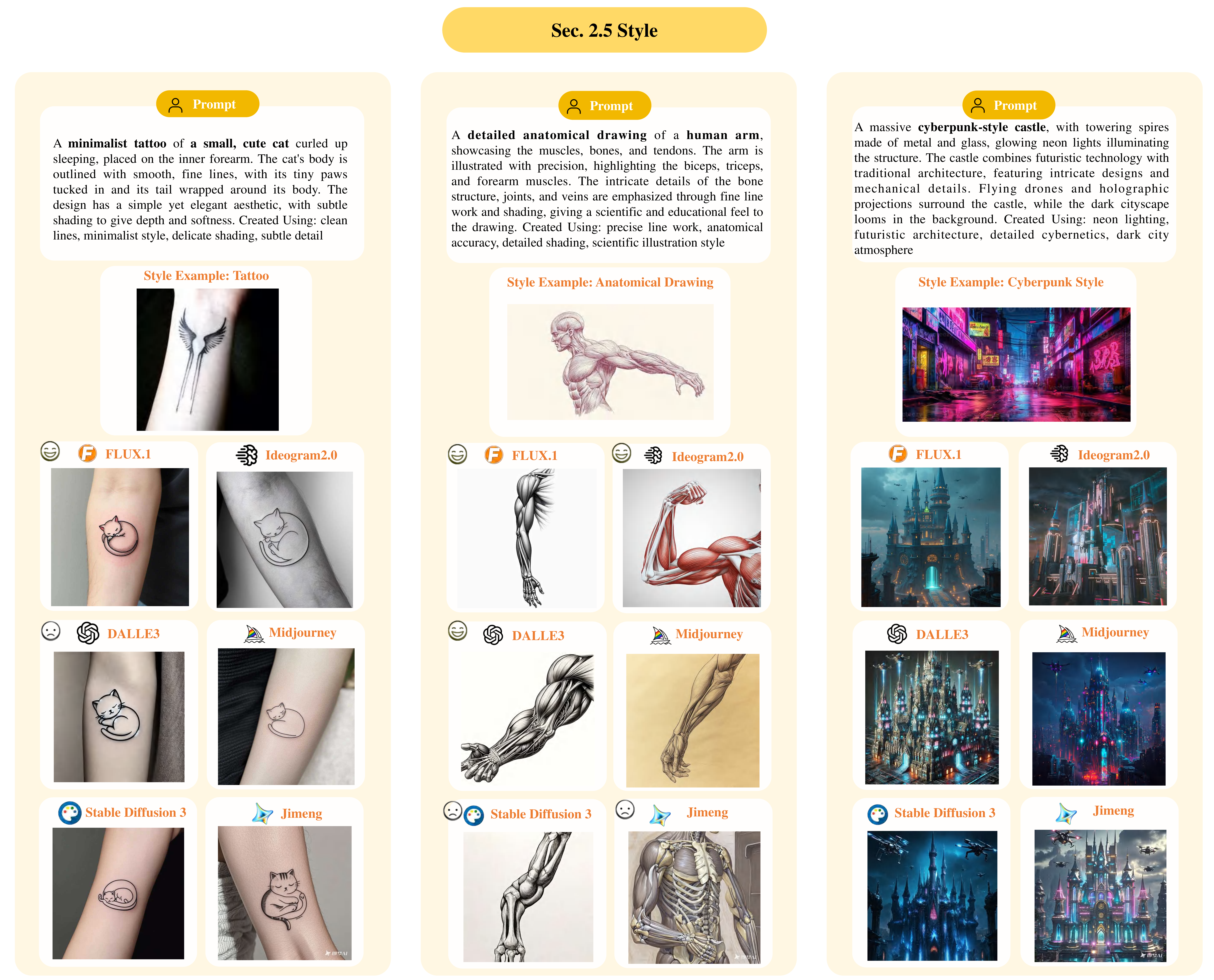}}
  \caption[Section~\ref{style}: style.]{Results on different style image generation task. Refer to Section \ref{style} for detailed discussions.}
  \label{fig_style_3}
\end{figure*}

\begin{figure*}[!ht]
  \centering 
  \makebox[\textwidth][c]{\includegraphics[width=0.9\textwidth]{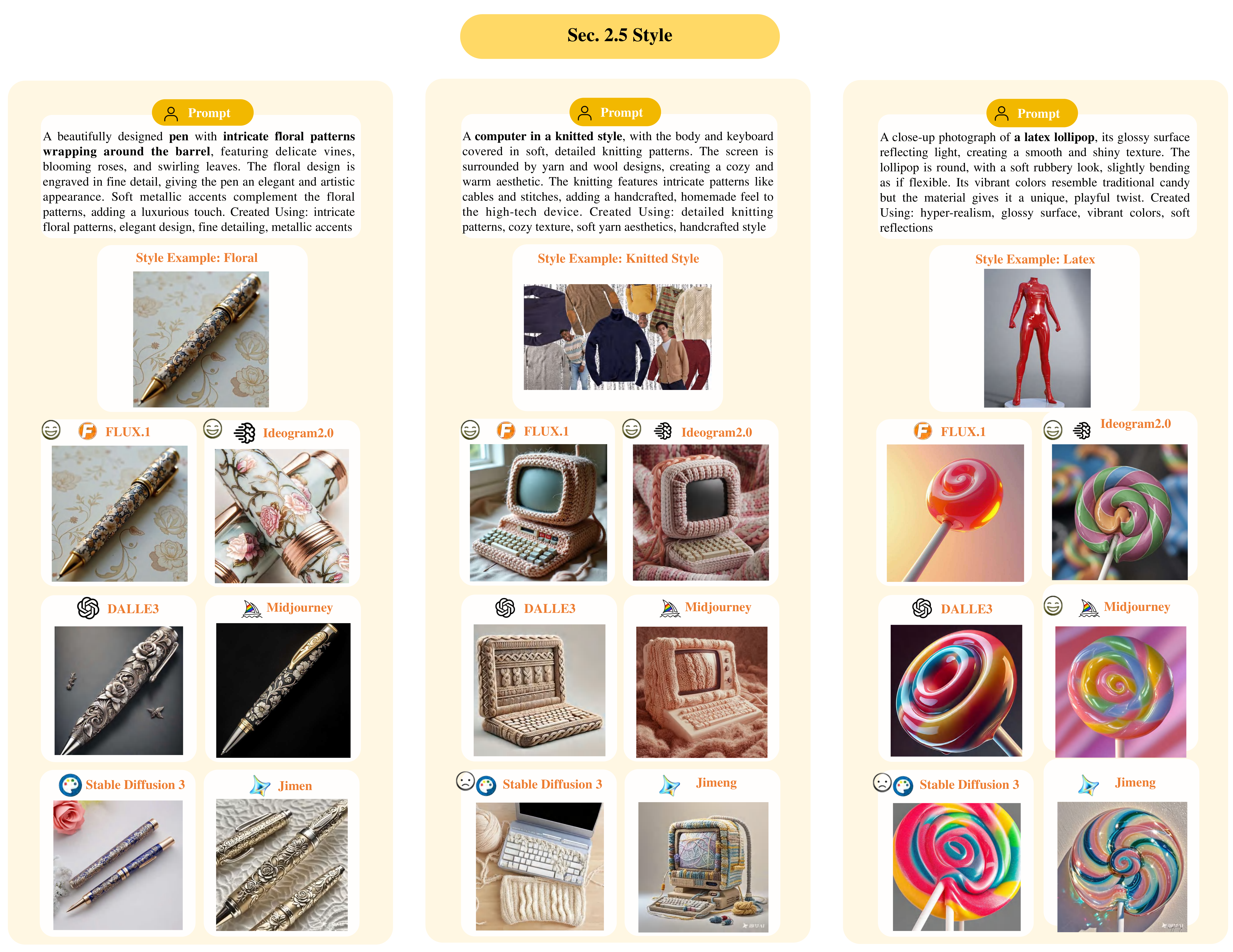}}
  \caption[Section~\ref{style}: style.]{Results on different style image generation task. Refer to Section \ref{style} for detailed discussions.}
  \label{fig_style_4}
\end{figure*}

\begin{figure*}[!ht]
  \centering 
  \makebox[\textwidth][c]{\includegraphics[width=0.9\textwidth]{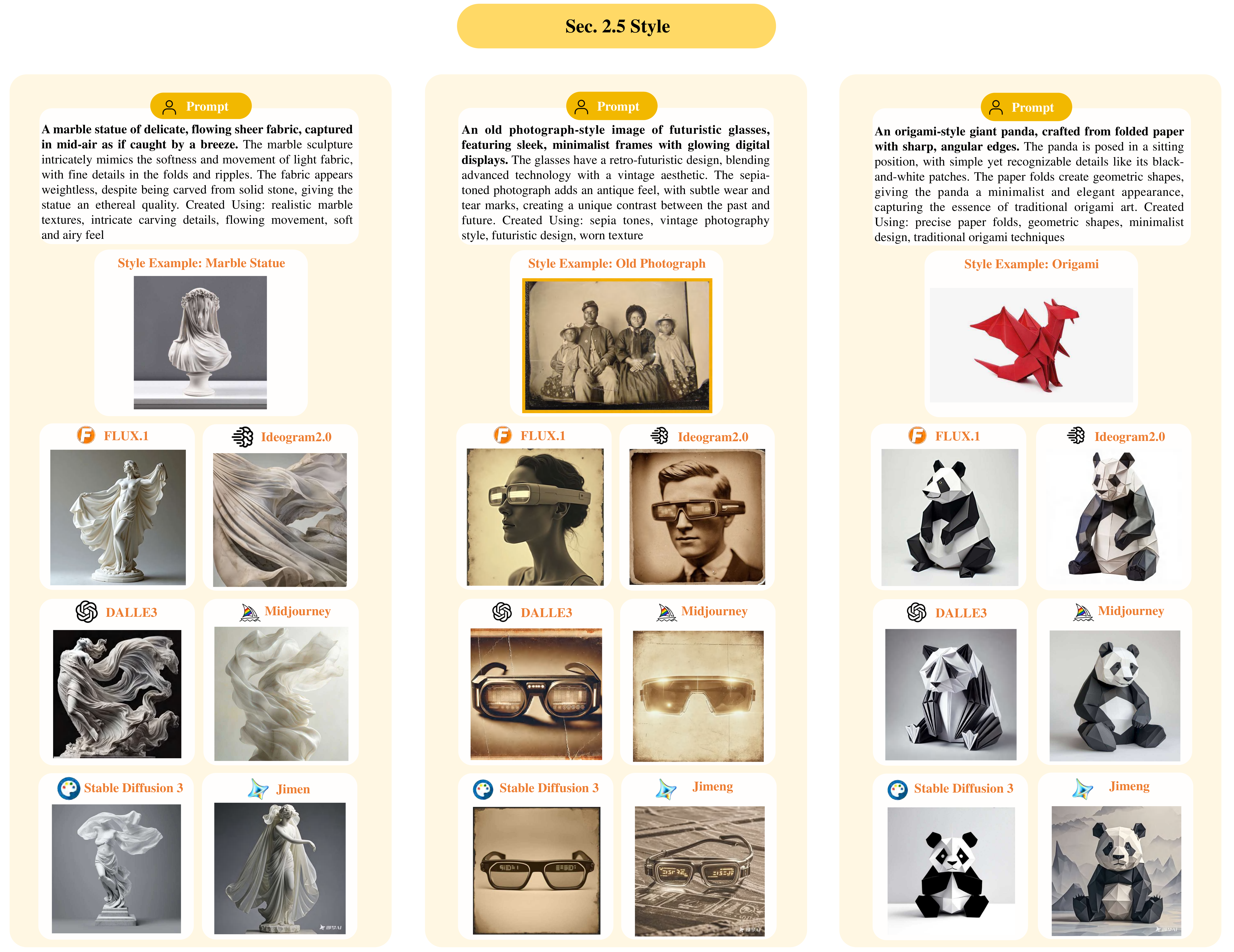}}
  \caption[Section~\ref{style}: style.]{Results on different style image generation task. Refer to Section \ref{style} for detailed discussions.}
  \label{fig_style_5}
\end{figure*}

\begin{figure*}[!ht]
  \centering 
  \makebox[\textwidth][c]{\includegraphics[width=0.9\textwidth]{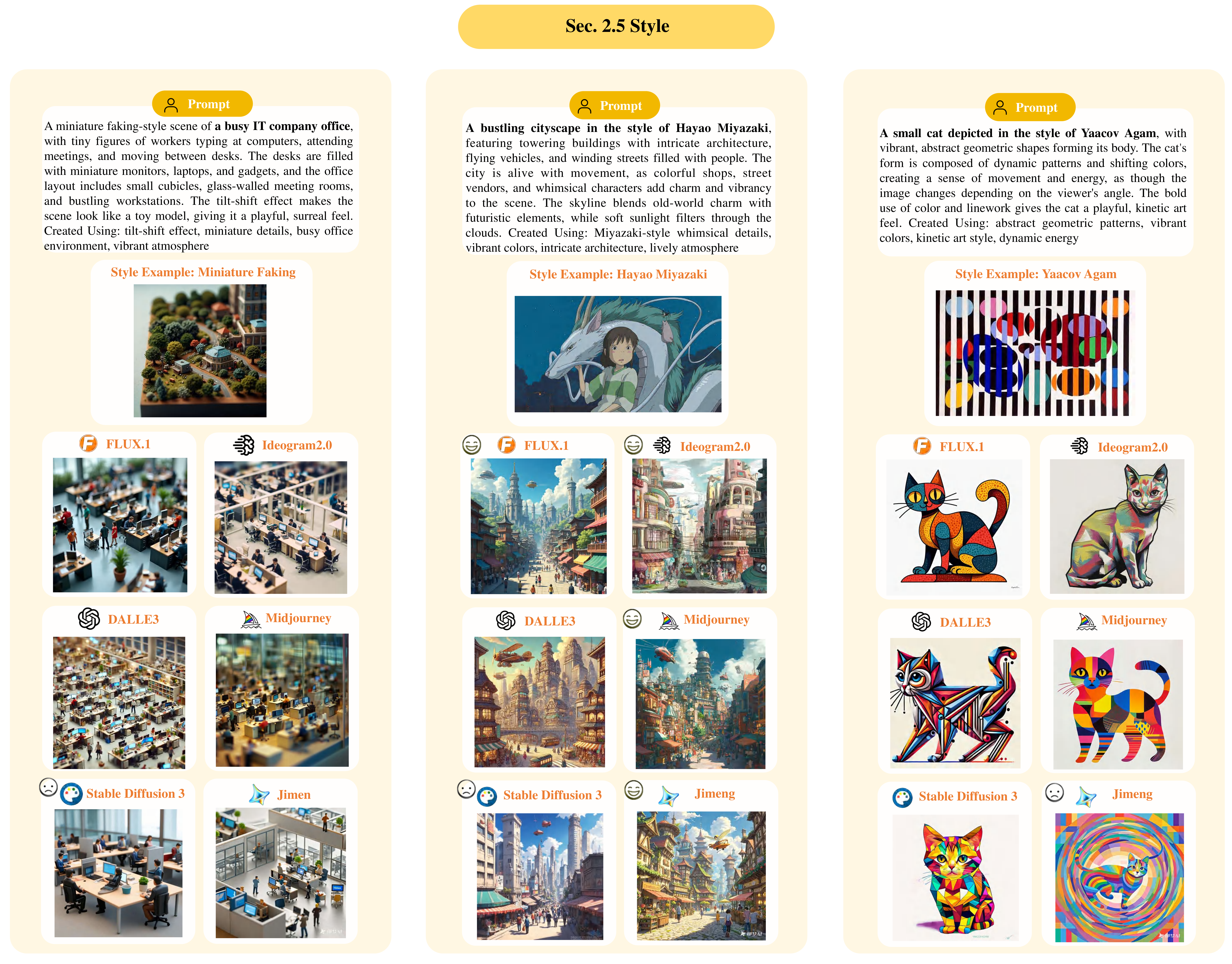}}
  \caption[Section~\ref{style}: style.]{Results on different style image generation task. Refer to Section \ref{style} for detailed discussions.}
  \label{fig_style_6}
\end{figure*}

\begin{figure*}[!ht]
  \centering 
  \makebox[\textwidth][c]{\includegraphics[width=0.9\textwidth]{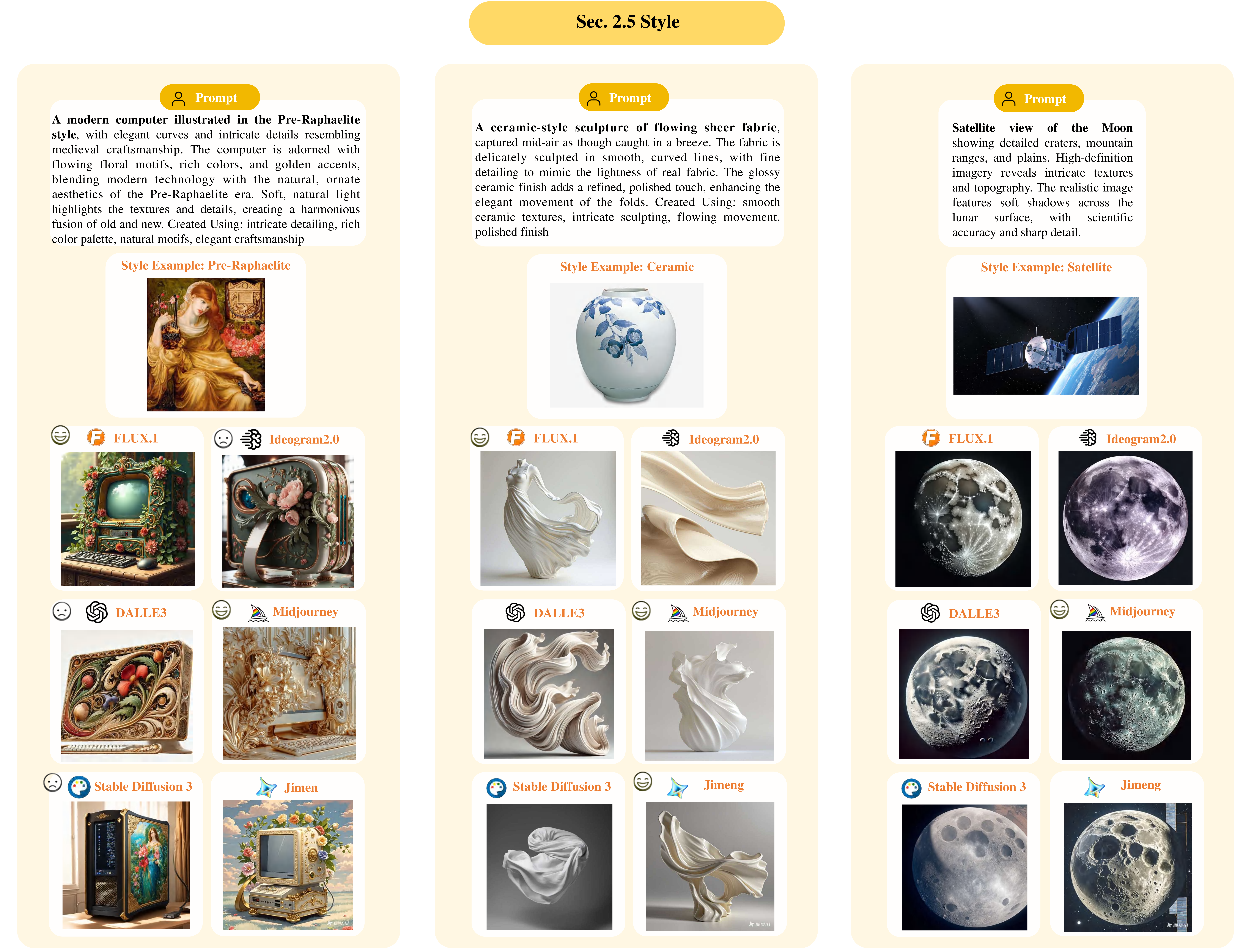}}
  \caption[Section~\ref{style}: style.]{Results on different style image generation task. Refer to Section \ref{style} for detailed discussions.}
  \label{fig_style_7}
\end{figure*}

\begin{figure*}[!ht]
  \centering 
  \makebox[\textwidth][c]{\includegraphics[width=0.9\textwidth]{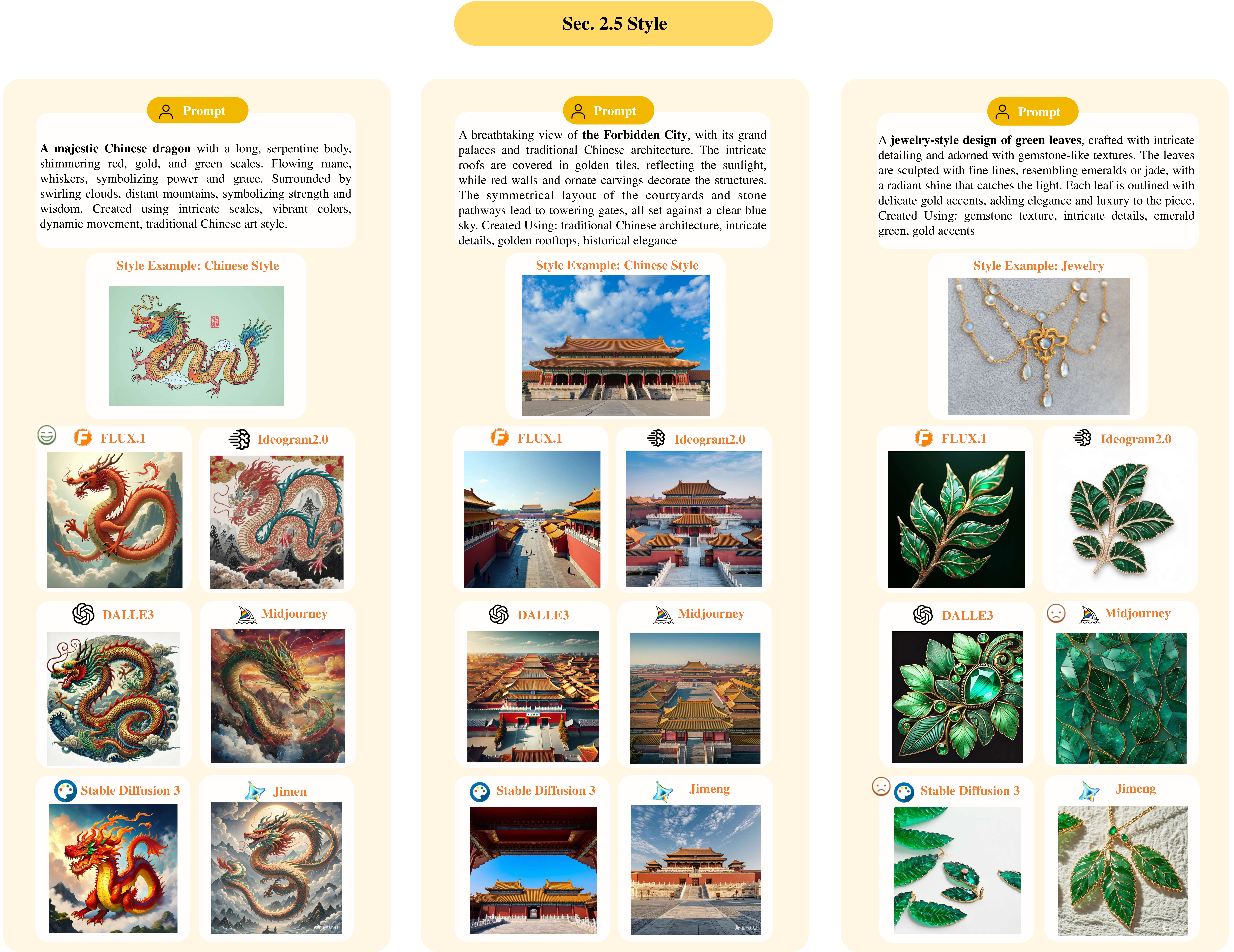}}
  \caption[Section~\ref{style}: style.]{Results on different style image generation task. Refer to Section \ref{style} for detailed discussions.}
  \label{fig_style_8}
\end{figure*}

\begin{figure*}[!ht]
  \centering 
  \makebox[\textwidth][c]{\includegraphics[width=0.9\textwidth]{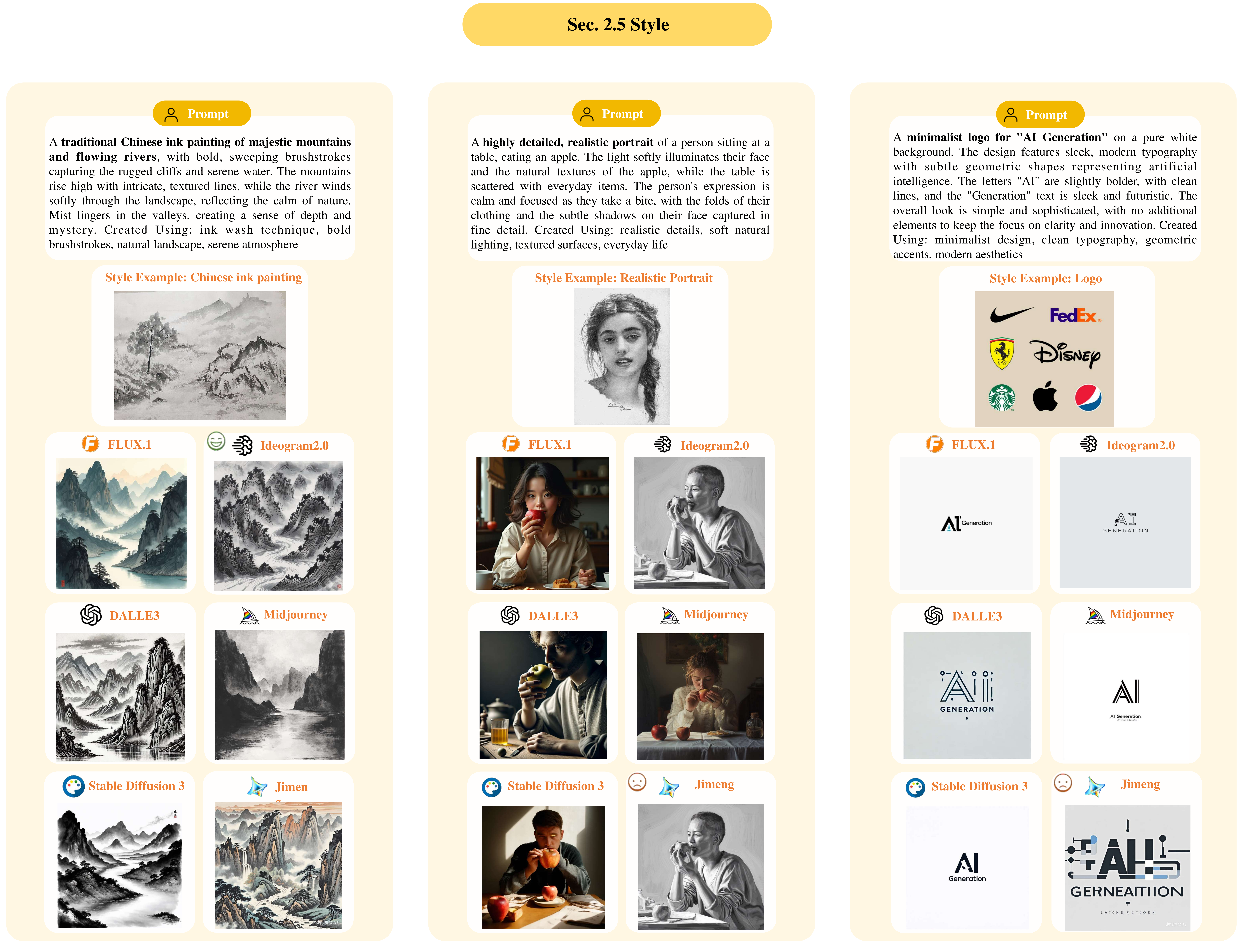}}
  \caption[Section~\ref{style}: style.]{Results on different style image generation task. Refer to Section \ref{style} for detailed discussions.}
  \label{fig_style_9}
\end{figure*}

\begin{figure*}[!ht]
  \centering 
  \makebox[\textwidth][c]{\includegraphics[width=0.9\textwidth]{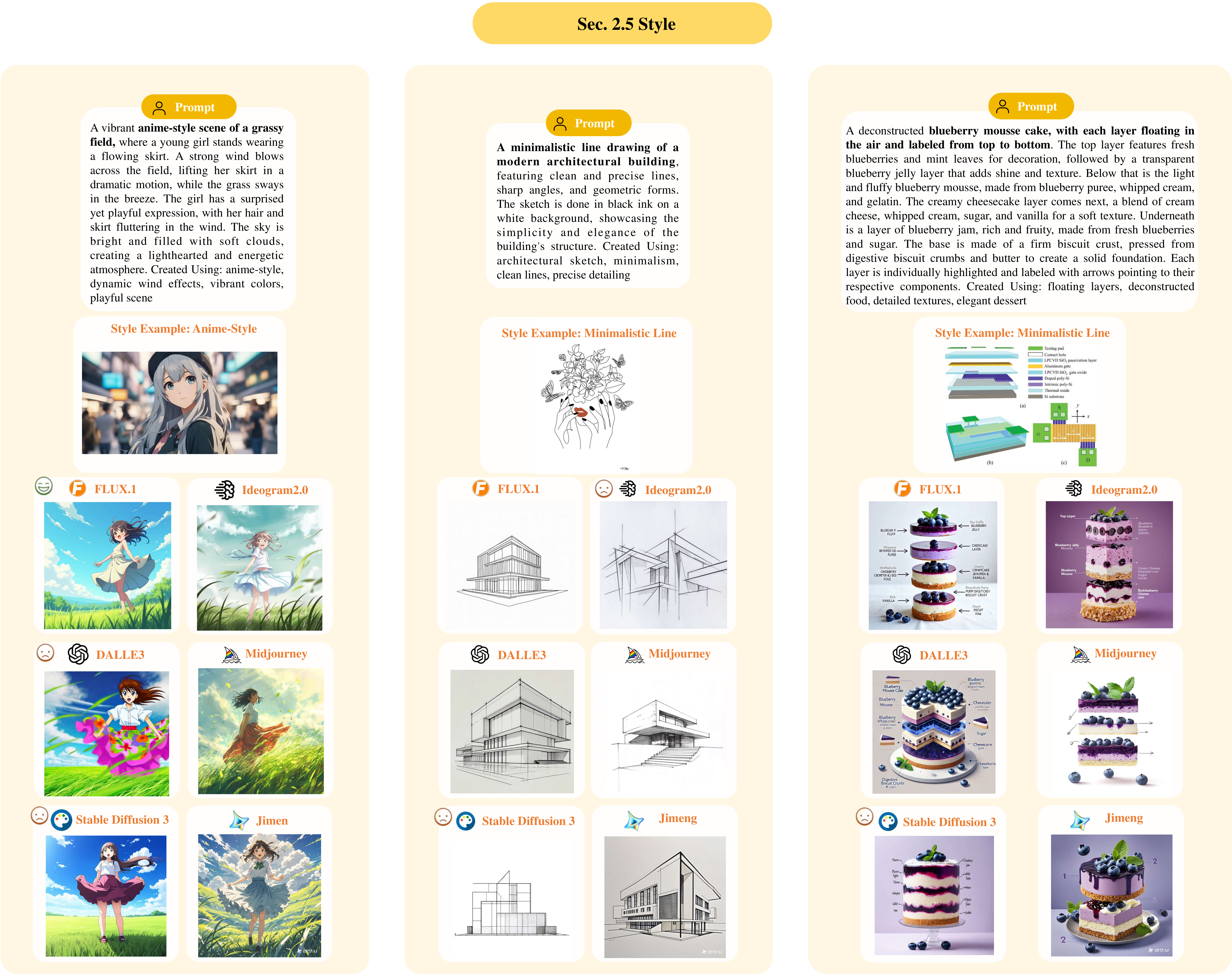}}
  \caption[Section~\ref{style}: style.]{Results on different style image generation task. Refer to Section \ref{style} for detailed discussions.}
  \label{fig_style_10}
\end{figure*}

\begin{figure*}[!ht]
  \centering 
  \makebox[\textwidth][c]{\includegraphics[width=0.9\textwidth]{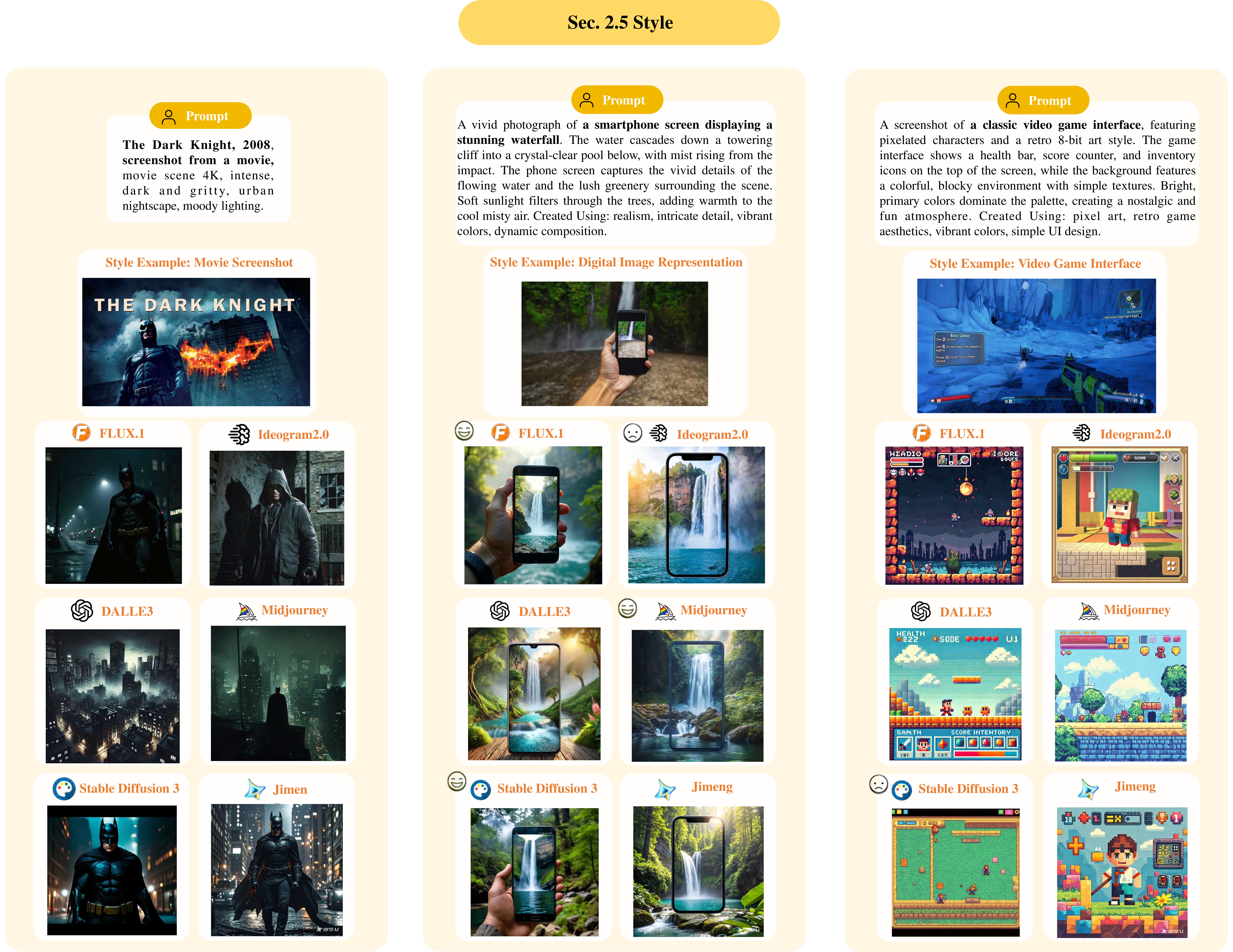}}
  \caption[Section~\ref{style}: style.]{Results on different style image generation task. Refer to Section \ref{style} for detailed discussions.}
  \label{fig_style_11}
\end{figure*}

In this experiment, we selected over 30 commonly used styles in text-to-image tasks and tailored the prompts accordingly. Additionally, we examined the model's imagination by combining seemingly contradictory styles and objects, such as floating chiffon in a marble texture, to test how well the model can generate coherent and creative images from these combinations.
We evaluated the generation results from six models, with detailed scores presented in Table \ref{tab_style} and visualizations provided in Figures \ref{fig_style_1}-\ref{fig_style_11}. 

We found that the performance of the models in this task did not vary significantly, but the images generated by Midjourney often exhibited greater aesthetic appeal and better alignment with human perception of beauty. For example, in the middle subplot of Figure \ref{fig_style_1}, all models can produce images in the isometric anime style, but the images generated by Midjourney and Jimeng stand out for their aesthetic quality. Similarly, in the left subplot of Figure \ref{fig_style_5}, only Midjourney is able to generate a marble sculpture that conveys the most fluidity and grace.

\textbf{Score.} Notably, while Stable Diffusion 3~\cite{rombach2022high} demonstrated room for improvement, the other models exhibited relatively similar performance. Among the six models, Midjourney exhibited superior generalization capabilities and higher aesthetic quality, securing higher scores. The results from GPT-4o were closely aligned with human aesthetic judgments, while the outputs from other evaluation systems showed notable discrepancies from human perceptions of aesthetics.

\begin{table}[h]
    \centering
    \caption[Section~\ref{style}: style.]{The scoring of generation results by six models on different style image generation under different evaluation systems. Refer to Section \ref{style} for detailed discussions.}
    \begin{tabular}{l|c|c|c|c|c}
        \midrule
        Metric & CLIPScore & HPSv2 & Aesthetic Score & GPT-4o & Human \\
        \midrule 
        FLUX.1 & 30.49 & \textbf{0.29} & 6.07 & \textbf{7.13} & 9.26 \\
        Ideogram2.0 & 30.96 & \textbf{0.29} & \textbf{6.12} & 6.69 & 9.26 \\
        Dall-E3 & \textbf{31.10} & \textbf{0.29} & 6.04 & 6.94 & 9.44 \\
        Midjourney & 30.26 & 0.28 & 5.99 & \textbf{7.13} & \textbf{9.61} \\
        SD3 & 29.51 & 0.28 & 6.01 & 6.89 & 8.68 \\
        Jimeng & 30.55 & 0.28 & 5.98 & 6.30 & 9.36 \\
        \midrule
    \end{tabular}
    \label{tab_style}
\end{table}

\clearpage
\section{Conclusion}

\subsection{Summary}
In this report, we conducted a comprehensive evaluation of six powerful text-to-image models, termed \imaginee, including FLUX.1, Ideogram2.0, Dall-E3, Midjourney, Stable Diffusion 3, and Jimeng. We extensively explored the performance of these models across various levels of difficulty, including qualitative samples and quantitative benchmarks. To make our evaluation more thorough, we carefully collected numerous samples covering six aspects: structured output generation, realism and physical consistency tasks, specific domain generation, challenging scenario generation, and different style image generation. Each domain also includes several more detailed subtasks for in-depth discussion and analysis.

\subsection{Task Complexity Analysis}
In this work, unlike traditional text-to-image generation, we focus on exploring and tackling challenging and specialized image generation domains to test the robustness and potential of generic T2I models.

Currently, all models still face difficulties in code generation tasks, including generating simple Python code, QR codes, and barcodes, indicating that applying text-to-image models to general code generation remains a significant challenge. Moreover, in 3D generation tasks, all models perform poorly, and the generated images cannot be directly used for 3D-related applications. In structured output tasks, none of the models can accurately follow prompts to generate images or tables. Although FLUX.1 can generate chart images that are close to correct, there are still discrepancies in detail compared to the prompts. Additionally, all models are unable to output images containing Chinese text.

However, all models can accept JSON-format inputs and generate images according to the specified JSON content. In some basic tasks, such as generating images in different styles or based on photographic terminology, the models perform well and can produce high-quality images following the prompts.

In summary, the current performance of these models in specific image generation domains still faces some common bottlenecks: on the one hand, there are certain unachieved functionalities or difficult tasks to overcome, and on the other hand, there are some basic capabilities that have already been realized. Further analysis is needed to identify the challenges of each task in the current design.

\subsection{Model Performance Evaluation}
\textbf{FLUX.1 and Ideogram2.0.} Overall, FLUX.1 and Ideogram2.0 perform the best. In structured output tasks, FLUX.1's outputs can closely match the images, tables, and web pages described in the prompts. In realism and physical consistency tasks, both models can largely address issues of human deformities and have a basic understanding of the fundamental laws of the physical world, though the generated images still exhibit some logical inconsistencies. In specific domain generation tasks, these two models excel in understanding foundational knowledge in disciplines such as chemistry, biology, and medicine, showing potential as general models, and can generate high-quality data for autonomous driving and embodied intelligence tasks. In challenging scenario generation tasks, FLUX.1 and Ideogram2.0 perform exceptionally well in generating content with dense text, accepting inputs in different languages and emojis.

\textbf{Midjourney.} Midjourney produces images with the best aesthetic appeal, particularly evident in generating images of different styles. In everyday contexts, Midjourney's images are more visually pleasing, making them more practical.

\textbf{Dall-E3.} Dall-E3 has a profound understanding of the physical world and boasts higher safety measures, being sensitive to copyright, watermarks, and NSFW prompts and images. Additionally, the images generated by Dall-E3 have a distinct style that sets them apart from all other models.

\textbf{Stable Diffusion 3 and Jimeng.} Stable Diffusion 3 and Jimeng do not match FLUX.1 in overall generation quality. Jimeng is particularly sensitive to special symbols in prompts and also has high safety measures.

\subsection{Quantitative Benchmark Assessment}
Currently, the concentrated quantitative evaluation benchmarks, such as CLIPScore, HPSv2, and Aesthetic Score, cannot reasonably assess model outputs in more challenging tasks. The evaluation results from GPT-4o are more comprehensive and reasonable; however, in tasks that require comparing image details for evaluation, GPT-4o's results still significantly differ from human intuitive perceptions.

{
\bibliographystyle{plain}
\bibliography{imaginee}
}

\end{document}